# Bringing order into the realm of Transformer-based language models for artificial intelligence and law

Candida M. Greco and Andrea Tagarelli[a]

Dept. Computer Engineering, Modeling, Electronics, and Systems Engineering (DIMES),
University of Calabria, 87036 Rende (CS), Italy

**Abstract.** Transformer-based language models (TLMs) have widely been recognized to be a cutting-edge technology for the successful development of deep-learning-based solutions to problems and applications that require natural language processing and understanding. Like for other textual domains, TLMs have indeed pushed the state-of-the-art of AI approaches for many tasks of interest in the legal domain. Despite the first Transformer model being proposed about six years ago, there has been a rapid progress of this technology at an unprecedented rate, whereby BERT and related models represent a major reference, also in the legal domain. This article provides the first systematic overview of TLM-based methods for AI-driven problems and tasks in the legal sphere. A major goal is to highlight research advances in this field so as to understand, on the one hand, how the Transformers have contributed to the success of AI in supporting legal processes, and on the other hand, what are the current limitations and opportunities for further research development.

## 1 Introduction

**AI for law.** Artificial Intelligence (AI) is increasingly finding its application to the legal domain. This is mainly motivated by an evidence that the volume of information produced in the legal domain is overwhelming, and also variegate due to the involvement of many different actors, such as legal professionals (lawyers, attorneys), law courts, legislators, law firms, and even citizens [Surden, 2019]. AI tools, especially those for natural language processing (NLP), are indeed conceived to help legal actors handling huge amounts of legal documents, lighten the workload in searching for and understanding text passages of interest, thus ultimately easing streamline processes and saving time for tackling more complex tasks.

AI is increasingly powering legal tech companies with the advanced technical backbone for better serving their clients more cheaply, efficiently and accurately than ever. For instance, one of the most required applications of AI in the legal tech industry is contract review and analysis, whereby companies like Klarity (klarity.com) offer solutions to review sales contracts under a particular company legal policy, or to enhance risk management (e.g., in response to legal and regulatory changes) in order to assist firms in preparing a remediation plan, like in the cases of Kira Systems (kirasystems.com) and ThoughtRiver (thoughtriver.com). Legal research is also fundamental to pursue goals of prediction of case outcomes or court rulings. In this regard, Lex Machina (lexmachina.com) can help lawyers manage a case strategy based on previous results in similar cases, or to predict how a particular judge might rule on a case; Ravel Law's Judge Analytics tool (ravellaw.com) can also visually explore data points related to specific judges, organized by court and case type; Blue J Legal (bluej.com) has developed a litigation prediction tool with a focus on tax law, which can serve for speeding up settlement negotiations; DeepJudge AG (deepjudge.ai) can automatically highlight crucial information in documents, referencing with external sources such as codes, court rulings, commercial registries; yet, Luminance (luminance.com) can facilitate the discovery of background information to carry out necessary due diligence more efficiently. Other companies, such as Intraspexion (intraspexion.com), have a focus on providing attorneys with early warning indicators when their AI tools identify lawsuit dangers. Last but not least, chatbots are increasingly being developed to improve the user experience through online self-serving, also for particular application scenarios; e.g., DoNotPay's tool (donotpay.com) helps users to automatically claim asylum in some countries.

---

[a] *Corresponding author:* andrea.tagarelli@unical.it

Moreover, AI development is highly fostered by political institutions and governments. In the 2020 report commissioned by the Administrative Conference of the United States on the use of AI in federal administrative agencies [Engstrom et al., 2020], AI tools are recognized as already enhancing agency operations across the full range of governance tasks, including enforcing regulatory mandates (e.g., for market efficiency, or workplace safety), adjudicating benefits and privileges, monitoring and analyzing risks to public health and safety, extracting information from the government's massive data streams (e.g., from consumer complaints), communicating with the public about its rights and obligations as welfare beneficiaries, taxpayers, asylum seekers, and business owners. Also, the European Union's approach to AI focuses on excellence and trust, aiming to boost research and industrial capacity while ensuring safety and fundamental rights.[1] This has been translated into concrete actions such as a review of the Coordinated Plan on AI (with EU member states) and a proposal for a regulation laying down harmonised rules on AI, known as *AI Act*.

**The breakthrough of large language models.**    On a variety of application domains, ranging from scientific literature to Web and social media, NLP research has traditionally focused on shallow text feature models, then evolved to logical and probabilistic language models, and more recently to word embeddings based on machine learning and to deep learning architectures. The latter have certainly shown improved results in a wide range of NLP benchmarks, including benchmarks specific to retrieval and classification tasks. In particular, the pre-trained *Transformer-based Language Models* (TLMs) are leading to a significant advance in NLP research to support human decision-making processes in several textual data domains [Greco et al., 2022]. TLMs possess unique characteristics in the machine and deep learning realm, so that exploiting them to solve such tasks as retrieval, understanding and prediction has a number of key advantages that can be summarized as follows. First, like any other deep neural network models, TLMs avoid to manually handle feature engineering, and hence the need for selecting features from the input text and measuring their importance. Second, like sophisticated recurrent and convolutional neural networks, TLMs represent language semantics and non-linear relationships between terms; however, TLMs are better to capture subtle and complex lexical patterns including the sequential structure and long-term dependencies, thus obtaining the most comprehensive local and global feature representations of a text sequence. Third, TLMs incorporate the so-called *attention* mechanism, which allows a learning model to assign higher weight to text features according to their higher informativeness or relevance to the learning task.

Whilst the hype for successfully using TLMs is expected to hold true also for such a challenging domain as law, however, like for other specialized fields, legal language has distinct characteristics compared to general natural languages. Given the variety of legal sources and their heterogeneous functionality, legal language has indeed specific terminology and linguistic patterns, formal syntax, and semantics relying on a particular knowledge domain, to the extent that it is often regarded as a *sublanguage*. The objective of this article is to shed light on the research advances in AI for law that TLM-based methods are achieving in the last few years, since the advent of the first Transformer [Vaswani et al., 2017], and especially of the BERT model [Devlin et al., 2019] and its many variants and extensions, which, as a whole, are sometimes referred to as "*BERTology*".

**Objectives, limitations, and scope.**    In this article, we examine the landscape of works that aim to solve legal problems by means of a subset of artificial intelligence and NLP methodologies, which corresponds to the state-of-the-art in legal AI and is represented by TLMs. We believe this is not trivial since, despite being a relatively recent technology, its evolution in the last few years has been dramatic and progressed at an unprecedented rate, also in the legal domain. To the best of our knowledge, this is the first systematic study on problems, tasks and benchmarks in the legal AI area, for which AI approaches and methods based on TLMs have been developed to address several tasks, such as retrieval, classification, prediction, entailment, information extraction and many others.

Despite our effort, this article cannot be intended to discuss definitive solutions to legal problems, tasks and applications. Moreover, please note that other deep learning technologies, from convolutional and recurrent networks to word embeddings, as well as topic modeling and classic machine-learning techniques are regarded as out of scope of this work, unless they are used in combination with TLMs.

This article is also necessarily a snapshot in time, until mid 2022, with an update to early 2023. It is hence likely that recently appeared relevant works were missing at the time of writing of this work. As such, the presented classification will need to be updated as new challenges come into focus and additional perspectives are brought to bear on legal problems.

**Comparison with existing overviews.**    The use of AI within law has long been a topic of interest in the law area, mainly focusing on understanding the relations of AI to the practice and administration of law as well as on the ethical limits of the application of AI to legal data (e.g., [Callister, 2020, Surden, 2019]). While a purely law-based perspective

---

is out of scope of this survey, it is nonetheless important to place our work into the computer-science literature that has recently surveyed the topic.

[Zhong et al., 2020] overview existing tasks and applications in legal AI, organized into three categories, namely judgment prediction, similar case matching, and legal question answering. In this regard, a summary of main characteristics is provided about existing approaches, also including a mention to early TLMs used for legal AI. An overview on legal information retrieval approaches, divided into natural language based, ontology-based, and deep-learning-based systems is also presented in [Sansone and Sperlí, 2022]. [Francesconi, 2022] describe the vision of the IAAIL president at ICAIL 2021 on the status of the AI and law discipline, including possible future perspectives that envisage the use of machine and deep learning to extract knowledge from legal data, combined with legal knowledge representation and models for legal reasoning. Also, [Song et al., 2022] provide a comparative empirical evaluation of TLMs on various NLP tasks, which aims to gain insights into the performance difference between domain-specific models and general domain models.

While the above surveys are general in the domain of AI and law, a focus on the task of querying for ad-hoc case law retrieval is taken in [Locke and Zuccon, 2022], where approaches and methods are discussed into the following categories: Boolean and natural language, conceptual search and case-based retrieval, question answering, query expansion, query reduction, search result diversification, use of citation networks, and deeper understanding of texts.

None of the above works, however, provides a systematic analysis of approaches and methods based on TLMs for legal problems and tasks, which is instead the objective of this article.

**Plan of this paper.** The remainder of this paper is organized as follows. Section 2 provides an overview of BERT and subsequent TLMs, focusing on those being used in the legal domain. Section 3 introduces main problems and relating tasks in the legal AI area that are being addressed by TLM-based methods, along with major relevant benchmarks in the legal domain. Accordingly, Section 4 describes in detail major existing TLM-based methods. In Section 5, we provide a discussion on main findings, limitations, challenges and future perspectives for legal AI based on TLMs. Section 6 concludes the paper.

## 2 Background on Transformer-based Language Models

Since their debut in the Natural Language Processing (NLP) field, Transformer models [Vaswani et al., 2017] have become a standard de-facto in the development of revolutionary deep-learning models that pushed the state-of-the-art in many challenging NLP tasks. The key point in the Transformer architecture is the use of *attention* mechanisms [Bahdanau et al., 2015] as an alternative to the conventional recurrent neural networks (RNNs) and convolutional neural networks (CNNs). The attention takes into account all the hidden states of a neural network and chooses which ones is to ignore and which ones is to remember, giving the proper weight to words' dependencies regardless of the distance in the sequence. The weights are computed through an attention function, whose purpose is to value the relevance of an input word with respect to a target word. Nowadays, attention mechanisms are ubiquitous in NLP and have boosted performances on a wide range of tasks.

Transformer is the first model to rely solely on attention mechanisms, leaving out the use of RNNs and CNNs. The attention is modeled in two ways. Firstly, it estimates the importance of words in the same sequence, i.e., input words are treated as both sources and targets, with the goal of calculating appropriate representations of the input sequence that reflect syntactic-semantic connections. This mechanism is also referred to as *self-attention*. Secondly, the input representations are weighted through attention to predict target tokens in an auto-regressive manner. The Transformer's architecture consists of multiple stacked encoder-decoder structures (Fig. 1).

The attention function used in the Transformer is called *scaled dot-product attention* [Vaswani et al., 2017]. For each token, three types of vectors are defined, namely query, key and value, with dimensions $d_k$, $d_k$ and $d_v$, respectively. A weight is assigned to each value vector, depending on the compatibility of the query and the key. The query represents the token against which the importance of the other tokens, namely the keys, have to be evaluated. Formally, given $n$ tokens in input, $\mathbf{Q} \in \mathbb{R}^{n \times d_q}, \mathbf{K} \in \mathbb{R}^{n \times d_k}, \mathbf{V} \in \mathbb{R}^{n \times d_v}$ are the representation matrices of queries, keys and values, respectively, the scaled dot-product attention $Att(\mathbf{Q}, \mathbf{K}, \mathbf{V}) \in \mathbb{R}^{n \times d_v}$ is defined as follows:

$$Att(\mathbf{Q}, \mathbf{K}, \mathbf{V}) = softmax(\frac{\mathbf{Q}\mathbf{K}^\top}{\sqrt{d_k}}) \cdot \mathbf{V}, \tag{1}$$

where term $\sqrt{d_k}$ is used as a scaling factor. By default, both $d_k$, $d_q$ and $d_v$ have a value of 64. All the layers in the model, as well as the embedding layers, produce outputs of dimension $d_{model} = 512$.

The encoders consist of multiple heterogeneous self-attention mechanisms, each capturing a different kind of word relation in order to reflect several syntactic-semantic nuances. A single type of self-attention is called *head*, so that the

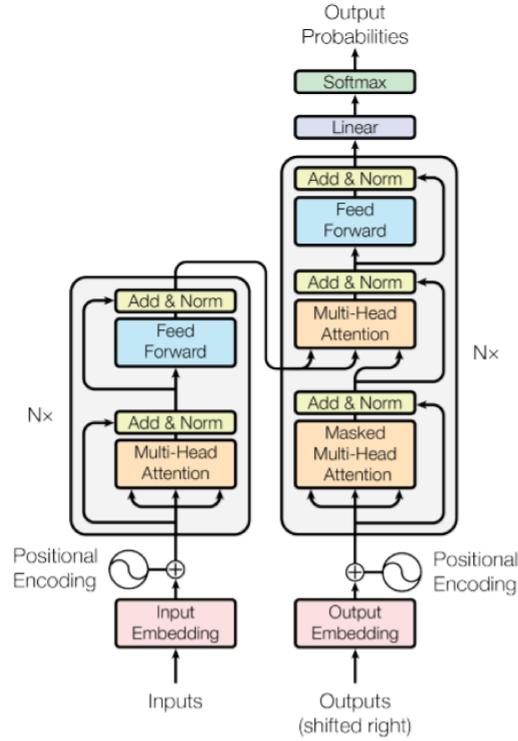

**Fig. 1.** Transformer architecture [Vaswani et al., 2017]. On the left side, the encoder component with the multi-head self-attention mechanism. On the right side, the decoder component with two attention mechanisms, a masked multi-head self-attention for decoder outputs and a multi-head attention for encoder-decoder outputs.

encoder has a *Multi-Head Self-Attention mechanism*:

$$MultiHead(\mathbf{Q}, \mathbf{K}, \mathbf{V}) = concat(\mathbf{head}_1, \ldots, \mathbf{head}_h) \cdot \mathbf{W}^{(O)}$$

$$\mathbf{head}_i = Att(\mathbf{Q}\mathbf{W}_i^{(Q)}, \mathbf{K}\mathbf{W}_i^{(K)}, \mathbf{V}\mathbf{W}_i^{(V)}), \tag{2}$$

where $\mathbf{W}_i^{(Q)} \in \mathbb{R}^{d_{model} \times d_q}$, $\mathbf{W}_i^{(K)} \in \mathbb{R}^{d_{model} \times d_k}$, $\mathbf{W}_i^{(V)} \in \mathbb{R}^{d_{model} \times d_v}$ are the projection matrices for queries, keys and values, respectively. $W^{(O)} \in \mathbb{R}^{h d_v \times d_{model}}$ is a projection matrix for the output.

The decoders have a similar structure, but they require the use of *masked* multi-head self-attention mechanism in order to predict the current word without having access to subsequent words, i.e., in an auto-regressive manner. In addition, they also use an *Encoder-Decoder Attention mechanism*, i.e., a multi-head attention over the output of the encoder stack, in order to catch relevant information for the current target prediction.

Queries, keys and values may have different meanings depending on the type of attention. In the self-attention, they all come from the output of the previous layer in the encoder (as well as in the decoder). In the encoder-decoder attention, queries come from the previous decoder layer, while keys and values come from the encoder output.

In the following, we describe main characteristics of the TLMs that have been used to address legal problems and tasks (cf. Section 3). We organize our presentation of TLMs into three parts: Section 2.1 is devoted to BERT, which is the most commonly used TLM also in the legal domain, Section 2.2 overviews main variants and extensions of BERT, and Section 2.3 contains further and more recent TLMs.

It is worth noticing that our discussion concerns not only the conceptual architecture and components of TLMs but also includes information about their implementation; moreover, in Table 1, we provide further details summarizing the pre-training configuration of each of the models, in particular: number of steps, batch size, lr (learning rate), optimization method, $\epsilon$ (i.e., constant for numerical stability for the optimization method), dropout probability, weight decay (i.e., parameter for weights regularization), warmup steps (i.e., number of steps for the warmup part of training), the activation function, vocabulary size, and max length (i.e., maximum number of tokens). In this regard, we point out that most of the publicly available software resources, including pre-trained models, can be found on the online platform Hugging Face.[2] Throughout this section, we will refer to the model implementations available in this platform, unless otherwise specified, e.g., repositories provided by the authors of the TLMs.

---

[2] https://huggingface.co/transformers, https://github.com/huggingface/transformers.

**Table 1.** Summary of pre-training settings of frequently used TLMs in works discussed in this article.

| Model | #steps | batch size | lr | optimization | ε | dropout | weight decay | warmup steps | activation | vocab. size | max length |
|---|---|---|---|---|---|---|---|---|---|---|---|
| BERT | 1M | 256 | 1e-4 | Adam $\beta_1$ = 0.9 $\beta_2$ = 0.999 | 1e-6* | 0.1 | 0.01 | 10K | gelu | 30K | 512 |
| RoBERTa | 500K | 8K | 6e-4 (base) 4e-4 (large) | Adam $\beta_1$ = 0.9 $\beta_2$ = 0.98 | 1e-6 | 0.1 | 0.01 | 24K (base) 30K (large) | gelu | 50K (sub-words) | 512 |
| DeBERTa | 1M 500K (v3) | 2K 8K (v3) | 2e-4 (base, large) 1.5e-4 (1.5B) 3e-4 (v3 large) 6e-4 (v3 base, small) | Adam beta1=0.9 beta2=0.999 beta2=0.98 (v3) | 1e-6 | 0.1 | 0.01 | 10K | gelu ** | 50K (base, large) ** 128K (1.5B) *** | 512 ** |
| DistilBERT | 3 (epochs)° | 5° | 5e-4° | AdamW° $\beta_1$ = 0.9 $\beta_2$ = 0.98 | 1e-6° | 0.1** | 0.0° | in the code° | gelu ** | 28996 (cased) ** 30K (uncased) ** | 512 ** |
| ALBERT | 125K | 4096 | 1.76-3 | Lamb°°° $\beta_1$ = 0.9 $\beta_2$ = 0.999 | 1e-6°°° | 0°° | 0.01°°° | 3125 °°°° | gelu | 30K | 512 |
| T5 | 524K | 128 | 1e-2 | AdaFactor†† $\beta_1$=0 | $\epsilon_1$= 1e-30†† $\epsilon_2$= 0.001†† | 0.1 | 0 ††† | 10K | relu | 32K | 512 |
| ELECTRA | 1M (small) 766K (base) 400k-1.75M (large) | 128 (small) 256 (base) 2048 (large) | 2e-4 (small) 2e-4 (base,large) | Adam $\beta_1$ = 0.9 $\beta_2$ = 0.999 | 1e-6 | 0.1 | 0.01 | 10K | gelu ** | 30K † | 128 (small) 512 (others) |
| Longformer | like RoBERTa | 64 | 3e-5 | like RoBERTa | like RoBERTa | like RoBERTa | like RoBERTa | 500 | like RoBERTa | like RoBERTa | 4096 |

* from `https://github.com/google-research/bert/blob/master/optimization.py`
** from `https://github.com/huggingface/transformers`
*** from `https://github.com/microsoft/DeBERTa`
° from `https://github.com/huggingface/transformers/tree/77321481247787c97568c3b9f64b19e22351bab8/examples/research_projects/distillation`
°° from ALBERT paper: «We also note that, even after training for 1M steps, our largest models still do not overfit to their training data. As a result, we decide to remove dropout to further increase our model capacity»
°°° from `https://github.com/google-research/albert/blob/master/optimization.py`
°°°° from `https://github.com/google-research/albert/blob/master/run_pretraining.py`
† from `https://github.com/google-research/electra/blob/master/configure_pretraining.py`
†† from `https://github.com/google-research/text-to-text-transfer-transformer`
††† from `https://huggingface.co/docs/transformers/main_classes/optimizer_schedules`

## 2.1 BERT

The first known Transformer-based model which has revolutionized the way to approach the natural language understanding challenges, pushing the state-of-the-art in many demanding NLP benchmarks, is Bidirectional Encoder Representations from Transformers (BERT) [Devlin et al., 2019].[3] BERT is a Transformer-based model that has represented a breakthrough in several NLP benchmarks and it is still considered a must-have baseline [Rogers et al., 2020]. It essentially consists of a stack of Transformer encoder layers (Fig. 1). The key aspects of BERT are the *bidirectional unsupervised pre-training* and the beneficial effect of having a unified architecture across different tasks. The framework uses the pre-training fine-tuning paradigm. In the pre-training stage, the model uses a masked language objective to get deep bidirectional representations from unlabeled text by jointly conditioning on left and right context-words in all encoder layers. A deep bidirectional model has shown to outperform a conventional unidirectional model or even a shallow left-to-right and right-to-left concatenated model (like ELMo).

To obtain a deep bidirectional representation of unlabeled text in the pre-training phase, BERT employs a particular pre-training objective task called *Masked Language Modeling* (MLM). This task is to predict masked input words from unlabeled text by conjointly conditioning on left and right context-words in all encoder layers. In other words, BERT considers a deeply bidirectional context in the prediction of masked tokens, and as a result it builds more expressive and meaningful word embeddings than a conventional unidirectional or shallow bidirectional models (Fig. 2). In addition, BERT uses a second pre-training objective task called *Next Sentence Prediction* (NSP) to cover a variety of downstream tasks involving sentence pairs. Given two word sequences as input, the NSP task is to determine if the second sequence is subsequent to the first in the original document.

Each word is first tokenized into *word pieces* [Wu et al., 2016], which forms a vocabulary of 30,522 tokens.[4] The input representation is the sum of token embeddings with segment and position embeddings. Segment embeddings take into account the token location, i.e., the tokens of a sentence have all the same segment encoding. Position embeddings keep track of the token absolute position in the text. The first token of the sequence is always a special token called [CLS], and the output representation of this token is considered to be representative of the whole input sequence. In the NSP task, the two sentences in an input sequence are separated to each other by another special token called [SEP]. In the MLM task, words to be masked are replaced with the [MASK] special token. The introduction of [MASK] has the advantage to allow the model to rely jointly on left and right context. However, it also cause a mismatch between the pre-training and the subsequent, downstream fine-tuning phase, since the latter does not introduce any mask. Therefore, to mitigate this issue, not all the tokens are replaced with [MASK], but only a certain percentage; in particular, from all tokens, 15% are selected to be masked, using a random token in 10% of cases, [MASK] in 80% of cases and leave the original token in the remaining 10%. The model does not know which words have been randomly replaced, so it has to keep a distributional contextual representation for each token.

---

[3] `https://github.com/google-research/bert`
[4] Using the WordPiece tokenization process, the vocabulary is obtained using a data-driven approach: given a training corpus and a number $w$ of word pieces for the vocabulary, the task is to select $w$ word pieces such that the segmented corpus contains as much unsegmented words as possible. WordPiece tokenization has shown to deal with the out-of-vocabulary words better than standard tokenization procedures.

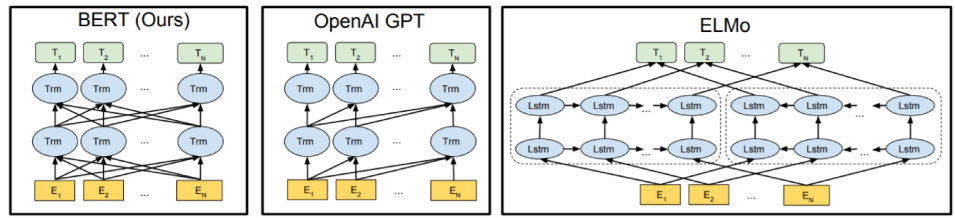

**Fig. 2.** Difference between deeply bidirectional BERT architecture, left-to-right GPT architecture and shallow bidirectional ELMo architecture [Devlin et al., 2019].

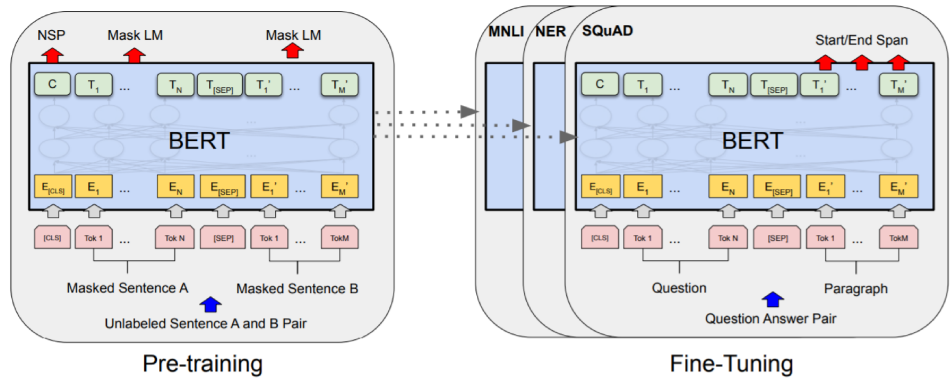

**Fig. 3.** BERT Pre-training and Fine-tuning [Devlin et al., 2019].

BERT is originally pre-trained using two unlabeled corpora, named Book Corpus and English Wikipedia, for a total of 3,300M words. The input sequence has a limit of 512 tokens. For NSP, the training examples are sampled so that in 50% of cases the second sentence is actually subsequent to the first, while in the remaining 50% the second sentence is randomly selected from the corpus.

The authors released BERT in two format sizes: BERT-*base* and BERT-*large*. BERT-*base* consists of 12 stacked encoder layers with 12 attention heads and a hidden size of 768, for a total of 110M parameters, while BERT-*large* has 24 stacked encoder layers, 16 attention heads and a hidden size of 1024 (340M total parameters).

Using BERT for downstream tasks is straightforward, as it just requires one additional task-specific layer on top of the model and a fine-tuning phase on parameters. The input representation is analogue to the pre-training one described above. Typically, the final hidden state corresponding to the [CLS] token is used as the input of the additional top layer for tasks such as classification and entailment; analogously, the final hidden states of the input tokens are fed into the additional top layer for token-level tasks, such as sequence-tagging and question-answering.

BERT contextual representations of words can also be used as word embeddings for feature-based models. The authors experimented using BERT embedding as input for a randomly initialized two-layer BiLSTM. They found that great performance can be achieved by combining the outputs of BERT layers, especially when the last four layers' outputs of BERT-*large* are concatenated.

Several studies have examined the encoded knowledge in the BERT weighted representations [Rogers et al., 2020]. It has been shown that BERT naturally learns syntactical information, such as parts of speech and syntactic chunks, and its representations are characterized as being hierarchical. BERT has also semantic knowledge, as it encodes information like semantic roles, entity types, relations etc. Several studies stated that the most information about the order of words is in the lower layers of BERT, syntactic information is primarily in the middle layers, while the most task-specific information is in the final layers. Semantic information, instead, is all over the model since it pervades all the language. In the knowledge induction by filling in the blanks, a study has shown that, for some relation types, BERT can be competitive with methods supported by knowledge bases.

BERT was tested on a number of benchmarks, including GLUE [Wang et al., 2018], SWAG [Zellers et al., 2018], SQuAD v1.1 [Rajpurkar et al., 2016] and v2.0 [Rajpurkar et al., 2018], getting the top results on eleven tasks. In fact, it advanced the state-of-the-art, obtaining 80.5 score (7.7% point improvement compared to previous models) on the official GLUE leaderboard,[5] 86.3% test accuracy on SWAG (8.3% point improvement compared to GPT), 1.5 more points on SQuAD v1.1 Test F1 and 5.1 more points on SQuAD v2.0 Test F1 compared to previous models.

Several instances of BERT model have been released over the years, which mainly differ from each other in terms of model size (as above discussed) as well as training objective and case folding. For the legal domain, the following models appear to be mostly used in the works discussed in Section 4: the case-sensitive `bert-base-cased`[6], the uncased `bert-base-uncased`, the instance trained on sentence-pair classification task for FAQ retrieval, dubbed `bert-based-faqir`,[7] the Japanese instance called `bert-base-japanese`[8] as well as the Japanese `bert-base-japanese-whole-word-masking`[9] and `BERT-base_mecab-ipadic-char-4K_whole-word-mask` (abbrv. BERTjpcwwm).[10] The latter are the Japanese instances trained using the MLM objective with the *whole word masking* technique, in which all the tokens related to a single word are masked.

There is also a multilingual variant of BERT, called *m-BERT*.[11] Actually, only the base version of the model is available (12 layers, 12 attention heads, 768 hidden size, 110M parameters in total), trained for more than 100 languages and a specific version for Chinese.

Over the time, BERT has inspired extensive research in the development of different variants and enhancements of the model. In the next section, we discuss the most relevant BERT-inspired models that have been applied in legal tasks.

## 2.2 BERT variants and extensions

**RoBERTa.** RoBERTa [Liu et al., 2019] is conceived to improve BERT based on the presumed under-training and non-optimal setting of the key hyperparameters of BERT.[12] RoBERTa indeed develops a more robust and optimized approach through two main interventions. Firstly, the amount of training data is increased by adding three further corpora besides Book Corpus and English Wikipedia, namely CC-News, which consists of the English portion of CommonCrawl news dataset (76 GB), Stories (31 GB), and OpenWebText (38GB). Secondly, RoBERTa is pre-trained much longer, which is related to the finding that decreasing BERT's number of steps and increasing the batch size leads to better results, with the same computational cost. Table 1 reports further details regarding the pre-training configuration settings.

Substantial changes on the pre-training procedure have been made too. Pre-training sequences have a length of 512 tokens, discarding BERT short-sequence strategies — in BERT, 90% of the pre-training steps involve sequences of length 128 tokens, and the remaining 10% use sequences of 512 tokens. The next sentence prediction task is removed, as it is not regarded as relevant to the model's performance. Moreover, the masked language modeling task is modified, by introducing a *dynamic* masking that avoids the use of the same mask for each instance and for each epoch.

The text representation of RoBERTa is obtained through a byte-level vocabulary containing 50K subwords units, in contrast with the character-level vocabulary of 30K tokens used by BERT. The text segmentation method used in RoBERTa is the Byte-Pair Encoding (BPE) [Sennrich et al., 2016].

Like for BERT, two sizes of RoBERTa have been released: RoBERTa-*base* (12 layers, 12 attention heads, 768 hidden size) and RoBERTa-*large* (24 layers, 16 attention heads, 768 hidden size). RoBERTa has shown to overcome BERT in classic benchmarks such as GLUE, SQuAD v2.0 and RACE [Lai et al., 2017].

**SBERT and SRoBERTa.** SBERT and SRoBERTa [Reimers and Gurevych, 2019] are modified versions of BERT and RoBERTa, respectively, especially designed for tasks such as semantic textual similarity, clustering, and information retrieval via semantic search.[13] One problem is the massive computational overhead concerning the search for the most similar sentence-pairs in a collection of thousands of sentences. SBERT and SRoBERTa drastically reduce the

---

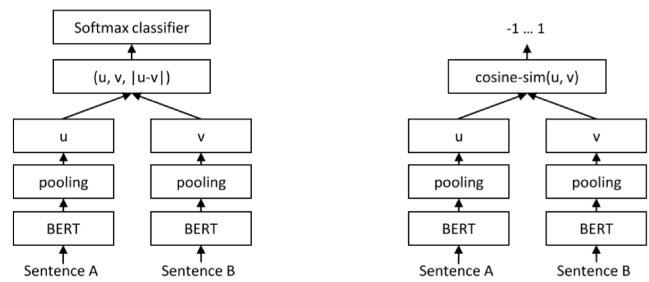

**Fig. 4.** SBERT architectures for classification (on the left) and inference/regression (on the right) tasks. [Reimers and Gurevych, 2019]. In the first case, the concatenation of the sentence embeddings $u$ and $v$ and the element-wise difference $|u - v|$ is the input for a softmax classifier. In the second case, the cosine similarity between $u$ and $v$ is calculated.

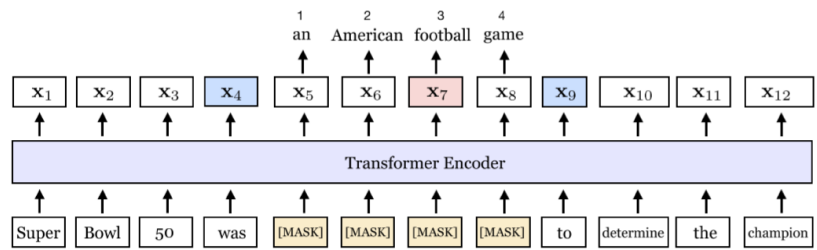

**Fig. 5.** SpanBERT training [Joshi et al., 2020]. The masked span $x_5, x_6, x_7, x_8$ is predicted using the boundaries $x_4$ and $x_9$.

computational effort of BERT for finding the most similar pair in a collection, while maintaining the same accuracy. SBERT (resp. SRoBERTA) fine-tunes BERT (resp. RoBERTa) based on a Siamese network architecture which, given two pre-trained BERT (resp. RoBERTa) models, one for each input sentence, it ties the models' weights which are updated during fine-tuning, to obtain semantically-expressive fixed-sized sentence embeddings based on a pooling operation. The two resulting sentence embeddings are concatenated with their element-wise difference, prior to the softmax layer for class prediction.

Siamese and triplet networks are used to obtain semantically meaningful sentence embeddings, then these embeddings can be compared using similarity criteria (e.g., cosine similarity). The final layer structure depends on the task at hand to be optimized (Fig. 4). For classification purpose, the sentence embeddings are concatenated with their element-wise difference and fed into a softmax classifier, using an optimized cross-entropy loss. For regression purpose, cosine similarity between sentence embeddings and mean-squared-error loss are used. In addition, the authors experimented the model with a triplet objective function.

SBERT has been trained on NLI data (SNLI and MultiNLI) and evaluated on semantic textual similarity (STS) tasks, both unsupervised, i.e., without any task-specific training data, and supervised STS tasks, in which the STS benchmark was used for fine-tuning. SBERT sentence embeddings has been evaluated using the *SentEval* toolkit, in which the embeddings are used as features for a logistic regression classifier. Using RoBERTa in place of BERT (SRoBERTa), the model generates quite similar sentence embeddings and obtain similar performance. Although SBERT embeddings are not designed for transfer-learning on other tasks, experiments reveal that they can achieve good performance in most SentEval transfer learning tasks.

**SpanBERT.** SpanBERT [Joshi et al., 2020] introduces a novel pre-training approach that differs from BERT in three aspects: the masking scheme, the training objective and the sequence training procedure.[14]

While maintaining the same percentage distribution of masking as BERT, SpanBERT masks adjacent random *spans* rather than random single tokens. The span masking is performed by sampling spans of text. The span length is picked from an unbalanced geometric distribution that favors short length selection, where the starting point of span masking is randomly selected too. Eventually, complete words only (instead of sub-words) are masked.

In addition, SpanBERT introduces a novel pre-training objective, called *span-boundary objective* (SBO). This task is to predict each token of the span by using only the observed tokens at the boundaries, i.e., the first token before the start and first token after the end of the span (Fig. 5). The idea is to encourage the model to record as much span information as possible in the boundary output encodings. Each token of the span is then represented using the boundary output encodings and relative position embedding of the target token. Like in MLM, SBO minimizes the cross-entropy loss. The overall loss is the sum of both MLM and SBO objectives for each token in the span:

$$
\begin{aligned}
L(x_i) &= L_{MLM}(x_i) + L_{SBO}(x_i) \\
&= -log P(x_i | \mathbf{x}_i) - log P(x_i | \mathbf{y}_i),
\end{aligned}
\tag{3}
$$

where $x_i$ is the $i$-th token in the span, $\mathbf{x}_i$ is the encoder output and $\mathbf{y}_i$ is the model representation of $x_i$, which corresponds to a function of the external boundaries and the relative token position in the span:

$$
\mathbf{y}_i = f(x_{s-1}, x_{e+1}, p_{i-s+1}),
\tag{4}
$$

where $s$ and $e$ indicate the start and the end positions of the span, $\mathbf{x}_{s-1}$ is the boundary on the left of the span, $\mathbf{x}_{e+1}$ is the boundary on the right of the span, $p_{i-s+1}$ is the relative token position and $f(\cdot)$ is a function implemented as a 2-layer feed-forward network. The model gets rid of the NSP task, since the authors suggest that using only single sequences in the training process benefits from two aspects: the model can take advantage from longer contexts and the noise, coming from the context of other documents, is removed.

---

[14] https://github.com/facebookresearch/SpanBERT

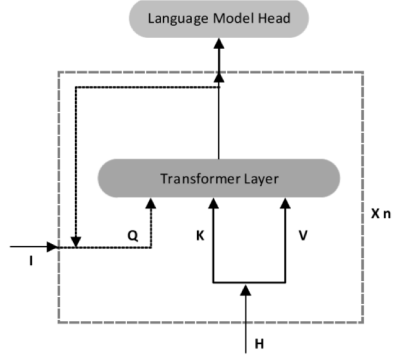

**Fig. 6.** DeBERTa's enhanced mask decoder (EMD) [He et al., 2021]. There are $n$ stacked EMD layers and two inputs: $H$ and $I$. $H$ is the hidden state from previous Transformer layer, while $I$ represents any additional information ($H$, absolute position or output from previous EMD layer). When $I = H$ and $n = 1$, EMD is equivalent to a BERT layer.

The implementation is the same as BERT-*large*, also in terms of corpora (i.e., Book Corpus, English Wikipedia) and tokenization method (i.e., WordPiece), but changing some configuration setting (Table 1). Inspired by RoBERTa, in the training process the model uses different masks at each epoch and considers only sequences of 512 tokens till the end of the document, discarding BERT's short-sequence strategies.

SpanBERT has been tested on several benchmarks regarding question answering, coreference resolution and relation extraction, as well as nine GLUE tasks. Experimentation reveals that SpanBERT can outperform BERT on most benchmarks.

**DeBERTa.** DeBERTa [He et al., 2021] introduces two novel techniques to improve upon BERT and RoBERTa: the *disentangled attention mechanism* and an *enhanced mask decoder*.[15] The former consists in computing the attention weights among words using two disentangled matrices, which represent word contents and positions, respectively. In fact, unlike BERT, DeBERTa separates content and relative position encodings in two vectors, so that the attention score for a word pair is calculated by summing up the content-to-content, content-to-position and position-to-content attention:

$$
\begin{aligned}
A_{i,j} &= \{\mathbf{h}_i, \mathbf{p}_{i|j}\} \times \{\mathbf{h}_j, \mathbf{p}_{j|i}\}^\top \\
&= \mathbf{h}_i \mathbf{h}_j^\top + \mathbf{h}_i \mathbf{p}_{j|i}^\top + \mathbf{p}_{i|j} \mathbf{h}_j^\top,
\end{aligned}
\tag{5}
$$

where $A_{i,j}$ is the attention score between tokens $i$ and $j$, $\mathbf{h}_i$ (resp. $\mathbf{h}_j$) is the content representation of $i$ (resp. $j$), $\mathbf{p}_{i|j}$ (resp. $\mathbf{p}_{j|i}$) is the relative position with token $j$ (resp. $i$).

DeBERTa is pre-trained using the MLM task, but unlike BERT, it has no absolute positional embeddings. The absolute position is very important for the prediction of masked tokens, in order to discern syntactical nuances in a sentence. For this purpose, DeBERTa incorporates the absolute positional embeddings just before the *softmax* layer responsible of masked word predictions (Fig. 6). This approach is called *enhanced mask decoder* (EMD) and allows the model to consider relative position in all layers and absolute information when masked words have to be decoded. The pre-training settings of large DeBERTa models is analogue to BERT except for the use of the BPE vocabulary. Like BERT, training data is composed by English Wikipedia and Book Corpus (already used in BERT) with the addition of OpenWebText and Stories, for a total of about 78Gb. Following RoBERTa, DeBERTa uses a dynamic batch data to handle documents shorter than 512 tokens. Inspired by SpanBERT, it includes also span masking strategies. DeBERTa pre-training parameters settings are reported in Table 1.

---

[15] https://github.com/microsoft/DeBERTa

DeBERTa is implemented with three version sizes: DeBERTa-*base* (12 layers, 12 attention heads, 768 hidden size), DeBERTa-*large* (24 layers, 16 attention heads, 1024 hidden size) and DeBERTa-1.5*B* (48 layers, 24 attention heads, 1536 hidden size). DeBERTa-1.5*B* is fine-tuned on a new virtual adversarial training method, called *Scale-invariant-Fine-Tuning* (SiFT), that normalizes word embeddings into stochastic vectors, which reveals to be beneficial for improving training stability in downstream tasks.

In GLUE tasks, DeBERTa outperforms BERT, RoBERTa and ELECTRA (cf. Section 2.3), with much less pre-training data than RoBERTa and ELECTRA (78G against 160G). DeBERTa shows outstanding performance also in SQuAD, RACE and SWAG benchmarks. DeBERTa-1.5*B* was tested on SuperGLUE [Wang et al., 2019] benchmarks, where it surpassed for the first time human performance in terms of macro-average score.

In [He et al., 2023], a novel version of DeBERTa, called DeBERTaV3, is proposed. It is based on a previous model update, DeBERTaV2, where the major changes are in the vocabulary, tokenizer, input encoding and shared projection matrices (position and content) in attention layers. DeBERTaV3 is pre-trained using the same settings as the original DeBERTa but following ELECTRA [Clark et al., 2020] in the use of a generator-discriminator setting and replacing the MLM task with the ELECTRA's pre-training task, called *Replace Token Detection*. In eight GLUE tasks, DeBERTaV3-*large* (24 layers, 12 attention heads, 1024 hidden size) obtains an average score of 91.37 on development set, outperforming ELECTRA and RoBERTa. It was also tested on SQuAD v2.0, RACE and SWAG, outperforming previous models on development set.

In [He et al., 2023], a multilingual version of DeBERTAV3-*base* (12 layers, 12 attention heads, 768 hidden size), called mDeBERTa-*base*, is also proposed. It has the same model structure as DeBERTAV3-*base*, and it was trained with the same corpus as XLM-RoBERTa [Conneau et al., 2020] and the same SentencePiece vocabulary as mT5 [Xue et al., 2021]. The pre-training setting is similar to XLM-RoBERTa except for the number of steps. It was tested on XNLI [Conneau et al., 2018] and obtained higher results than previous base models (like mT5-*base* and XLM-RoBERTa-*base*) in all languages and under both zero-shot cross-lingual-transfer and zero-shot translate-train-all settings.

**DistilBERT.** DistilBERT [Sanh et al., 2019] stands as a counterpart of the ever larger pre-trained language models, demonstrating that a smaller but still general-purpose language model can achieve comparable performance in much less time.[16] This leads to significant benefits, since large-scale models often have hundred million parameters and their training phase have considerable computational costs and memory requirements.

The key aspect of DistilBERT corresponds to the *knowledge distillation*, a transfer learning technique whereby the knowledge of a larger pre-trained model, called "teacher", is passed to a smaller model, the "student", through a training phase wherein the student has to reproduce the same results of the teacher. In DistilBERT, the training objective is a linear combination of three loss functions: distillation loss, masked language modeling loss (same as BERT) and cosine embedding loss. The distillation loss is a cross-entropy loss function taking into account both the probabilities estimated by the teacher and the student:

$$L_d = \sum_i t_i * \log(s_i), \tag{6}$$

where $t_i$ (resp. $s_i$) is the probability coming from the teacher (resp. from the student). Cosine embedding loss is used to measure the degree of similarity of two input vectors and aims to align the directions of student and teacher vectors.

The architecture of DistilBERT is the same as BERT except for the absence of the token embedding and the pooler layer. Moreover, the number of layers is halved compared to BERT. Following RoBERTa, the model was trained using dynamic masking, without the NSP task and with very large batches, but keeping the same corpora as the original BERT. As a result, DistilBERT has 40% fewer parameters than BERT-*base* and is 60% faster. Further details about pre-training parameter setting are reported in Table 1.

In the GLUE benchmark, DistilBERT obtains almost the same score as BERT in the development set. In SQuAD v1.1, it obtains good results, using a second knowledge distillation in the fine-tuning phase with BERT fine-tuned on SQuAD as teacher model.

---

[16] `https://github.com/huggingface/transformers/tree/main/examples/research_projects/distillation`

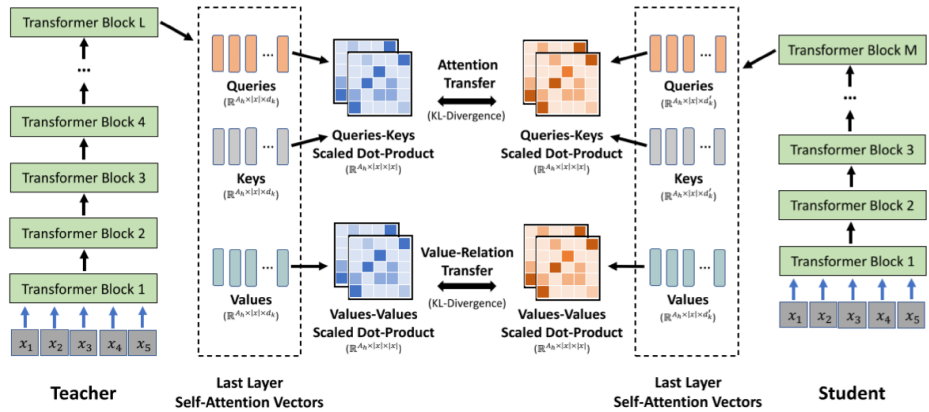

**Fig. 7.** Deep Self-Attention Distillation of MiniLM [Wang et al., 2020b].

**MiniLM**  MiniLM [Wang et al., 2020b] also adopts a knowledge distillation approach. By limiting the distillation process to the last teacher level, MiniLM aims to alleviate the difficulty in performing a layer mapping between the teacher and the student and to make the number of the student layers more flexible. Moreover, the model introduces the scaled dot-product between values, in addition to the scaled dot-product of queries and keys, to transfer the value relations (Fig. 7), thus achieving a deeper mimicking of the teacher as well as introducing more knowledge about word dependencies. Optionally, an intermediate-size student model ("the teacher assistant") can be interposed between the teacher and the student in the knowledge distillation process to alleviate the size gap between student and teacher and improve the model performance of smaller students. The knowledge of the teacher is hence distilled into the teacher assistant, which guides the training of the student.

MiniLM is trained for mono and multi-lingual tasks. In the mono-lingual setting, the teacher is the BERT-*base*, while for the multi-lingual setting the teacher is XLM-RoBERTa-*base*. The model proves to outperforms DistilBERT on SQUAD v2.0 and several tasks of GLUE, and achieves competitive performance on XNLI w.r.t. mBERT.

**AlBERT.**  AlBERT [Lan et al., 2020] stems from the same considerations of the DistilBERT's authors regarding the excessive size of pre-trained language models, which raises concerns in terms of memory and time costs.[17] In ALBERT, two parameter reduction techniques are proposed to mitigate the memory consumption and speed up the training time.

The first technique is a *factorized embedding parameterization*, which consists in the decomposition of word embeddings in smaller matrices. In BERT, the size $E$ of the context-independent wordpiece embeddings is equivalent to the context-dependent hidden layer size $H$ ($E \equiv H$); however, it is assumed in AlBERT that these sizes should be untied, and in particular $H \gg E$ should hold. In the factorized embedding parameterization, vocabulary vectors of size $V$ are first projected in a lower-dimensional space of size $E$ and then projected to the hidden space of size $H$. Therefore, the embedding parameters are reduced from $\mathcal{O}(V \times H)$ to $\mathcal{O}(V \times E + E \times H)$, which is a significant reduction since $H \gg E$.

The second technique is the *cross-layer parameter sharing*. As default the model shares all parameters across layer, although other sharing techniques can be used. As a result, AlBERT has much smaller parameter size and the training is 1.7 times faster than BERT without severely affecting the performance. In addition, AlBERT configurations can be scaled up much larger than BERT. There are four sizes of AlBERT models: AlBERT-*base* (12 layers, 12 attention heads, 768 hidden size), AlBERT-*large* (24 layers, 16 attention heads, 1024 hidden size), AlBERT-*xlarge* (24 layers, 16 attention heads, 2048 hidden size), AlBERT-*xxlarge* (12 layers, 64 attention heads, 4096 hidden size). BERT-*base* has 108M total parameters, while AlBERT-*base* has only 12M parameters and AlBERT-*xlarge* has 60M parameters. AlBERT-*xxlarge* is around 70% of BERT-*large* parameters (235M against 334M), but is about 3 times slower because it has a larger structure. ALBERT-*large* has about 18 times fewer parameters compared to BERT-*large*.

To further improve the performance, the authors replaced the NSP task with another pre-training objective task, called *Sentence-Order Prediction* (SOP). Given two segments of text from the same document, the task is to predict if the second segment comes next the first segment in the document, i.e., they are consecutive (positive examples), or if the original order of the segments has been swapped (negative examples). In other words, the task is designed to train the model to understand the order of segments, focusing on *discourse coherence* predictions.

AlBERT is pre-trained on the same BERT corpora, with input length up to 512, vocabulary size of 30k. Text is tokenized using SentencePiece. Table 1 contains further details on pre-training parameters. AlBERT was fine-tuned under two settings: single-model and ensembles. It achieves better results than previous models like BERT in several downstream tasks.

---

[17] https://github.com/google-research/ALBERT

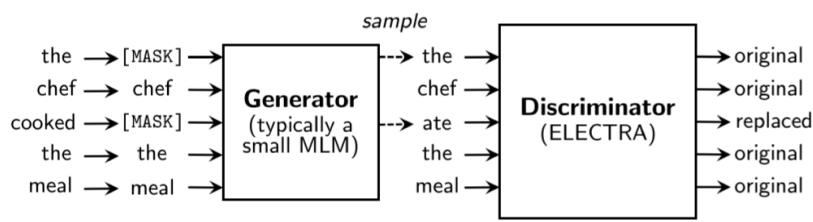

**Fig. 8.** Replaced token detection task in ELECTRA [Clark et al., 2020].

## 2.3 Other Transformers

Besides BERT and its close variants and extensions, there is a body of TLMs that extend different or additional parts of the full architecture of Transformer compared to BERT. Following [Yang et al., 2023], we organize our overview of such TLMs by distinguishing encoder-only models, decoder-only models, and encoder-decoder models, depending on whether the language modeling is also auto-regressive. Furthermore, the last part of this section contains TLMs that, regardless of the above categorization, have been especially designed for a specific task, or for dealing with long documents.

Again, please note that our overview of each of the models is necessarily brief and limited to those TLMs that have been used so far for legal tasks, to the best of our knowledge.

### 2.3.1 Encoder-only models

**ELECTRA.** The ELECTRA model [Clark et al., 2020] originates from the observation that the use of a discriminative approach in the pre-training phase, which allows the model to learn from all input tokens, is computationally more efficient compared to a generative approach like MLM in BERT.[18] Within this view, a new pre-training task, called *Replaced Token Detection* (RTD), is proposed such that, instead of using a mask like in MLM, tokens are replaced with plausible generated tokens. In this way, the model is trained to discriminate real tokens from credible yet fake tokens. This type of corruption overcomes the BERT's mismatch between pre-training phase, in which artificial [MASK] tokens are used, and fine-tuning phase (where there are no artificial tokens). The key aspect of the task is that the model can learn from all input tokens and not just from a small subset, thus achieving a significant speed-up in the training phase compared with BERT.

ELECTRA consists of a generator followed by a discriminator, both consisting of a Transformer encoder. The generator is trained to produce plausible tokens, replacing real tokens in a random set of positions. The training is performed in a MLM manner: given a random set of masked tokens, the generator learns to predict the original tokens. If the prediction is correct then it is regarded as "real", otherwise as "fake". The discriminator is trained to predict if the current token is real or fake (Fig. 8). After pre-training, the generator is discarded so that only the discriminator is used in the fine-tuning phase.

The model architecture and most hyperparameters are equal to BERT's (see Table 1). ELECTRA is available in three model sizes: ELECTRA-*small* (12 layers, 4 attention heads, 256 hidden size), ELECTRA-*base* (12 layers, 12 attention heads, 768 hidden size), ELECTRA-*large* (24 layers, 16 attention heads, 1024 hidden size). In the token masking process, the 15% of tokens are masked out (like in BERT), except for ELECTRA-*large* in which the percentage is increased to 25%. ELECTRA-*large* is trained with 400 steps or 1.75M steps: in the first case, it achieved performance in GLUE comparably to RoBERTa, with much less computational cost (less then 1/4 pre-training cost), whereas in the second case, it outperforms RoBERTa still with less computational costs. ELECTRA-*large* outperforms RoBERTa also in SQuAD v2.0. Indeed, ELECTRA-*base* overcomes BERT-*large* on GLUE development set (85.1 against 84.0).

**XLM-RoBERTa.** XLM-RoBERTa [Conneau et al., 2020, Pant and Dadu, 2020, Dadu and Pant, 2020] is a highly scalable cross-lingual model which was designed to improve cross-lingual language understanding (XLU).[19] It has indeed shown that low-resource language performance can be improved if scaled with similar high-resource language during pre-training, yet there is a trade-off between the positive transfer and the capacity dilution of the model (i.e., the number of parameters). More precisely, since model capacity is limited by time and memory constraints, the capacity reserved for a single language is diluted among all languages as the number of languages grows in a fixed-sized model, and once a certain level of dilution is exceeded, the overall performance is degraded. Adding more capacity to the model alleviates this condition, but it is unsuitable for models with modest sizes. The authors also deal with the appropriate allocation of the model capacity across high-resource and low-resource languages and the scaling of both model and vocabulary size with respect to the number of languages.

---

[18] https://github.com/google-research/electra
[19] https://github.com/facebookresearch/xlm

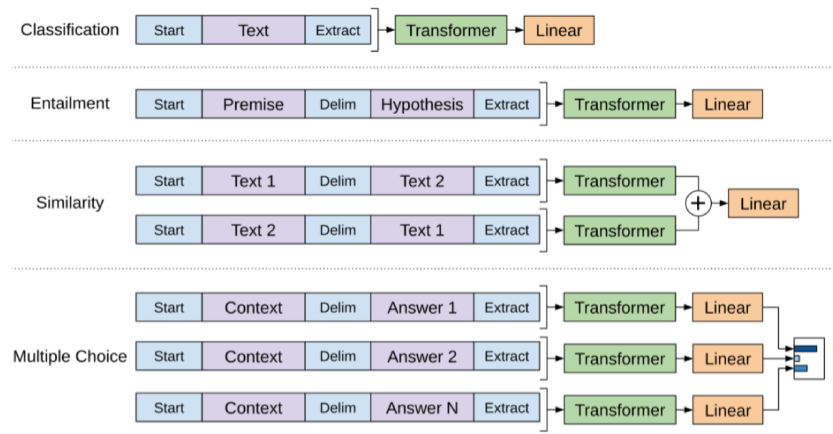

**Fig. 9.** Task-specific input conversion to token sequences in GPT [Radford et al., 2018].

The architecture of XLM-RoBERTa emulates the XLM approach [Conneau and Lample, 2019] for training objective, languages, and data, with few refinements to increase performance at scale. The model is pre-trained with a multilingual MLM objective, using monolingual data and sampling pieces of text from each language. The training data is obtained from a filtered and cleaned version of CommonCrawl and consists of more than 2TB of data in 100 languages. Sub-word tokenization is directly applied on the raw text for all languages using a SentencePiece model. Following RoBERTa, the authors show that training the model longer and with a larger-scale corpus yields improvements in the model performance.

XLM-RoBERTa is available in two model sizes: XLM-RoBERTa-*base* (12 layers, 12 attention heads, 768 hidden size, 270M parameters) and XLM-RoBERTa-*large* (24 layers, 16 attention heads, 1024 hidden size, 550M parameters). In [Goyal et al., 2021], two larger versions are proposed: XLM-RoBERTa-*xl* (36 layers, 32 attention heads, 2560 hidden size, 3.5B parameters) and XLM-RoBERTa-*xxl* (48 layers, 32 attention heads, 4096 hidden size, 10.7B parameters).

XLM-RoBERTa has been tested on cross-lingual understanding tasks as XNLI, CoNLL (2002 [Sang, 2002] and 2003 [Sang and Meulder, 2003]) and MLQA. With regard to XNLI, the fine-tuning was carried out on the English training set (cross-lingual transfer) and also on all training sets, previously translated (translate-train-all). On CoNLL, the fine-tuning considered the English set (cross-lingual transfer), each set separately (per-language performance) and all sets together (multilingual learning). XLM-RoBERTa has shown to outperform previous models like m-BERT and XLM in each of these cross-lingual benchmarks. Also, competitive results were obtained also in monolingual benchmarks like GLUE, proving that it is possible for a multilingual model to achieve on a single language the same performance than monolingual models.

### 2.3.2 Decoder-only models

**GPT.** GPT [Radford et al., 2018] is the first model of the Generative Pre-trained Transformer family developed by OpenAI.[20] Based on a Transformer decoder, with 12 layers, 12 attention heads, and 768 hidden layer size, GPT is an autoregressive model pre-trained on the BookCorpus dataset with a causal language modeling objective (CLM), i.e., to predict the next token given the left-side context. The model uses a BPE vocabulary size of 40K and maximum sequence length of 512 tokens. To address the downstream tasks in the fine-tuning phase, the inputs of the different tasks are converted into token sequences through the use of special tokens, in order to be processed by the model (Fig. 9). The model achieved better performance then previous competitors at that time on several tasks, like text classification, natural language inference, question answering, semantic similarity.

---

[20] https://openai.com/

**GPT-2.** GPT-2 [Radford et al., 2019] is the successor of GPT, conceived to perform downstream tasks in a zero-shot learning setting. Like GPT, GPT-2 is pre-trained on a CLM task but using WebText, a larger amount of free text obtained through web scraping and consisting of millions of webpages whose selection has been curated by humans. The architecture of GPT-2 is largely based on GPT, but it has more than one order of magnitude of additional parameters (1.5B parameters). The context size is increased to 1024 tokens and the vocabulary reaches 50K tokens.

To address downstream NLP tasks, the output of GPT-2 is conditioned on the input and task type. That is, in addition to the text, the model receives as input indications of the task to be performed. For example, to induce the summarization behavior, the text "TL;DR:" is added after the text to be summarized.

GPT-2 has shown to outperform previous state-of-the-art models in several language modeling datasets (such as LAMBADA [Paperno et al., 2016], enwik8 and text8).[21] On reading comprehension tasks, the model proves to be comparable to supervised baselines in a zero-shot setting.

**GPT-3.** GPT-3 [Brown et al., 2020a] significantly increases the size of GPT models to 175B parameters, about two orders of magnitude more than the predecessor GPT-2. The pre-training approach is similar to GPT-2, but the data are larger and more heterogeneous, as they contain the CommonCrawl dataset and high-quality reference corpora, including an expanded version of WebText, internet-based books corpora and the English Wikipedia. GPT-3 is evaluated in zero-shot, one-shot and few-shot settings. In zero-shot setting, a description of the task to be performed is given to the model, whereas for one-shot and few-shot settings one or few examples of how the task is to be executed are provided to the model at inference time, after which the model can be prompted to execute the task on a new example.

GPT-3 is evaluated on several traditional language model benchmarks, e.g., LAMBADA, achieving better results compared to the previous fine-tuned state-of-the-art models in many cases. Results also show that scaling up the model greatly improves the performance in task-agnostic/few-shot setting, becoming even competitive with fine-tuned competitors in some NLP tasks. For example, on question-answering tasks, GPT-3 gets mixed results, with the zero-shot model outperforming the fine-tuned T5-11B on TriviaQA but being outperformed on NQ. In machine translation tasks, GPT-3 exceeds previous unsupervised competitors when translating to English, but underperforms when translating in the opposite direction. On SuperGLUE, the few-shot model obtains higher performance then the fine-tuned BERT$-large$ on four tasks and almost reaches the state-of-the-art at that time (a fine-tuned 11B-parameter model) in two tasks. Regarding text summarization, some critical issues on the GPT-3 samples are found, e.g., the presence of semantically document-level repetition and loss of coherence on long passages.

**ChatGPT.** Perhaps the world's most controversial AI language tool of our time, ChatGPT[22] is a GPT-based model trained to interact with humans in a conversational way. ChatGPT shares the underlying architecture with Instruct-GPT [Ouyang et al., 2022], which is specifically designed and fine-tuned for generating detailed instructions given a prompt. In particular, like InstructGPT, ChatGPT utilizes the Reinforcement Learning from Human Feedback (RLHF) technique, which combines supervised fine-tuning and reinforcement learning by exploiting the preferences of human trainers as reward signals in the training process. First, dialogue samples are obtained with the support of the human trainers, which suggest the desired output behavior for the given prompt and help the model formulate the responses in a conversational scenario. The model is, therefore, fine-tuned on a dataset obtained by mixing such samples with the InstructGPT data converted in a dialogue format. A reward model is then obtained through reinforcement learning on comparison data. The latter consists of prompt and relative model responses ranked by human trainers on the basis of response quality. The reward models are then employed to fine-tune the model using Proximal Policy Optimization, a reinforcement learning algorithm [Schulman et al., 2017]. The result is a model that allows for human dialogue on a wide variety of topics, answering follow-up questions while maintaining a conversation flow.

**GPT-NeoX-20B.** GPT-NeoX-20B [Black et al., 2022] is an autoregressive language model with 20B parameters and trained on Pile [Gao et al., 2021].[23] The GPT-NeoX-20B architecture largely follows GPT-3 with some differences. It applies the rotary positional embeddings [Su et al., 2021] on the first 25% of embedding vector dimensions, which consist in rotating the embedding space in such a way that the attention between two tokens is linearly dependent on their distance in the text. Moreover, GPT-NeoX-20B computes attention and feed-forward layers in parallel, whose outcomes are subsequently summed. The BPE tokenizer and the vocabulary size is similar to GPT-2, but the tokenizer is trained on the Pile dataset and, unlike GPT-2, handles the presence of prefix spaces and repeated space tokens.

---

[21] enwik8 and text8 are available at http://www.mattmahoney.net/dc/text.html

[22] https://openai.com/blog/chatgpt

[23] https://github.com/EleutherAI/gpt-neox

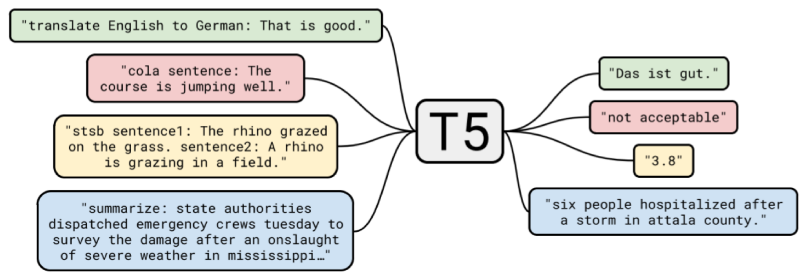

**Fig. 10.** T5 diagram [Raffel et al., 2020]. Every task is considered as text. A task-specific prefix is added to the input text to specify the current task.

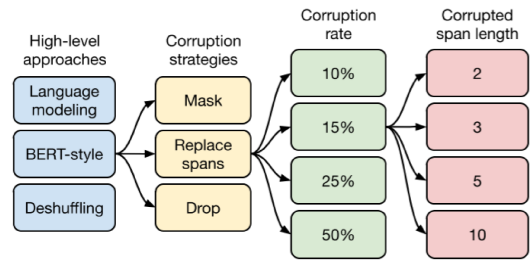

**Fig. 11.** A flowchart of the choices about unsupervised objective in T5 [Raffel et al., 2020]. After accurate experimentation, the authors opted for a BERT-style approach, with spans replacing as corruption strategies, 15% corruption rate and average corrupted span length chosen in {2,3,5,10}.

GPT-NeoX-20B is evaluated on several benchmarks (e.g., LAMBADA and TriviaQA), as well as mathematical and knowledge-based tasks. Results demonstrate that the model reaches higher performance than the similarly sized GPT-3 and fairseq models [Artetxe et al., 2022] when evaluated in five-shot setting.

**mGPT.** [Shliazhko et al., 2022] propose a GPT-like model in two size versions (1.3B and 12B parameters) for multilingual tasks.[24] mGPT aims to reproduce the architecture of GPT-3 starting from a GPT-2 implementation, based on the information described in [Brown et al., 2020a]. The resulting models are trained on 60 languages through Wikipedia and Colossal Clean Crawled (C4) corpora, and evaluated under zero-shot and few-shot settings (as in [Brown et al., 2020a]) on several multilingual benchmarks, including text classification, text generation and sequence labeling. On sequence labeling tasks, both zero-shot and few-shot settings show high scores. On most text classification tasks, the model is competitive with a state-of-art multilingual model XGLM (1.7B parameters) [Lin et al., 2022]. Overall, results show that larger models correspond to better generation abilities for all given languages. In addition, experimental evaluation on knowledge probing indicates a certain ability of mGPT to preserve factual knowledge.

### 2.3.3 Encoder-Decoder models

**T5.** Text-to-Text Transformer (T5) [Raffel et al., 2020] provides a unified approach in which one single model can be used for every task with the same objective, the same decoding process and the same training procedure.[25] To specify the current task, a task-specific prefix is added to the input text before feeding it to the model. The architecture of T5 is quite similar to the original Transformer [Vaswani et al., 2017], with few changes regarding the normalization layer and the position embedding scheme. The baseline is similar to BERT-*base* in terms of size and configuration, for both the encoder and the decoder, but with 220 million parameters in total, which are twice BERT-*base* parameters, as T5 contains two layer stacks instead of one. The text encoding method is SentencePiece [Kudo and Richardson, 2018] with a multilingual vocabulary of 32K wordpieces.

The training objective for pre-training and fine-tuning is a maximum likelihood objective using teacher forcing and a cross-entropy loss. More specifically, T5 utilizes a BERT-style denoising objective that randomly samples and drops out spans of tokens (loosely inspired by SpanBERT), with corruption rate of 15% of tokens and varying the span length, and the aim is to predict the dropped-out tokens. The span of tokens in replaced with only one special token. Figure 11 shows the experimentation choices for the pre-training objective.

---

[24] https://github.com/ai-forever/mgpt
[25] https://github.com/google-research/text-to-text-transfer-transformer

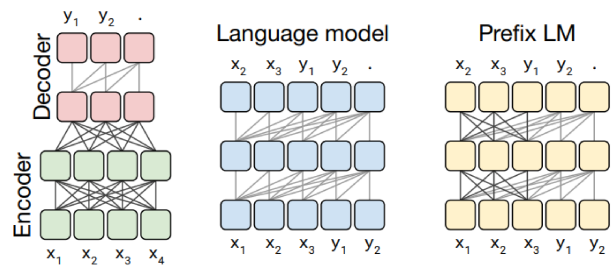

**Fig. 12.** Illustrations of the Transformer variants reviewed and compared in [Raffel et al., 2020]. On the left, the encoder-decoder architecture which uses the fully-visible masking when the encoder is involved, while the causal attention is used inside the decoder. In the middle, a standard language model which uses a causal masking. On the right, the Prefix LM which involves fully-visible masking over the input and causal masking for the output.

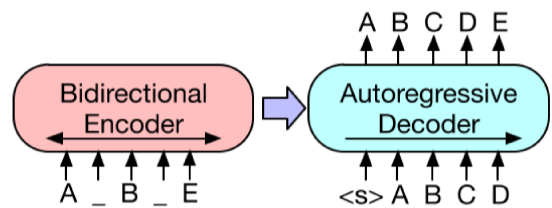

**Fig. 13.** BART architecture [Lewis et al., 2020].

For the model architecture, the authors reviewed and compared several Transformer variants (encoder-decoder, language model, prefix LM, see Figure 12) and eventually stated that the original encoder-decoder is the most suitable form for the text-to-text framework. In addition, they found that a small domain-specific unlabeled dataset, repeated many times in the pre-training, can degrade performance in some downstream tasks. For this reason they adopted a new massive dataset called Colossal Clean Crawled Corpus (C4), consisting in hundreds of gigabytes of clean, high-quality text picked from the web. Using C4 in the pre-training phase led the model to be flexible enough to be fine-tuned to a variety of tasks while obtaining state-of-the-art results. In this respect, T5 is provided in five versions: T5-*base* (i.e., the baseline, 12 layers, 12 attention heads, 768 hidden size, 220M parameters), T5-*small* (6 layers, 8 attention heads, 512 hidden size, 60M parameters), T5-*large* (24 layers, 16 attention heads, 1024 hidden size, 770M parameters), T5-*3B* (24 layers, 32 attention heads, 1024 hidden size, 3B parameters), T5-*11B* (24 layers, 128 attention heads, 1024 hidden size, 11B parameters). It should be noted that C4 is regarded as big enough to allow the exploration of the effect of scaling up the amount of pre-training without overfitting, in particular when training on more data, with larger versions of T5 or ensemble of models. However, small models can still be useful when limited computational resources are available in fine-tuning.

T5 has been tested on various downstream tasks, such as classification, question-answering, translation and summarization. When fine-tuning on GLUE (as well as on SuperGLUE), the datasets of all the benchmarks are concatenated so that the model is fine-tuned just once, using small batch size in order to mitigate overfitting in low-resource tasks. For task with long output sequences, the performance improved with the use of beam search, while the other task are reported with greedy decoding. T5 achieves state-of-the-art result in GLUE and overcomes RoBERTa and AlBERT in SuperGLUE and SQuAD.

A multilingual variant of the model, called *mT5*, is also available [Xue et al., 2021].[26] It was pre-trained on a Common Crawl dataset, called mC4, based on 101 languages. The architecture and training procedure is quite similar to T5. The model is available in four sizes: mT5-*small* (300M parameters), mT5-*base* (580M parameters), mT5-*large* (1.2B parameters), mT5-*xl* (3.7B parameters), mT5-*xxl* (13B parameters). It was tested on several XTREME [Hu et al., 2020] tasks, including XNLI and XQuAD [Artetxe et al., 2020]. For each task, three fine-tuning settings are considered: zero-shot, translate-train, in-language multitask. mT5 achieved strong performance on each task, overcoming previous models like m-BERT.

---

[26] https://github.com/google-research/multilingual-t5

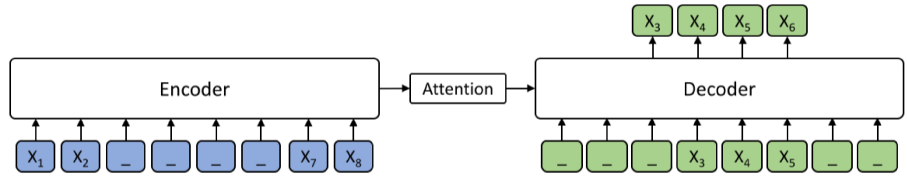

**Fig. 14.** MASS architecture [Song et al., 2019].

**BART.** BART [Lewis et al., 2020] is a denoising auto-encoder implemented as a sequence-to-sequence model, combining bidirectional and auto-regressive Transformers.[27] More precisely, it consists of a bidirectional encoder, following BERT, and a left-to-right auto-regressive decoder, following GPT (Fig. 13). The architecture follows the standard sequence-to-sequence Transformer, except for the activation function and the initialization of the parameters which are set up according to the GPT instructions. It is also related to BERT except for the decoder layers, in which there is an additional cross-attention with the final hidden layer of the encoder, and the absence of the additional feed-forward network before word prediction.

BART is available in a base size (6 layers, 12 attention heads,[28] 768 hidden size) and a large size (12 layers, 16 attention heads,[29] 1024 hidden size).

BART is pre-trained by first corrupting input text and then training the model to get back the original document. A key advantage of BART is the noising flexibility, i.e., it allows for arbitrary type of document corruption, including changing its length. This makes it possible to mask single tokens (like in the MLM task) as well as the entire input. The best results have been obtained shuffling the order of the original sentences and masking spans of text. This approach generalizes both MLM and NSP pre-training tasks. The masking method is similar to SpanBERT, but in this case the sampling is obtained using a different distribution (Poisson) and each span is replaced with a single [MASK] token.

BART can be fine-tuned for a wide range of downstream tasks, such as sequence classification, token classification, sequence generation and machine translation. Documents are tokenized with BPE, the corruption rate is the 30% of tokens for each document, and all sentences are permuted. BART obtains similar results to RoBERTa in GLUE and SQuAD benchmarks, and achieves new state of the art compared to previous models in abstractive dialogue, summarization and abstractive QA.

In [Liu et al., 2020], a multilingual version of BART is provided, called *mBART*. It is the application of BART to large-scale monolingual corpora across many languages. mBART represents the first multi-language denoising pre-training method for a complete sequence-to-sequence model, which makes it possible a direct fine-tuning for supervised and unsupervised machine translation without task-specific modifications. mBART recognizes the language through a language id token ($\langle LID \rangle$) placed at the beginning of the sentence. The authors built several versions of the model such as mBART25 (pre-trained on 25 languages) and mBART06 (pre-trained on 6 European languages).

**MASS.** MASS [Song et al., 2019] introduces a masked sequence-to-sequence pre-training, in which the encoder maps a sequence to several masked fragments, each composed of consecutive tokens, and the decoder predicts the masked fragments. The encoder and decoder are pre-trained simultaneously and in two stages: the encoder is forced to understand the meaning of the masked tokens to predict the tokens on the decoder side, while the decoder has the input tokens completely masked, so it relies primarily on the representation of the input passed by the encoder. More precisely, the decoder predicts the first token of the masked fragment based solely on the encoder input representation, while subsequent tokens are predicted based on the encoder representation and the previously predicted fragment tokens (Fig. 14). This enables the model to learn generating sequences during pre-training.

Available in a base size (6 layers, 12 attention heads, 768 hidden size) and a middle size (6 layers, 16 attention heads, 1024 hidden size), MASS is pre-trained on the WMT monolingual corpus,[30] and fine-tuned on several benchmarks for machine translation, text summarization, and conversational response generation.

**UniLM.** UniLM [Dong et al., 2019] is a Transformer model pre-trained on unidirectional, bidirectional and sequence-to-sequence prediction tasks.[31] It uses self-attention masks to control the context of the prediction, thus enabling a single architecture for uni/bidirectional and sequence-to-sequence approaches. The parameter sharing across the different language modeling objectives allows the model to learn more general text representations (Fig. 15). Different segment embeddings are used based on the pre-training task. Two special tokens, namely [SOS] and [EOS], are added at the beginning and at the end of each segment in the input, respectively. Such tokens play an important role for both natural language understanding and generation downstream tasks.

---

[27] https://github.com/pytorch/fairseq
[28] from https://huggingface.co/facebook/bart-base/blob/main/config.json
[29] from https://huggingface.co/facebook/bart-large/blob/main/config.json
[30] https://www.statmt.org/wmt16/translation-task.html
[31] https://github.com/microsoft/unilm

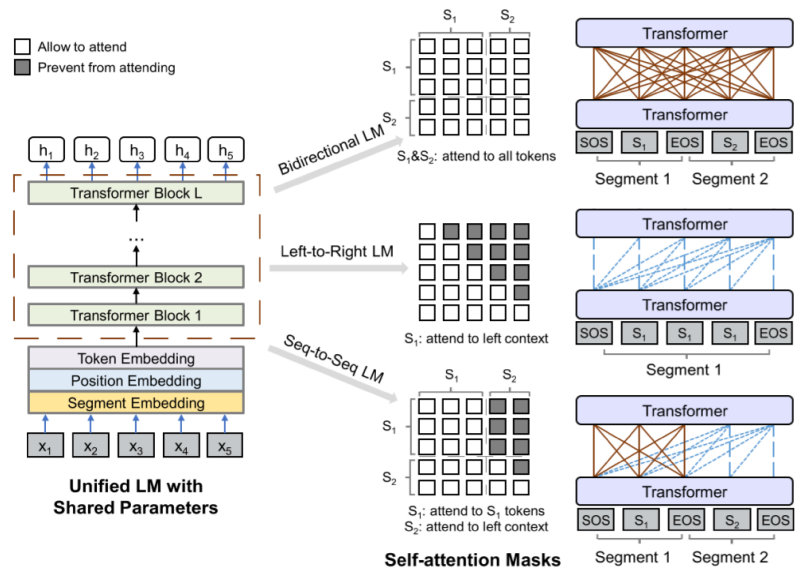

**Fig. 15.** UniLM Pre-training [Dong et al., 2019]. A unique model with shared parameters across all the pre-training objectives is employed. Self-attention masks control the context for the prediction, based on the objective.

UniLM is initialized and pre-trained following BERT-*large*. For language understanding tasks, it is fine-tuned as a bidirectional encoder, where the contextual [SOS] embedding represents the input encoding. For language generation tasks, it is fine-tuned on a sequence-to-sequence task, where the [EOF] token can used to learn when to stop the decoding process. UniLM is evaluated on several benchmarks, such as GLUE, SQuAD and CNN/DailyMail. Results demonstrate that UniLM is competitive against BERT on GLUE and better than BERT on SQuAD 2.0 development set, and outperforms previous state-of-the-art model on several generative tasks (CNN/DailyMail included).

Inspired by UniLM, [Bao et al., 2020] propose *UniLMv2*, which is pre-trained on bidirectional and sequence-to-sequence tasks in a unified manner with a pseudo-masked language modeling (PMLM) procedure. The model parameters and the context encodings are shared across the two tasks. Given a corrupted input with masked tokens, UniLMv2 uses standard masks and an autoencoding approach to learn the relations between masked tokens and the context. In addition, it uses pseudo masks and a partially autoregressive modeling approach to learn the relations between masked spans or tokens (token-to-token, token-to-span, and span-to-span relations). The accessible context for each token in the partially autoregressive modeling is controlled according to a factorization order. Similarly to [Yang et al., 2019], for each text in input a random factorization order is sampled. The model predicts one token or a span at each factorization step (for this reason it is partially autoregressive). The masks used in autoencoding pre-training provide global masking information, which can be used to access the position embeddings in the factorization steps. A standard mask, named [M], is employed to corrupt input tokens in the autoencoding pre-training. To handle factorization steps in the partially autoregressive pre-training, the [P] masks are appended to the input without replacing the original tokens. Such masks have the same position embedding of the corresponding tokens and their final embeddings are used for the predictions.

UniLMv2 size is similar to that of BERT-*base* (i.e., 12 layers with 12 attention heads and 768 hidden size). Fine-tuned and evaluated on several language understanding and generation tasks, UniLMv2 outperforms the base version of BERT, RoBERTa and XLNet on SQuAD and eight tasks of GLUE. On CNN/DailyMail, it proves to be more effective than other TLMs such as UniLM, MASS (base version) and BERTSUMABS (cf. Section 2.3.4).

### 2.3.4 Task-specific and long range models

**XLNet.** XLNet [Yang et al., 2019] exploits the advantages of autoregressive language modeling and autoencoding denoising. Instead of masking the input, XLNet enables bidirectionality by permuting the order of context factorization, and is trained to predict a token in an auto-regressive manner, but the prediction order is randomized and not sequential. This means that the model predicts a token based on previously predicted tokens, whose positions in the

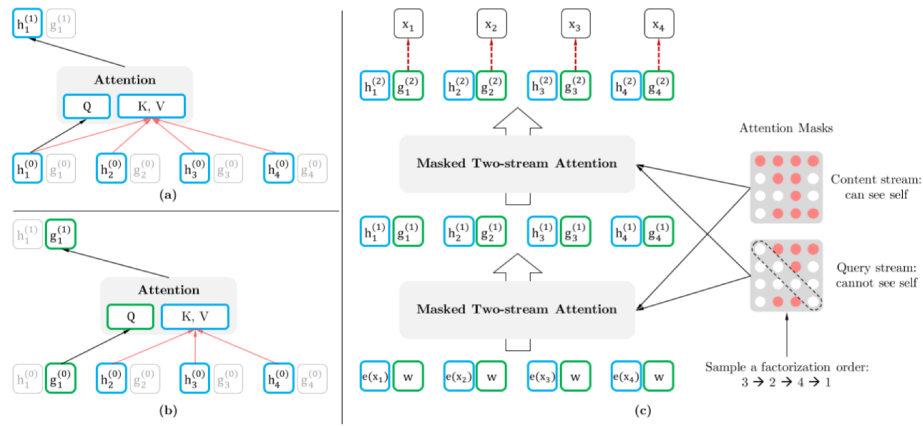

**Fig. 16.** XLNet (a) content stream attention, (b) query stream attention, and (c) two-stream self-attention [Yang et al., 2019].

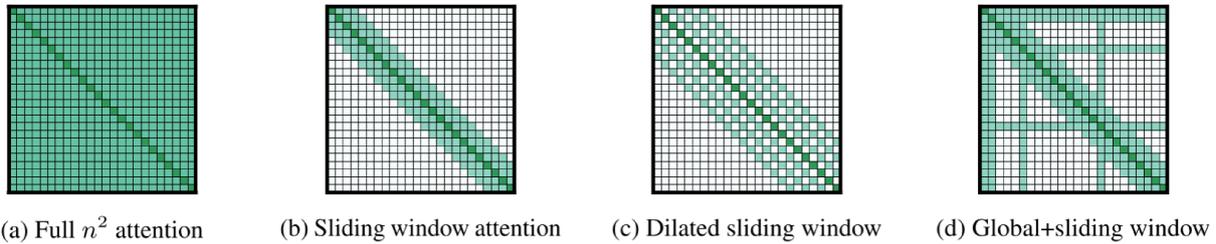

(a) Full $n^2$ attention          (b) Sliding window attention          (c) Dilated sliding window          (d) Global+sliding window

**Fig. 17.** Longformer attention pattern [Beltagy et al., 2020].

text may be before or after the token to be predicted. In this way, the model is forced to learn predicting the target with a randomly ordered context, thus enabling the bidirectionality. XLNet can be employed for tasks that include text of any length. The architecture is inspired by Transformer-XL [Dai et al., 2019], from which it takes the segment-level recurrence mechanism and relative positional encoding scheme. The segment-level recurrence mechanism consists in caching the representations computed for the previous segment so that they can be used as extended context for processing the next segment, thus increasing the maximum perceivable distance between dependencies. The relative positional encoding scheme adapts the input positional encodings to the segment-level recurrence mechanism. XLNet also introduces a new self-attention mechanism, called two-stream self-attention, which consists in a standard self-attention (content stream attention) and a query stream attention that mimics the masking behavior by obscuring the content of the token to be predicted but keeping its positional encoding visible. Figure 16 shows the difference between content stream attention and query stream attention: the former has access to the representations of both the context and the query token, while the latter has access only to the contents of previous tokens and to the positional encoding of the query.

XLNet is pre-trained with various pre-training corpora, which also include the BERT pre-training corpora. Two sizes of the model are available: XLNet-*large* (24 layers, 16 heads and 1024 hidden size) and XLNet-*base* (12 layers, 12 heads and 768 hidden size). XLNet achieves substantial improvements over BERT and RoBERTa on various benchmarks, such as RACE, SQuAD v2.0 and GLUE.

**Longformer.**    Longformer [Beltagy et al., 2020] is specifically designed to handle long text sequences.[32] Its attention mechanism combines local attention and global attention (Fig. 17), reducing the standard self-attention operations and scaling complexity from quadratic to linear with respect to the sequence length. Local attention is necessary to get contextual representations. In particular, the model uses a sliding window that can be dilated, i.e., it can have gaps of positions. For each attention head, different dilation configuration and different window size are provided. Different dilatation allows some heads to focus on local context through compact windows, while the others can focus on more distant contexts through enlarged windows. Global attention is added on few elected tokens and is crucial to get full sequence representations that are required for many NLP tasks (e.g., QA, document classification). How to select tokens for global attention is task-specific; for example it may consider the [CLS] token in classification tasks or all query tokens in question answering tasks. For auto-regressive tasks, the model has small non-dilated windows in lower layers, and increases the size and dilation (only on 2 heads) moving up to higher layers. The model is evaluated on the character-level text8 and enwik8 datasets.

---

[32] https://github.com/allenai/longformer

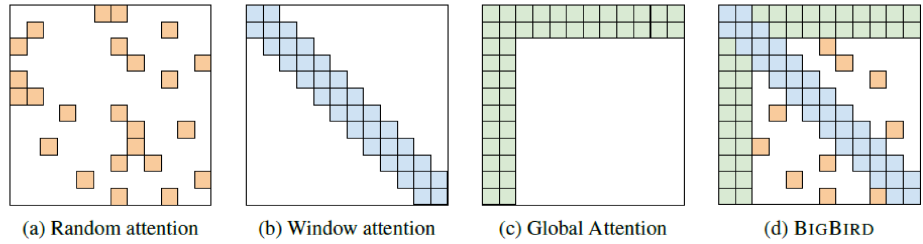

**Fig. 18.** BigBird attention pattern [Zaheer et al., 2020].

Longformer is pre-trained with MLM objective, starting from a RoBERTa checkpoint with some minimal non-architectural changes to add the new attention pattern. Besides English Wikipedia and Bookcorpus, the pre-training includes a portion of RealNews, consisting of documents longer than 1200 tokens, and a portion of Stories corpus. The tokenizer is the same as RoBERTa, as well as the vocabulary, except for the introduction of special tokens for question answering data. The sliding window is set with a size of 512 and extra position embeddings are added (up to 4096, while RoBERTa has a maximum of 512) to support longer documents. Thereby, the model can handle up to 8 times longer documents than BERT and RoBERTa. This leads Longformer to outperform RoBERTa in tasks where long-document datasets are used, such as WikiHop [Welbl et al., 2018] and TriviaQA for question-answering, OntoNotes [Pradhan et al., 2012] for co-reference resolution. Longformer is provided in two model sizes: Longformer-*base* (12 layers, 12 attention head, 768 hidden size) and Longformer-*large* (24 layers, 16 attention head, 1024 hidden size).

An encoder-decoder variant of Longformer, dubbed *Longformer-Encoder-Decoder* (LED), handles long text sequences in sequence-to-sequence tasks. The attention pattern of Longformer is applied to the Encoder part, while the Decoder part contains the full attention of standard Transformer. LED parameters are initialized following BART architecture in terms of number of layers and hidden sizes, with position embedding length extended to 16K tokens and enhanced initialization, in order to deal with longer texts. The model is released in two versions: LED-*base* (6 layers, 12 attention head, 768 hidden size) and LED-*large* (12 layers, 16 attention head, 1024 hidden size).

**BigBird.** BigBird [Zaheer et al., 2020] is a Transformer-based model that, like Longformer, can reduce the computational and memory cost by passing from a full-attention mechanism to a sparse-attention mechanism.[33] In particular, the model uses a combination of three attention patterns: random attention, sliding window attention and global attention (Fig. 18). In the random attention, each query focuses on a set of random keys. In the sliding window attention, each query focuses on a set of their right and left neighboring tokens selected by a sliding window. In the global attention, some tokens (from the text or special tokens like [CLS]) attend to the whole sequence. Formally, given an input sequence $\mathbf{X} = (\mathbf{x_1}, ..., \mathbf{x_n}) \in \mathbb{R}^{n \times d}$, the generalized attention mechanism for $\mathbf{x}_i$ is as follows:

$$Att(\mathbf{X})_i = \mathbf{x}_i + \sum_{h=i}^{H} softmax((x_i \mathbf{W}_Q^{(h)}) \cdot (\mathbf{X}_{N(i)} \mathbf{W}_K^{(h)})^{\top}) \cdot (\mathbf{X}_{N(i)} \mathbf{W}_V^{(h)}), \tag{7}$$

where $H$ is the number of attention's heads, $\mathbf{W}_Q^{(h)} \in \mathbb{R}^{d \times m}$, $\mathbf{W}_K^{(h)} \in \mathbb{R}^{d \times m}$ and $\mathbf{W}_V^{(h)} \in \mathbb{R}^{d \times d}$ are query, key and value matrices of the $h$-th head, respectively, $N(i)$ denotes the set of keys attended by the query $i$, while $x_i$ and $\mathbf{X}_{N(i)}$ are the vector representations of query $i$ and its keys, respectively. Like Longformer, this type of attention mechanism can reduce the Transformer complexity from quadratic to linear in the number of tokens, allowing the model to scale to much longer sequences in input, up to 8 times longer than previous Transformer-based models. The authors stated that sparse attention mechanisms have the same power and expressiveness as full-attention mechanisms, demonstrating that, on the one hand, when used in standalone encoders like BERT they are universal approximate methods of sequence-to-sequence functions and, on the other hand, that the sparse encoder-decoder models are Turing-complete under standard assumptions regarding precision.

In the pre-training settings, BigBird follows BERT and RoBERTa to create base and large models. It is pre-trained using the MLM objective and four corpora (Books, CC-News, Stories and Wikipedia), starting from a RoBERTa checkpoint. The model is available in two base versions: BigBird-ITC-*base* (in which only internal tokens in the text attend global attention) and BigBird-ETC-*base* (in which only the external token like [CLS] attends global attention). Both versions have 12 layers, 12 attention heads and hidden size of 768. Each version has its own setting regarding the number of tokens with global attention and random attention as well as the window size for local attention. Two large versions are also available: BigBird-ITC-*large* and BigBird-ETC-*large*, both with 24 layers, 16 attention heads and 1024 hidden size.

---

[33] https://github.com/google-research/bigbird

BigBird has been fine-tuned on QA challenging datasets (Natural Questions, HotpotQA-distractor, TriviaQA-wiki, WikiHop) outperforming competitors like Longformer and RoBERTa. In document classification and GLUE tasks, BigBird proves to be more advantageous than existing methods when longer documents and fewer training examples are being used (e.g., Arxiv [He et al., 2019]). Moreover, like Longformer, an encoder-decoder setup of BigBird is also proposed, in which the sparse attention mechanism is introduced only in the Encoder side.

**MonoT5.**  Inspired by the success of T5, [Nogueira et al., 2020] applied the model to document ranking, actually using a sequence-to-sequence model instead of the commonly adopted encoder-only pre-trained architectures.[34] Document ranking is considered as a relevance prediction task, i.e., given a document and a query, to assign a relevance score that indicates how much the document is relevant to the query. The new model, called *monoT5*, is fine-tuned to predict if the candidate document is relevant to the query, i.e., the target tokens are "true" and "false", and the underlying logits of the target tokens are considered as relevance probabilities for ranking. The core idea is to take advantage of the latent knowledge captured in the pre-training phase by connecting fine-tuned latent representations with output target tokens.

MonoT5 is fine-tuned on MS MARCO passage dataset [Nguyen et al., 2016] and then applied on other three datasets (Robust04 [Voorhees, 2004], Core17 [Allan et al., 2017] and Core18 [Naseri et al., 2018]), experimenting with zero-shot document ranking. Three sizes of T5 are considered in the fine-tuning process: T5-*base*, T5-*large* and T5-*3B*. Due to the high computational costs, it was not possible to fine-tune T5-11*B*. T5 is used as a re-ranker, applied directly to the output of BM25 model (and variants). In the experimentation, the zero-shot transfer-learning approach on the three dataset outperforms the in-domain cross-validation approach used in previous models. Furthermore, the study reveals that, in data-poor conditions, the model outperforms the classic encoder-based approaches.

**DPR.**  Dense Passage Retriever (DPR) [Karpukhin et al., 2020] is based on two BERT-*base* encoders and focuses on text retrieval in open-domain question-answering, where the task is to select query-relevant passages from candidate contexts.[35] The first encoder maps the passage to vector in a low-dimensional and continuous representation space of a given size, and builds an index for all the passages. The second encoder maps the query to a real-valued vector, with the same fixed size as the passage vectors, and retrieves from the index the passages whose vectors are the closest to the query vector in terms of inner-product similarity.

In general, dense vector representation needs multiple query-passage labeled pairs. In DPR, the focus is to find a proper training scheme that can allow the use of a relative small number of query-passage pairs. For this purpose, the embeddings are built so that the inner product between query and relevant passages is maximized by using a objective that compares all pairs in a batch. The embeddings correspond to the representation of the [CLS] tokens.

DPR is trained with the English Wikipedia corpus, where each article is split into passages and each passage is preceded by the article's title along with the [SEP] token. Three strategies of positive/negative sampling have been considered: random passages from the text corpus, positive passages of other questions (named *gold* passages) and top BM25 passages that do not contain the answer. Best results corresponded to the use of gold passages in the same mini-batch and one BM25 passage.

DPR has been tested on five question-answering datasets: Natural Questions [Kwiatkowski et al., 2019], TriviaQA [Joshi et al., 2017], WebQuestions [Berant et al., 2013], CuratedTREC [Baudis and Sedivý, 2015] and SQuAD v1.1. In TriviaQA, WebQuestions and CuratedTREC there are only query-answer pairs, so that DPR uses the highest BM25 passage containing the answer as positive passage. In SQuAD and Natural Questions, the passages are managed differently from the DPR's pool of candidates, so that a matching and replacing process of the gold passages with the corresponding passage in the pool is performed. DPR was also trained on each specific dataset as well as on all datasets, except SQuAD, with the latter demonstrating to perform better in retrieving top-20 and top-100 passages than single-dataset DPR.

**BERTSUM.**  BERTSUM [Liu and Lapata, 2019a] is a BERT-based approach for extractive and abstractive summarization.[36] The base architecture consists of a sentence-level BERT encoder. Differently from the original BERT,

---

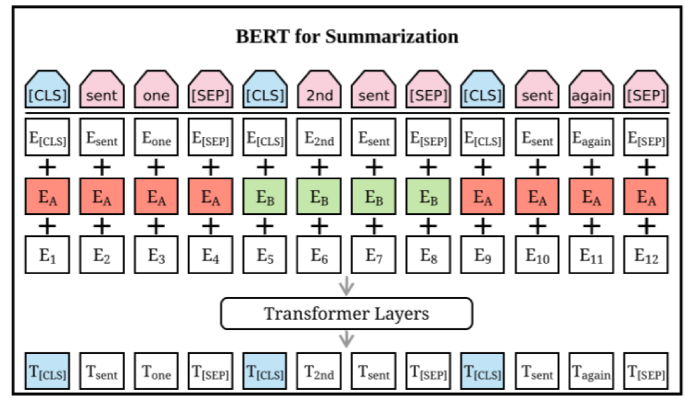

**Fig. 19.** BERTSUM architecture [Liu and Lapata, 2019a].

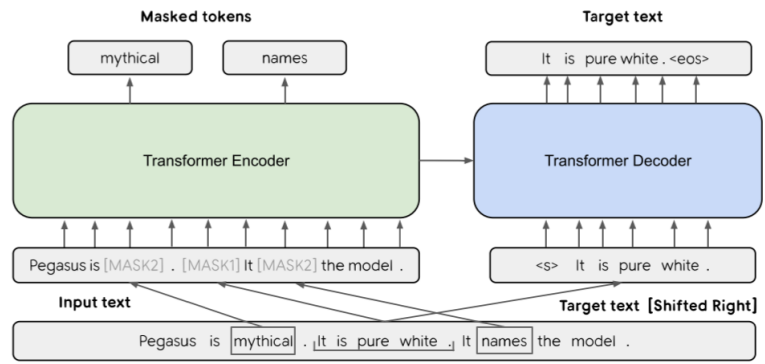

**Fig. 20.** PEGASUS architecture [Zhang et al., 2020]. [MASK1] refers to sentence masking of the Gap Sentences Generation objective, while [MASK2] refers to the token masking of MLM.

BERTSUM introduces a [CLS] token at the beginning of each input sentence (Fig. 19). Two interspersed segment embeddings are used to distinguish multiple sentences in the document (i.e., sentences in even positions are assigned one segment embedding, while those in odd positions are assigned the other segment embedding). Moreover, the input length limit of 512 tokens is removed by adding more position embeddings, randomly initialized and to be fine-tuned with the other parameters. The resulting contextual [CLS] embedding from the top layer of the encoder is representative of the sentences. The encoder is implemented with `bert-base-uncased`.

To address the extractive summarization task, two stacking Transformer layers are put on top of the encoder to learn document-level features from the sentence representations in a hierarchical manner. In particular, the lower Transformer layers capture adjacent sentences information, while the higher layers represent the multi-sentence discourse. The final sentence representations are fed to a classification layer to decide what sentences include into the summary. The resulting model is called *BERTSUMEXT*.

For abstractive summarization, a 6-layered Transformer decoder is added to the architecture. Since there is knowledge disalignment between encoder and decoder (the former is pre-trained while the latter needs to be trained from scratch), [Liu and Lapata, 2019a] propose to separate the optimizers, learning rates and warmup-steps of the two modules in order to train the encoder with more accurate gradients when the decoder becomes stable. Additionally, a two-stage fine-tuning strategy is proposed, which consists in first fine-tuning the encoder on the extractive task and then fine-tuning it on the abstractive task. The purely abstractive model is named *BERTSUMABS*, while the two-stage fine-tuned model is *BERTSUMEXTABS*.

The models are fine-tuned and evaluated on three summarization benchmarks, containing news articles and related summaries: *CNN/DailyMail* [Hermann et al., 2015, Nallapati et al., 2016, See et al., 2017], *NYT*[37] and *XSUM* [Narayan et al., 2018].

---

[37] `https://catalog.ldc.upenn.edu/LDC2008T19`

**PEGASUS.** PEGASUS [Zhang et al., 2020] is a TLM specifically conceived for abstractive summarization. It introduces a new self-supervised pre-training objective called *Gap Sentences Generation* (GSG), which consists in masking relevant sentences from the input documents and inducing the model to generate the masked sentences. Each selected sentence is replaced with a mask token. The concatenation of such gap-sentences becomes a pseudo-summary of the document. The importance of a sentence is deduced calculating the ROUGE F1 score between the sentence and the remainder of the document.

PEGASUS is pre-trained using the C4 corpus and *HugeNews*, a dataset collected by the authors, consisting of 1.5B news articles. The pre-training phase is conducted with or without the addition of the MLM objective (Fig.20). PEGASUS is provided in two sizes: PEGASUS-*base* (12 layers, 12 attention heads, 768 hidden size) and PEGASUS-*large* (16 layers, 16 attention heads, 1024 hidden size).

PEGASUS is evaluated on several summarization datasets, such as CNN/DailyMail and XSum. The best performing setting for the base model reveals to be when using the GSG objective without MLM, selecting the top-m sentences whose importance scores are calculated independently, considering the double counting of identical n-grams in the computation of ROUGE1-F1 and a SentencePiece unigram vocabulary of 96K tokens. The same settings are then used for PEGASUS-*large*. Results show that the model achieves state-of-the-art performance on all the considered downstream task datasets. Moreover, on CNN/DailyMail it reaches better ROUGE2-F1 score than GPT-2 in zero-shot setting and, using just 1K examples, it outperforms the previous best few-shot learning model.

**PRIMERA.** PRIMERA [Xiao et al., 2022] is designed for multi-document summarization. To this purpose, it incorporates a particular pre-training strategy, called Entity Pyramid, which selects salient sentences from a cluster of related documents. Such sentences are then masked and the model is trained to reconstruct and aggregate them using the remainder of the documents. The model concatenates the documents and adds separator tokens among them to keep the boundary information. The documents are processed with LED, which is pre-trained on an unlabeled multi-document dataset. PRIMERA is evaluated on several multi-document summarization datasets, showing to be effective in zero- and few-shot settings.

# 3 Problems and Tasks

Transformer-based language models are leading a significant advance in AI-based NLP research to bring in better support for human decision-making processes in the legal domain. In this section, we present the main types of legal problems that are recognized as those particularly benefiting from AI-based NLP research, and we discuss the associated tasks that are being powered by BERT and related models. We organize our discussion on the legal problems into three broad areas, namely *search* (Section 3.1), *review* (Section 3.2), and *prediction* (Section 3.3) — through our discussion, we attempt to organize the flow of presentation by distinguishing tasks involving *codes* (e.g., statutes, regulations, contracts) from those concerning *case law*; however, it is often the case that a task can be regarded as relevant for any type of legal document. Please note that the three macro categories are actually interleaved and interrelated in many practical scenarios, therefore our purpose of classification should be taken with a grain of salt, mainly for the sake of presentation. Throughout the remainder of the paper, we will use abbreviations whose description is reported in Table 2.

Moreover, to complement our presentation supporting the description of the TLM-based methods, we end this section with an overview of relevant benchmarks for the TLM-based legal learning context (Section 3.4).

## 3.1 Legal Search

Legal search corresponds to a need for legal information, and hence requires the detection and retrieval of documents potentially relevant to support legal decision-making. For instance, lawyers may search for laws enacted by parliaments or civil codes (similar to legislation in a civil law jurisdiction), but also for documents in litigation, patents, and several other documents that can support a law firm [Locke and Zuccon, 2022].

The searched documents are also called legal authorities in [Dadgostari et al., 2021], which points out how the legal search is driven by a notion of *relevance*, which should be «determined functionally with respect to norms and practices concerning legal reasoning and argumentation within a legal community». Thus, a document is regarded as «legally relevant exactly when it is understood by the dominant legal community as containing information that bears on a legal question of concern» [Dadgostari et al., 2021].

Legal search has been addressed in [Dadgostari et al., 2021] as a citation recommendation problem: given a citation-free legal text (CFLT), to find the most suitable set of opinions, from a reference legal corpus, to be cited for the input CFLT. Then, if the CFLTs are opinions from the corpus where all citation information is deleted, the search results can be compared to the actual citation information. More in general, legal search tasks are mainly addressed from two perspectives, namely Information Retrieval and Textual Entailment. While the former is intuitively seen as an essential part of any legal search task, the latter actually corresponds to Natural Language Inference, since it aims to determine, given any two textual fragments (e.g., two sentences), whether the one can be inferred from the other one; the entailment is said "positive", resp. "negative" when the first text can be used to prove that the second text is true, resp. false, otherwise (i.e., if the two texts have no correlation) the entailment is regarded as "neutral" [Kim et al., 2021].

**Table 2.** Abbreviations and descriptions of most relevant tasks in this article.

| abbreviation | description | abbreviation | description |
|---|---|---|---|
| AS/ES | abstractive/extractive summarization | NLI | natural language inference |
| AVP/ALVP | article/alleged violation prediction | NSP | next sentence prediction |
| CIR | case importance regression | OR | overruling |
| CJP(E) | court judgment prediction (and explanation) | PR/CR | passage/case retrieval |
| CLM | causal language modeling | QA; MCQA | question answering; multiple choice QA |
| CTR | case term recognition | RC; MCRC | reading comprehension; multiple choice RC |
| DR | document retrieval/recommendation | RIR | regulatory information retrieval |
| DS/SS | document/sentence similarity | RRL | rhetorical role labeling |
| IE | information extraction | SA | sentiment analysis |
| IR | information retrieval | SAR | statutory article retrieval |
| LPP | legal precedent prediction/retrieval | SF | slot filling |
| LJP | legal judgment prediction | STP | same topic prediction |
| MLM | masked language modeling | TC/SC; TpC | text/sentence classification; topic classification |
| NER | named entity recognition | TM/CM | text/case matching |

Since 2014, the Competition on Legal Information Extraction/Entailment (COLIEE) has served as an international forum to discuss issues related to legal information retrieval and entailment.[38] The COLIEE editions from 2014 to 2017 focus on a two-phase legal question answering task: given a legal bar exam question $q$, the first phase is to retrieve a set of articles from a target Civil Code corpus (i.e., Japanese civil code) that are deemed as appropriate for answering $q$, whereas the second phase is to determine if the (gold) relevant articles entail $q$ or *not* $q$. Since the 2018 edition, both the retrieval and entailment tasks are also applied to *case law* texts, which are relatively long documents consisting of the facts (i.e., factual statements) in a case. Searching for case law is a peculiarity of common-law jurisdictions, which comes about from the principle of "stare decisis" (doctrine of precedent), and has unique challenges that have emerged in law research [Locke and Zuccon, 2022]; conversely, for the civil-law jurisdictions, statutes are applied in the decision-making for a given legal issue in a *mutatis mutandis* approach, i.e., when asserting the substantial identity of two facts, we want to ignore the circumstances of a contingent nature, which are naturally different. The most recent edition at the time of writing of this article, COLIEE-2021 [Rabelo et al., 2022], proposes five tasks:

- Legal Case Retrieval (Task 1) – the goal is to identify the cases from a court case corpus that support the decision of a query case; such cases are also called "noticed" with respect to the query case, i.e., precedent cases that are referenced by the query case. Formally, given a set of candidate cases $C = \{c_1, \ldots, c_n\}$ and a query case $q$, the task is to identify the supporting cases $C_q = \{c \mid c \in C \land noticed(q, c)\}$, where $noticed(q, c)$ denotes that $c$ should be noticed given the query case $q$.
- Legal Case Entailment (Task 2) – given a query case, the goal is to identify one or more paragraphs from a case relevant to the query that entail(s) the decision of the query. Formally, given a query case $q$ and a case $c_i$ relevant for $q$ represented by its paragraphs $\{c_{i1}, \ldots, c_{in_i}\}$, the task is to identify the set of paragraphs $\{c_{ij} \mid c_{ij} \in c_i \land entails(c_{ij}, q)\}$, where $entails(c_{ij}, q)$ is true if the paragraph $c_{ij}$ entails $q$.
- Statute Law Retrieval (Task 3) – this is the former phase-1 in COLIEE-2014, i.e., given a civil code $S$ and a legal bar exam question $q$, to retrieve the set of articles $S_q$ from $S$ such that $entails(S_q, q)$ or $entails(S_q, not\ q)$.
- Statute Law Entailment (Task 4) – this is the former phase-2 in COLIEE-2014, i.e., given a legal bar exam question $q$ and relevant articles $S_q$, to determine if it holds that $entails(S_q, q)$ or $entails(S_q, not\ q)$.
- Legal Question Answering (Task 5) – this is regarded as a combination of Task 3 and 4 (although, in the COLIEE competition, any knowledge source other than the results of Task 3 can be used).

Training data are pairs ⟨*query, noticed case(s)*⟩ for Task 1, triplets ⟨*query, noticed case(s), entailing paragraph IDs of the case(s)*⟩ for Task 2, pairs ⟨*query, relevant article(s)*⟩ for Task 3, triplets ⟨*query, relevant article(s), Y/N answer*⟩ for Task 4 and Task 5; the test data are only queries for Tasks 1, 3, and 5, whereas they include queries and relevant texts for Tasks 2 and 4.

It is worth noticing that supporting cases are relevant factors in court decision-making and are actually used in the attorney's ligation [Nguyen et al., 2021a]. Legal case entailment is also useful in practice, since a decision for a new case can be predicted by implication of previous cases; it can also be treated in combination with case retrieval,

---

[38] webdocs.cs.ualberta.ca/~miyoung2/jurisin_task/index.html; webdocs.cs.ualberta.ca/~miyoung2/COLIEE201$i$ ($i \in \{5, 6, 7\}$); sites.ualberta.ca/~miyoung2/COLIEE2018/; sites.ualberta.ca/~rabelo/COLIEE20$i$ ($i \in \{19, 20, 21\}$)

as developed in [Vuong et al., 2023], where a supporting model is introduced to describe the case–case supporting relations and to define paragraph–paragraph and decision-paragraph matching strategies. Analogous considerations hold for the statute law tasks. Moreover, the latter are particularly challenging since, besides the need for addressing the long articles, legal bar exam questions describe specific legal cases, while the language used in statute law tends to be more general.

A further perspective on legal question answering is taken in [Zheng et al., 2021], where a *multiple choice question answering* task, dubbed CaseHOLD, is defined from legal citations in judicial rulings. The citing context from the judicial decision serves as the prompt for the question, whereas the answer choices are holding statements derived from citations following text in a legal decision. Holdings are central to the common law system, as they represent the predominating, precedential legal rule when the law is applied to a particular set of facts. Analogously, in [Xiao et al., 2021], legal question answering is addressed on the JEC-QA dataset, which consists of multiple-choice questions from the Chinese national bar exam, where the questions and candidate choices are concatenated together to form the inputs of the models.

### 3.2 Legal Document Review

Document review is another critical process for law practitioners and lawyers, as it usually involves document sets that are unmanageable for a team of humans given their amount, the cost of reviewers, and deadlines in the context of legal proceedings. The purpose of legal document review is for the parties to a case to organize and analyze the available documents so as to determine which are sensitive or otherwise relevant to the litigation. For instance, document review can be intended to negotiate or revise an agreement, ensure that the filings of an attorney's client comply with appropriate regulations, modify a brief for a trial motion, inspect a contract to avoid potential risks, or review client tax documents. Relevance, responsiveness to a discovery request, privilege, and confidentiality are essential criteria for any document in the review, but also in the analysis of the information to relate key documents to alleged facts or key legal issues in the case.

[Shaghaghian et al., 2020] recognizes four main tasks of document review, namely information, fact, comparative, and rule navigation, which are primarily characterized in terms of the following problems:

— *Passage retrieval* – Navigating a user to answers for non-factoid questions is in fact seen as equivalent to retrieving relevant text passages during the document review process. Passage Retrieval to answer non-factoid questions can be modeled as a binary text classification task, i.e., given a set of queries $\{q_i\}_{i=1..Q}$ and a set of candidate texts (e.g., sentences, snippets, paragraphs) $\{s_j\}_{j=1..N}$, each pair question-snippet $(q_i, s_j)$ is assigned label 1 if $s_j$ contains the answer to $q_i$, and label 0 otherwise.

— *Named entity recognition* – Examining factoid questions whereby the user is searching for specific facts or entities is instead modeled as named entity recognition (e.g., extraction of facts from a court decision document, such as Date of Argument, Date of Decision, Petitioner, Judge, Sought Damages and Damages Awarded Monetary Values). Named Entity Recognition to extract facts or elements of factoid questions can be modeled as a sequence labeling, multi-class classification task, i.e., given a set of fact-related classes $\{c_i\}_{i=1..C}$, each token is assigned a class (or a distribution over the classes).

— *Text similarity* – Computing text similarity is essential to identify matching texts according to different aspects; for instance, to identify the differences between a regulation and its amended version, or to discover the discrepancies of regulations in different jurisdictions. Text similarity to identify matching texts at various, pre-determined levels can be modeled as a binary, resp. multi-class, text classification task, i.e., given a set of matching levels $\{m_i\}_{m=1..M}$ and a set of texts $\{s_j\}_{j=1..N}$, each pair of texts $(s_j, s_k)$ is assigned a class $m_i$ depending on the degree of matching between $s_j$ and $s_k$.

— *Sentiment analysis* – This can be addressed to identify the polarity, or the mood, associate with certain legal statements, with the purpose of, e.g., identifying rules imposed by deontic modalities, which are of the form of obligations, prohibition and permission statements. This can be modeled as a binary, resp. multi-class, text classification task, i.e., given a set of texts $\{s_j\}_{j=1..N}$, each text is assigned a class depending on the polarity or sentiment expressed in the text.

In [Xiao et al., 2021], *legal reading comprehension* is addressed to predict the start positions and end positions given question-answer pairs with corresponding supporting sentences. Legal document review is also related to *document recommendation*. As discussed in [Ostendorff et al., 2021], a typical recommendation scenario occurs during the preparation of a litigation strategy, when the involved legal professionals are provided with recommended other decisions that possibly cover the same topic or provides essential background information (e.g., they overrule the target decision). Also, *text segmentation*, i.e., the task of dividing a document into multi-paragraph discourse units that are topically coherent, can be useful to one or more of the above tasks, especially when the existing logical boundaries imposed to the document might not be sufficient to detect fine-grain topic changes. In [Aumiller et al., 2021], text segmentation is used to solve a *topical change detection* problem (also called same topic prediction): Given two chunks of text of the same type (e.g., paragraphs, sections) and binary labels, to determine if the two chunks belong to the same topic, otherwise a change in topic is detected and so the beginning of a new chunk of text. Also, [Savelka et al., 2021] introduce the task of *automatic functional segmentation*, which is to segment adjudicatory decisions of cases according to the functional role of the parts.

*Contracts*, in various forms, are major target of interest for document review tasks. [Zheng et al., 2021] focus on contract documents such as *Terms-of-Services*, for the detection of potentially unfair contractual terms. A contractual term (clause) is regarded as unfair if it has not been individually negotiated, and it corresponds to an evident imbalance in the rights and obligations of the parties, to the detriment of the consumer [Zheng et al., 2021]. A binary classification task can hence be defined, whereby positive examples are the potentially unfair contractual terms. The Terms-of-Service task can help consumers better understand the terms they agree to when signing a contract and ease access to legal advice about unfair contracts. [Hendrycks et al., 2021] address the *legal contract review* task, which is to analyze a contract to understand rights and obligations of the signatories as well as to evaluate the associated impact. This task can be seen as similar to extractive question answering, where each question is the description of a label category and language models have to detect the spans of the contract is related to the label. [Leivaditi et al., 2020] specialize the legal contract review task to lease agreements and address this task from two perspectives: detection of sentences expressing a potential risk to one or more signatories (binary classification) and extraction on important entities for the domain (entity recognition). Unlike [Hendrycks et al., 2021] and [Leivaditi et al., 2020], which aim to find what kinds of terms are present, [Koreeda and Manning, 2021] focus on knowing what exactly each of these terms states. Given a set of hypotheses and a contract, the task is to decide if the contract entails, contradicts or is neutral to each hypothesis (three-class classification) and detect the evidence, i.e., spans, in the contract that determine the decision (multi-label binary classification). On privacy policies, [Ahmad et al., 2021] define the *intent classification* task, which is to predict sentences explaining privacy practices, along with a slot filling task to detect text spans within a sentence expressing specific details. A slot extraction task is also performed in [Bui et al., 2021] to detect spans in the text expressing different types of user data.

Another legal context that can be included in the document review category to a broader extent concerns a special case of retrieval task, namely *regulatory information retrieval* [Chalkidis et al., 2021b], i.e., to ensure a regulatory compliance regime regarding an organization's processes/controls. A compliance regime includes corrective, detective and preventive measures such that either, given a control/process, to retrieve relevant laws in order to apply corrective measures or, given a new law, to retrieve all the affected controls/processes in order to apply corrective or preventive measures. Regulatory information retrieval is defined as a special case of document-to-document information retrieval, since the query is an entire document — unlike traditional information retrieval, whereby queries are usually short texts.

More tasks concern case law documents. Legal cases are lengthy and unstructured, although they are actually characterized by an implicit thematic structure into sections such as "facts of the case", "arguments given by the parties", etc. These sections are often called as *rhetorical roles*. Identifying such semantic roles is essential for improving the roles readability of the documents but also helps in downstream tasks such as classification and summarization. The task is challenging since in most cases legal documents can vary in structure and rhetorical labels can be subjective. [Bhattacharya et al., 2019b] introduce the *rhetorical role labeling* task, which is to label sentences of a legal case with the corresponding rhetorical role. This task was also introduced in the context of the Artificial Intelligence for Legal Assistance (AILA) 2020 competition (Task 2), whereby the predefined labels are "Facts", "Ruling by Lower Court", "Argument", "Statute cited", "Precedent cited", "Ratio of the decision", and "Ruling by Present Court".[39]

To support legal document review, special cases of retrieval are also involved. For instance, [Martino et al., 2022] deal with the identification of *paragraph regularities* in legal cases, which is addressed by using a nearest-neighbor search method to efficiently select the most similar paragraphs appearing in a set of reference documents. *Explanatory sentence retrieval* [Savelka and Ashley, 2021] is instead to retrieve useful sentences to explain predetermined legal concepts. Explanations of legal concepts can be inferred looking at how they have been applied in previous cases, allowing a lawyer to elaborate supporting or contrary arguments related to particular accounts of meaning. Searching through legal documents, a lawyer can find sentences mentioning a particular concept, but not all of them could be useful for explaining that concept. Therefore, the aim is to automatically rank sentences in order to assign higher scores to explanatory sentences.

It is also highly desirable for legal professionals dealing with cases to access to their summaries, also known as *headnotes*. However, creating headnotes is certainly time-consuming, therefore automatic *summarization of legal judgments* is another meaningful problem in the legal domain. Two related tasks have been introduced in the Artificial Intelligence for Legal Assistance (AILA) 2021 competition, namely to identify "summary-worthy" sentences in a court judgment (Task 2a) and to generate a summary from a court judgment (Task 2b).[40] The former can be seen as a sentence classification task, whereas the latter can be addressed either by collecting the detected summary-worthy sentences so as to form *extractive summaries* or by using generative models to produce *abstractive summaries*.

---

[39] https://sites.google.com/view/aila-2020/task-2-rhetorical-role-labeling-for-legal-judgements

[40] https://sites.google.com/view/aila-2021/task-2-summarization-of-legal-judgements

### 3.3 Legal Outcome Prediction

Legal relevance is related to the well-known predictive theory of the law first introduced in [Oliver Wendell Holmes, 1897]. In contrast to previous definitions of the law, Holmes formulated the law as a prediction, particularly the behavior of a court, so as to build a more useful approach in practice when dealing with those individuals who care little for ethics or lofty conceptions of natural law (i.e., the "bad men"). Besides the Holmes' theory, predictive tasks in law are more generally concerned with judicial opinions. For instance, as discussed in [Dadgostari et al., 2021], given the content of a source judicial opinion, one task is to predict the other opinions that are cited in the source document; or, given a source document and a set of related opinions identified by law professionals, to predict their answers.

The primary predictive task in law is commonly referred to as *legal judgment prediction* (LJP), i.e., to predict the outcome of a judicial decision based on the relevant facts and laws [Aletras et al., 2016, Zhong et al., 2018, Chalkidis et al., 2019a]. For instance, [Aletras et al., 2016] define the problem of case prediction as a binary classification task, which is to predict whether one of a predetermined, small set of articles of the ECtHR Convention has been violated, given textual description of a case, which includes the facts, the relevant applicable law and the legal arguments. [41] In [Xiao et al., 2021], the LJP task is addressed on both criminal and civil cases from the CAIL-Long dataset. Fact descriptions are taken as input whereas the judgment annotations are extracted via regular expressions; each criminal case is annotated with the charges, the relevant laws, and the term of penalty, and each civil case is annotated with the causes of actions and the relevant laws. For criminal cases, the charge prediction and the relevant law prediction are formalized as multi-label classification tasks, whereas the term of penalty prediction task is formalized as a regression task. For civil cases, the cause of actions prediction is formalized as a single-label classification task, and the relevant law prediction is formalized as a multi-label classification task. In [Dong and Niu, 2021], the three types of prediction are addressed in a context of graph node classification, where a Transformer model is combined with a graph neural network model. [Malik et al., 2021] propose the *court judgment prediction and explanation (CJPE)* task, which requires to predict the decision of a case and to provide explanations for the final decision, where explanations correspond to portions in the case description that best justify the outcome.

A related axis of prediction is the one introduced in [Mahari, 2021], dubbed as *legal precedent prediction*, which is to predict passages of precedential court decisions that are relevant to a given legal argument posed in the context of a judicial opinion or a legal brief. Both judicial opinions and legal briefs usually contain a number of independent legal arguments, each citing its own set of precedent, where the precedent depends on the context of the entire case as well as on the specific legal argument being made [Mahari, 2021]. Clearly, in common law jurisdictions, this is particularly useful as legal professionals build their arguments by drawing on judicial precedent from prior opinions.

Another critical task is *overruling prediction*, i.e., to determine if a statement is an overruling, i.e., a sentence that nullifies a previous case decision as a precedent, by a constitutionally valid statute or a decision by the same or higher ranking court (which establishes a different rule on the point of law involved). In [Zheng et al., 2021, Limsopatham, 2021], the overruling prediction is modeled as a binary classification task, where positive examples are overruling sentences and negative examples are nonoverruling sentences from the law. The overruling task is clearly important for legal professionals, since verifying whether cases remain valid and have not been overruled is essential to ensuring the validity of legal arguments.

*Case importance* and *article violation* are also considered [Chalkidis et al., 2019a], [Limsopatham, 2021]. Predicting the importance of a case can be seen as a regression task, e.g., to measure on a scale from lower scores for key cases, to higher scores for unimportant cases. Given the facts of a case, article violation is to predict if any human rights article or protocol has been violated by the case (binary classification), or which human rights articles and/or protocols have been violated (if any) by the case (multi-label classification). A special case of the above task is the *alleged violation prediction* introduced in [Chalkidis et al., 2021c], whose aim is to predict the allegations made by applicants given the facts of each case. This can be useful to identify alleged violations for plaintiffs, facts supporting alleged violations for judges but also for legal experts to identify previous cases related to the allegations. The task is treated in [Chalkidis et al., 2021c] as a multi-label text classification, since the model might select multiple articles that were allegedly violated (according to the applicants).

*Employment notice prediction* [Lam et al., 2020] is to predict the number of months awarded for *reasonable notices* in employment termination cases. If the employer does not comply with the obligation to provide an appropriate employment notice or payment in lieu of notice, judges determine the compensation that an employer owes to an employee at the time of termination. Courts might rely on factors such as length of service, employee's age, character of employment, aggravated damages to establish what constitutes reasonable notices, but it is not clear how to weigh each individual factor and how they should be used. As a result, the case law on employment notice turns out to be inherently inconsistent and subjective. [Lam et al., 2020] define this problem as a text classification task, in order to obtain a similar decision-making process of a judge who would rely allegedly on past cases and differences of fact to decide the amount of reasonable notice.

---

[41] The above view has been recognized not only as one of the most challenging by the legal community, but it has also raised controversial debate on the role of AI applied to law. Adversarial opinion is on that based on the evidence that, in real-life scenarios, judges are unlikely to defer to AI to decide the outcome of a case. Nonetheless, the authors adopt an opinion that is commonly shared with most researchers and practitioners of AI in law, whereby it should be seen as a powerful tool to aid legal professionals to increase their access to justice, and ultimately address unmet needs of the legal community.

### 3.4 Benchmarks and Datasets

To complement our discussion so far, here we provide a summary of the main benchmarks and datasets that have been recognized as relevant in the TLM-based legal learning context. Our main focus is on those corpora that were used by the approaches covered in this work, which will be described next (Section 4). Note that we shall leave out of consideration the datasets used in the COLIEE Competitions, since they have already been described in Section 3.1.

Our presentation is organized into three subsections, which describe corpora concerning caselaw documents, codes, and a combination of both, respectively; moreover, each subsection is further organized by possibly grouping corpora that are cohesive in terms of data type and task. Table 3 summarizes the datasets that we shall describe through this section, according to the legal document category, the data type, the source, the size, and the tasks for which the benchmarks were designed.

#### 3.4.1 Caselaw data

[Strickson and Iglesia, 2020] propose a corpus of about 5K labeled UK court judgments, gathered from the web, for the task of JLP. Each law case is divided into separate judgments issued by individual judges, and each sentence in a judgment is labeled as "allow" or "dismiss" through a pattern matching approach. The dataset is used for the JLP task as a classification problem, whereby classic machine learning classifiers (e.g., support vector machine, random forest, logistic regression) are evaluated.

*ECHR* [Chalkidis et al., 2019a] contains allegations of violated provisions regarding the European Convention of Human Rights.[42] Each case includes a list of facts and a score, provided by the Convention, representing the importance of the case in the case law's development. Also, each case is mapped to the violated articles of the Convention. Moreover, [Quemy and Wrembel, 2022] present *ECH-OD*, a new database for storing and managing ECHR cases. This is designed to be automatically maintained and used as a unified benchmark to compare machine learning methods for the legal domain. The authors have provided the whole pipeline for the benchmark data extraction, transformation, integration, and loading as open-source software.

*Swiss-Judgment-Prediction* (SJP) [Niklaus et al., 2021] comprises 85K cases, in diachronic order, from the Federal Supreme Court of Switzerland (FSCS).[43] The evaluation task is a binary classification of the judgment outcome (i.e., approval or dismissal). The dataset includes cases written in German, French and Italian, and is annotated with publication years, legal areas and cantons of origin. [Niklaus et al., 2021] evaluate XLNet, RoBERTa, AlBERT (cf. Section 2), GermanBERT, UmBERTo, CamemBERT (cf. Section 4.6), and two variants namely Hierarchical BERT and Long BERT, both in monolingual or multilingual versions (cf. Section 4.7).

*GerDaLIR* [Wrzalik and Krechel, 2021] is a dataset for *legal precedent retrieval* on German language.[44] It is based on case laws gathered from the *Open Legal Data* [Ostendorff et al., 2020]. Passages containing references are considered queries, while the referenced law cases are labeled as relevant. The authors evaluate a set of retrieval methods on this dataset with Transformer-based re-ranking. In particular, they fine-tune GBERT and GELECTRA base versions using top-100 BM25 passage rankings and test the final models on top-1000 BM25 passage ranking; the use of ELECTRA for re-ranking has shown to lead to higher performances in most cases. [Urchs et al., 2021] introduce two further legal corpora for the German law. The first corpus[45] contains about 32K decisions, enriched with metadata, from hundreds of Bavarian courts. There are 22 different types of decisions in the corpus, such as resolutions, judgments and end-judgments. This corpus is not intended for a specific task (for example, it can be used to detect the type of the decision). The second corpus[46] is a subset of the former and contains 200 judgments, whose sentences (about 25K) were annotated by a domain expert w.r.t. four components of the text (written in the *Urteilsstil* style): "conclusion" (i.e., the overall result of the case), "definition" (i.e., abstract legal facts and consequences), "subsumption" (i.e., the ensemble of concrete facts and determination sentence), and "other" (i.e., sentences not labeled with any of the three previous labels). This corpus is intended for the automatic detection of conclusion, definition and subsumption components.

---

[42] https://archive.org/details/ECHR-ACL2019
[43] https://huggingface.co/datasets/rcds/swiss_judgment_prediction
[44] https://github.com/lavis-nlp/GerDaLIR
[45] https://zenodo.org/record/3936726#.ZAdMIXbMJD_
[46] https://zenodo.org/record/3936490#.ZAdN7HbMJD_

[Zhong et al., 2019b] provide 92 expert-annotated extractive summaries of *Board of Veterans' Appeals* (BVA) cases focused on post-traumatic stress disorder (PTSD), along with 20 test cases quadruple-annotated for agreement evaluation and two summaries for each test case and written by law experts.[47] Each sentence is annotated considering six labels, namely issue, procedural history, service history, outcome, reasoning, evidential support. Also, [Walker et al., 2019] introduce a dataset to test the performance of rule-based script classifiers, comprising 50 fact-finding decisions of BVA cases focused on veterans' appeals to a rejected disability claim for service-related PTSD. Each sentence of the dataset is assigned a rhetorical role by domain experts as follows: finding sentence, if it states a finding of the fact, evidence sentence, if it states the content of a testimony, reasoning sentence, if it reports the reasoning of the judge underlying the findings of facts, legal-rule sentence, if it states legal rules in the abstract, and citation sentence, if it refers to legal authorities and other materials. The dataset is used to test two hypotheses: whether distinctive phrasing allows automatic classifiers to be developed on a small set of labeled decisions, and whether semantic attribution theory can provide a general approach to develop such classifiers. Results demonstrate that some use cases can be addressed using a very small set of labeled data.

*Multi-LexSum*[48] is a collection of almost 9K expert-edited abstractive summaries for 40K writings of the Civil Rights Litigation Clearinghouse (CRLC),[49] which provides information on federal US civil rights cases for various target audiences (lawyers, scholars, and the general public) [Shen et al., 2022]. It is designed for multi-document and single-document summarization tasks. The source documents are extremely long, with cases often having more than two hundred pages. Multi-LexSum provides multiple summaries with different granularity (from an extreme one-sentence summaries to summaries with more than five hundred words). Although the provided summaries are abstractive, they present a high fraction of terms included also in the source document.

*RulingBR* [Feijó and Moreira, 2018] [50] comprises 10K Brazilian rulings for summarization on legal tasks, which were retrieved from the decision documents of the highest court in Brazil, *Supremo Tribunal Federal* (STF).[51] Each decision document is composed of the following four parts: "Ementa" (i.e., summary), "Acordao" (i.e., judgment), "Relatorio" (i.e., report), and "Voto" (i.e., vote). The Ementa part is used as gold summary for the dataset. [Lage-Freitas et al., 2022] also propose a dataset consisting of about 4K legal cases from a Brazilian State higher court (*Tribunal de Justica de Alagoas*), with a focus on the Brazilian appeals system, assigning the appeals with labels regarding court decisions. Following [Aletras et al., 2016], the authors assume that there is enough similarity between the case description of legal judgments and appeals lodged by attorneys. Brazilian courts data are scraped from the Web and segmented into sections, identifying the description, decision and unanimity parts; then, description sentences are labeled according to the decision outcome (yes, no, or partial) and unanimity information (unanimity vs. non-unanimity).

*CAIL2019-SCM* [Xiao et al., 2019] is a dataset of about 8K triplets of cases of the *Suprem People's Court of China*, concerning private lending.[52] It was collected from the *China Judgments Online*[53] for the *CAIL* competition, where participants were required to perform a similar case matching task, i.e., to detect which pair of cases in the triplet contains the most similar cases. Every document in the triplet refers to the fact description of a case. The most similar pair within each triple is detected by legal experts. The authors provide some baselines to compare the participants' performance, one of which uses BERT to obtain embeddings of the two cases, for which the similarity score is computed. The CAIL competition was first held in 2018 [Xiao et al., 2018]. In *CAIL2018*, participants were required to perform a legal judgment prediction task divided in three sub-tasks: law article prediction, charge prediction, and term-of-penalty prediction. The input is the fact description of a criminal case, and the associated dataset is divided into two sub-datasets: *CAIL-big* (with more than 1.6M cases) and *CAIL-small* (about 130K cases). [Yu et al., 2022b] extended the fact prediction task data of CAIL 2021 for the explainable legal case matching task. The sentences of a legal case in CAIL 2021 are associated with several tags regarding the issue of private lending. In the proposed dataset, called *eCAIL*,[54] the tagged sentences are considered as rationales. Given two legal cases, the cross-case sentences with identical labels are *pro-rationales* for the matching task, while sentences with different labels are *con-rationales*. A matching label is assigned for the case pair according the tag-overlapping: if there is a overlapping of more than 10 tags the cases are considered as matching, otherwise if there is no overlapping the label corresponds to mismatching, and an overlapping with less than (or equal to) 10 tags is considered as partially matching. The dataset provides 6K legal case pairs, with rationales and explanations (the concatenation of all the overlapped tags) for the matching labels. [Yu et al., 2022b] also provide *ELAM*, a dataset for explainable legal case matching task, containing 5K legal case pairs with the associated matching label, rationales, their alignments and the explanations for the matching decision.

---

[47] https://github.com/luimagroup/bva-summarization
[48] https://multilexsum.github.io/
[49] https://clearinghouse.net/
[50] https://github.com/diego-feijo/rulingbr
[51] https://portal.stf.jus.br/
[52] https://github.com/china-ai-law-challenge/CAIL2019/tree/master/scm
[53] https://wenshu.court.gov.cn/
[54] https://github.com/ruc-wjyu/IOT-Match

The authors collected the legal cases online,[55] which refer to the crime of obstruction of the social management order. Each case is associated with several legal-related tags. To pair the legal cases, the authors randomly selected 1250 query cases and constructed a pool of candidates for each query. From the candidate pool, a case is retrieved based on the number of overlapping tags between the case and the query. Each sentence of a legal case pair is associated with a rationale label, with the support of legal experts. The possible rationale labels are the following: not a rationale, a key circumstance, a constitutive element of a crime, or a focus of disputes. The alignment of the rationales (i.e., pro and con rationales) and the matching label (matching, partially matching or not matching) are then marked. Legal experts are also asked to provide explanations for their matching decision.

*ILDC* (Indian Legal Documents Corpus) [Malik et al., 2021] comprises about 35K cases from the Indian Supreme Court, annotated with the court decisions.[56] This is a corpus intended for court judgment prediction and explanation, which requires a model to predict the final outcome (accept or reject, w.r.t. the appellant) and to provide explanations for the given prediction. To this regard, a portion of the corpus is annotated with explanations given by legal experts, and ranked in order of importance in such as way that a higher rank corresponds to an explanation that is more important for the final judgment. The dataset is divided in $ILDC_{single}$ and $ILDC_{multi}$, depending on whether there is a single decision for documents having one or more petitions, or different decisions for documents with multiple appeals.

[Kalamkar et al., 2022] propose a corpus of 354 Indian legal judgment documents, annotated via crowd-sourcing activity with 12 different rhetorical roles, from different courts (Supreme Court of India, High Courts and district-level courts).[57] The annotation process is designed with the support of legal experts. The corpus is intended for the automatic structuring of legal documents. A Transformer-based model is proposed as baseline for the benchmark. Moreover, the authors propose extractive/abstractive summarization and court judgment prediction tasks as two applications of rhetorical roles, as they test how rhetorical roles could be useful for those tasks. For extractive and abstractive summarization, they experiment with the *LawBriefs* corpus, which comprises 285 expert-authored extractive summaries of Indian court judgments. For the court judgment prediction task, experiments were conducted using the ILDC corpus [Malik et al., 2021].

Two further legal datasets for rhetorical role identification are introduced in [Bhattacharya et al., 2021]. One dataset contains 50 cases from the Supreme Court of India belonging to five law domains: criminal, land and property, constitutional, labour/industrial and intellectual property rights. Such documents are gathered from Thomson Reuters Westlaw India website.[58] The other contains 50 cases from the UK Supreme Court, gathered from the official website of the court.[59]. Both the datasets are labeled with the following seven rhetorical roles: "Facts", "Ruling by Lower Court", "Argument", "Statute", "Precedent", "Ratio of the decision" and "Ruling by Present Court".

[Paul et al., 2022b] introduce a pre-training corpus consisting of about 5.4M Indian court cases. The documents are gathered from several web platforms and come from the Supreme Court and many High Courts of India. The corpus covers various court case domains as well as more than 1K central government acts. The authors further pre-train Legal-BERT@aueb and Legal-BERT@stanford on the proposed corpus and assess its pre-training effectiveness considering several downstream benchmarks, for the Indian as well as English languages. The performance of the pre-trained models have been compared to the BERT, the original Legal-BERT@aueb and Legal-BERT@stanford.

[Bhattacharya et al., 2019a] gather about 17K legal cases of the Supreme Court of India through the website of Westlaw India, which provides documents and related summaries written by domain experts. The authors perform a systematic comparison of several summarization algorithms, such as traditional unsupervised extractive methods (e.g., latent semantic analysis), neural unsupervised extractive methods (e.g., *Restricted Boltzmann Machines* [Verma and Nidhi, 2018] and summarization methods specifically conceived for legal documents, both unsupervised (*CaseSummarizer* [Polsley et al., 2016] and supervised (*LetSum* [Farzindar and Lapalme, 2004]).

[Shukla et al., 2022] provide three legal summarization datasets[60] gathering documents from the Indian and UK laws. The first dataset is *Indian-Abstractive dataset (IN-Abs)*, with about 7K cases of Indian Supreme Court judgments, obtained from the website of *Legal Information Institute of India*,[61] and corresponding abstractive summaries. The second dataset is *Indian-Extractive dataset (IN-Ext)*, with 50 case documents of the Indian Supreme Court labeled with six rhetorical roles (i.e., facts, argument, statute, precedent, ratio of the decision, and ruling by present court) and extractively summarized by domain experts, providing a summary for each rhetorical segment separately (with the exception of ratio and precedent segments that are summarized together). The third dataset is *UK-Abstractive dataset (UK-Abs)*, with 793 case judgments gathered from the website of the UK Supreme Court, which provides also the press (abstractive) summaries of the cases, divided in three segments: "Background to the Appeal", "Judgment", and

---

"Reasons for Judgment". The authors specify three criteria for the evaluation of methods: document-level summaries, segment-wise evaluations (i.e., how the summary represents the logical rhetorical segments in the legal case), and how the summaries are evaluated by domain-experts.

[Niklaus et al., 2022] augment the Swiss-Judgment-Prediction (SJP) dataset introduced in [Niklaus et al., 2021] via machine translation, i.e., translating a document written in one of the three languages (German, Italian, French) into the remaining two languages. A second version of the dataset is also provided by further augmenting SJP with Indian cases of the ILDC corpus, provided by [Malik et al., 2021]. To this regard, they translate all the Indian cases reported in the corpus to German, French and Italian. The authors evaluate several TLMs in relation to cross-domain (i.e., different legal ares), cross-regional (i.e., different regions) and cross-jurisdiction (from Indian to Swiss) transfer learning, whose discussion is demanded to Section 4.7.

*LEX Rosetta* [Savelka et al., 2021] propose a multilingual dataset of about 89K annotated sentences for the task of *automatic functional segmentation*, i.e., segmenting adjudicatory decisions of cases according to the functional role of the parts.[62] The sentences are from 807 documents of several courts, gathered from different sources that include seven countries (Canada, Czech Republic, France, Germany, Italy, Poland, USA), and annotated according to the following types: out of scopes (i.e., sentences that are outside the main document, such as editorial content and appendices), heading (i.e., markers of a section), background (i.e., sentences explaining facts, claims and procedural background), analysis (i.e., sentences containing the court reasoning and applications of law to the facts), introductory summary (i.e., a summary of the discussed case), and outcome (i.e., sentences describing the final decision). The dataset is used to test whatever GRU-based models generalize on different contexts (countries), in the segmentation of cases in three functional types (Background, Analysis and Outcome). To this end, the authors analyze the use of multilingual sentence embeddings of predictive models in three versions: training the model on a single context and evaluating transfer learning on other unseen contexts; training the model on a set of contexts and evaluating transfer learning on other unseen contests; and pooling the data of the target context with data from the other contexts. Results have shown that the second and third versions of the model are more effective.

### 3.4.2 Law code data

*SARA* [Holzenberger et al., 2020] is a dataset for *statutory reasoning* on US tax law.[63] This dataset is comprised of a set of rules extracted from the statutes of the US Internal Revenue Code (IRC),[64] along with a set of questions which would require to refer to the rules for being answered correctly. In fact, IRC contains rules and definitions for the imposition and calculation of taxes, and it is subdivided into sections defining one or more terms (e.g., employment, employer and wages). Each section is normally organized around a general rule, followed by a number of exceptions, and each of its subsections refers to a certain number of slots, which may be filled by existing entities. IRC can hence be framed as a set of predicates formulated in human language so as to require a system to determine whether a subsection applies, and to identify and fill the slots mentioned. Statutory reasoning is addressed as a task of entailment and a task of question answering. In the first task, two paragraphs are manually created for each subsection as test cases: the one describes a case to which the statutes apply, the other one describes a case to which the statutes do not apply. In the second task, test cases are created to predict how much tax a person owes, considering all the statutes and applying arithmetic calculations. In general, this dataset offers some features that allow for reasoning on several aspects, such as reasoning on time, numbers, cross-reference and common-sense. To test the abilities of NLP models on the statutory reasoning problem, the authors have pre-trained the models (e.g., Legal-BERT@jhu) on a large legal corpus obtained extracting tax law documents from the *Caselaw Access Project*,[65] private letter ruling from the *International Revenue Service* (IRS)[66] and unpublished US tax Court cases.

*BSARD* [Louis and Spanakis, 2022] is a French dataset composed of more than 1.1K legal questions labeled by domain experts with relevant articles selected from the 22K law articles gathered from 32 publicly available Belgian codes.[67] The set of questions and associated relevant articles are obtained in collaboration with *Droits Quotidiens* (DQ), an organization composed of a team of experienced jurists, which every year receives many questions from citizens seeking advice on legal issues, retrieves the articles relevant to the questions asked, answers the questions in a manner comprehensible to the applicant and categorizes the set of questions, legal references and answers with tags. The resulting corpus contains a large number of legal topics (related to social security, work, family, justice and so on). BSARD is intended for statutory article retrieval.

---

[62] https://github.com/lexrosetta/caselaw_functional_segmentation_multilingual

[63] https://nlp.jhu.edu/law/

[64] https://uscode.house.gov/browse/prelim@title26&edition=prelim

[65] https://case.law/

[66] https://www.irs.gov/tax-exempt-bonds/teb-private-letter-ruling-some-basic-concepts

[67] https://github.com/maastrichtlawtech/bsard

*GCL* [Papaloukas et al., 2021] is a dataset of about 47K documents regarding Greek legislation and designed for the multi-granular topic classification task, which requires to detect the thematic topic that is representative of a legal document.[68] The thematic topics are available in multi-level hierarchy. The main data source for this dataset is the *Permanent Greek Legislation Code - Raptarchis*, a catalogue of Greek legislation available through the portal e-Themis.[69] The portal provides a thematic index for the catalogue, reflecting the thematic hierarchical categories (topics). The hierarchy is dictated by the structural division in volumes, chapters and subjects, which reflect the levels of thematic topics. The classification task in GLC is divided into three sub-tasks, each of them deals with a level of the hierarchy.

Besides statutes, several benchmarks have been developed about *contracts* of different types. *CUAD* [Hendrycks et al., 2021] is a dataset specialized for *legal contract review*.[70] It includes more than 500 contracts, varying in type and length, with 13K annotations across 41 category labels provided by legal experts. Such category labels regard general information, such as party names, dates, renewal terms and so on, as well as restrictive covenants and revenue risks. Language models are required to detect the portions of a contract (the clauses) related to each label. Evaluations of such models as BERT, AlBERT, RoBERTa and DeBERTa have highlighted that better performance are influenced by model design and training set size.

[Leivaditi et al., 2020] provide a dataset containing 179 annotated documents regarding lease contracts. The annotations consist of entities (related to parties, property, terms and rent conditions, dates/periods) and red flags, i.e., terms or sentences indicating a potential risk for one o more parties (e.g., break option, guarantee transferable, right of first refusal to lease, bank guarantee), so that the dataset is mainly intended for supporting red flag detection and entity extraction tasks. The documents in the dataset are gathered through the EDGAR database, which is accessible through the US Securities and Exchange Commission (SEC).[71] The process of selecting the contracts to be annotated is performed using the BM25 ranking function, which evaluates the relevance of documents w.r.t. keywords/queries that may suggest the presence of red flags. The identification of such keywords/queries and the process of annotation are supervised by domain experts.

*ContractNLI* [Koreeda and Manning, 2021] contains 607 annotated contracts regarding non-disclosure agreements.[72] Such documents are gathered from Internet search engines and EDGAR. By comparing different non-disclosure agreements, a set of 17 hypotheses is obtained. Each document in annotated with respect to its relation with one of the hypotheses (i.e., entailment, contradiction, or not mentioned). If a document is annotated as entailing or contradicting, the spans (i.e., sentence or list item within a sentence) composing the documents are annotated as evidence or not (binary label) of the associated entailment relationship.

*ToS* [Aumiller et al., 2021] is a dataset consisting of Term-of-Service documents, specifically collected for *topic similarity* task. The documents include heterogeneous topics due to the different web sources. Some of the most frequent topics regard limitation of liability, law and jurisdiction, warranty, and privacy. Topics are obtained in a hierarchical way splitting the documents into smaller chunks. The authors define and test a system built on TLMs, which revealed to largely outperform segmentation baselines based on TF-IDF and bag-of-words.

*MAUD* [Wang et al., 2023] is a dataset for the legal multi-choice reading comprehension task and consists of legal texts extracted from 152 public merger agreements gathered from EDGAR. Merger agreements are legal documents regarding public target company acquisitions. In these documents there are special clauses, called deal points, that establish the conditions under which the parties are obliged to complete the acquisition. The deal points are extracted from the merger agreements by lawyers working on the American Bar Association's 2021 Public Target Deal Points Study ("ABA Study"). Moreover, a set of multiple-choice questions are answered by the lawyers for each deal point. One or more questions can be asked for a deal point, and each question can be answered by one or more answers. MAUD contains 92 questions, 8K unique deal points annotations, 39K question-answer annotations (the examples) and 7 deal point categories (e.g., Conditions to Closing, Deal Protection and Related Provisions, Material Adverse Effect).

[Manor and Li, 2019] provide a dataset containing legal contracts and summaries gathered from two websites, *TL;DRLegal*[73] and *TOS;DR*,[74] the purpose of which is to clarify the content of contracts through summaries. More precisely, the former, which deals mainly with software licences of companies, is used as a source for collecting 84 sets of contract agreement sections and corresponding summaries, whereas 412 sets are obtained from TOS;DR website, which focuses on user data and privacy topics of companies. The quality of the proposed summaries is verified by authors through an analysis of levels of abstraction, compression and readability.

---

[68] `https://huggingface.co/datasets/greek_legal_code`
[69] `https://www.secdigital.gov.gr/e-themis/`
[70] `https://github.com/TheAtticusProject/cuad/` The CUAD dataset is also available at `atticusprojectai.org/cuad`
[71] `https://www.sec.gov/edgar.shtml`
[72] `https://stanfordnlp.github.io/contract-nli/`
[73] `https://tldrlegal.com/`
[74] `https://tosdr.org/`

Another important target of interest for the development of benchmarks is represented by *privacy policies*. *PolicyIE* [Ahmad et al., 2021] is a corpus for automating fine-grained information extraction of privacy policies, especially through intent classification and slot filling tasks.[75] PolicyIE consists of about 5K intent and 11K slot annotations of several privacy policies related to website and mobile applications. The retrieved policy documents cover four privacy practices that are included in the *General Data Protection Regulation* (GDPR). Thus, sentences of such policy documents are categorized into the following GDPR-like intent classes: data collection/usage (i.e., what user information is collected, as well as the reason and the modality in which it collected), data sharing/disclosure (i.e., what user information is shared with third parties, as well as the reason and the modality in which it is shared), data storage/retention (i.e., location and time period in which user information will be saved), data security/protection (i.e., what protection measures for user information are taken), other (i.e., privacy practices not included in the other categories). Each sentence is annotated with 18 slot labels, which can be categorized into two overlapping types: type-I, which comprises data and participants to the policy practices (e.g., data provider, data collected, data collector) and type-II, i.e., purposes, conditions, polarity and protection methods. The annotation procedure was performed and monitored by domain experts.

*PrivacyQA* [Ravichander et al., 2019] contains 1750 questions with over 3.5K annotations of relevant answers regarding to privacy policies of mobile applications.[76] Questions for a particular privacy policy are provided by crowd-workers, while the identification of the related answers are committed to legal experts, which also provide meta-annotations on the relevance of the question, OPP-115 category, subjectivity, and the likelihood that the answer to the input question is contained into a privacy policy. The authors test the ability of different baselines on two tasks: deciding if a question is answerable, and identifying evidences in the policies for a given question.

*PolicyQA* [Ahmad et al., 2020] comprises about 25K triplets of question, passage and answer text, derived from segments of website privacy policy documents.[77] The corpus is designed so that the answer consists of small portions of the text that better identify the target information in relation to the question. It is curated from the existing *OPP-115* corpus [Wilson et al., 2016], which consists of 115 website policies (about 3.7K segments) annotated following domain-experts annotation schemes. The annotation schemes categorize the policy segments in ten data practice categories (e.g., first party collection/use), which are further categorized in several practice attributes (e.g., user type), and each practice attribute is assigned a set of values; for instance, user without account, user with account, other and unspecified. The annotated segments with the associated practice, attribute and value categories are used to form the PolicyQA corpus. Segments and categories are provided to skilled annotators to manually label the questions, for a total of 714 individual questions. The associated QA task is answer span prediction given a policy segment. To this regard, two neural baselines are evaluated, one of this is based on BERT.

[Bui et al., 2021] introduce a corpus[78] for the extraction and visualization in privacy policies of personal data objects, i.e., spans in the text expressing types of user data, and related privacy actions. The proposed corpus contains about 4.1K sentences and 2.6K annotated fine-grained data objects concerning several real-world privacy policies. It is obtained exploiting the OPP-115 dataset as a starting point, opting for the top US websites that cover several domains like banking, e-commerce, social network. The data objects in the privacy policies are detected by annotators with experiences in privacy and security research. The data objects are then labeled by the annotators, choosing among "collect", "not_collect", "share" and "not_share" labels. Such labels indicate the privacy action performed on the user data (collection or sharing). The resulting annotation has also been revised with a semi-automated process to improve the annotation quality, which involves the use of tools for correction and pre-annotation. The final corpus is used to train and evaluate a neural NER model, called *PI-Extract*, on the extraction of personal data objects and privacy actions. The task is formulated as a sequence labeling problem, which is to assign a label for each token of a given sentence.

Relevant benchmarks have been built by considering multilingual and/or multi-task evaluation scenarios. For instance, *COVID-19 Exceptional Measures* [Tziafas et al., 2021] is a collection of legal, manually-annotated documents regarding COVID-19 exceptional measures across 21 European countries for multilingual classification task. To this end, feature-based methods and XLM-RoBERTa pre-trained on the collection have been evaluated, showing best results in the use of the domain-adapted TLMs.

*MultiEURLEX* [Chalkidis et al., 2021a] consists of European Union laws, annotated with multiple labels and translated in 23 languages, with legal topic classification as supported task.[79] The authors experiment with monolingual BERT models, pre-trained in Romance, Slavic, Germanic and Uralic languages, and multilingual models (mT5 and XLM-RoBERTa) which are evaluated for cross-lingual legal text classification on this benchmark. The experimentation focuses mainly on the zero-shot cross-lingual transfer, namely one-to-many setting, in which a multilingual model is fine-tuned on one language and evaluated in the other 22 languages. However, models are also evaluated on one-to-one (training and testing on the same language) and many-to-many (training and testing on all languages) settings. Adaptation strategies on the multilingual models are applied to avoid the catastrophic forgetting of multilingual knowledge when fine-tuning on one source language only, significantly improving the zero-shot cross-lingual transfer. In the one-to-one setting, multilingual models prove to be competitive against monolingual models.

---

[75] https://github.com/wasiahmad/PolicyIE

[76] https://github.com/AbhilashaRavichander/PrivacyQA_EMNLP

[77] https://github.com/wasiahmad/PolicyQA

[78] https://github.com/um-rtcl/piextract_dataset

[79] https://huggingface.co/datasets/multi_eurlex

*EUR-Lex-Sum* [Aumiller et al., 2022] is a multi- and cross-lingual dataset containing about 32K pairs of documents and summaries in 24 languages.[80] Each language comprises up to 1500 pairs. Documents consist of legal acts retrieved from the European Union law website,[81] 375 of which are legal acts written in each of the languages. Summaries are structured following particular guidelines, for example there are sections dedicated to key points, background, key terms and so on. The authors evaluate several zero-shot extractive baselines, one of which is a version of LexRank that receives chunks (based on existing separators in the text) and uses embeddings generated by SBERT, and cross-lingual baselines, including one based on LED with capability of greedily chunking the text if document sizes exceed the model's maximum input length.

*LegalNERo* [Pais et al., 2021] contains 370 legal documents, designed for NER tasks and manually annotated according to five coarse-grained classes: person, location, organization, time expressions, and legal document references.[82] The documents are extracted from a larger corpus, called *MARCELL-RO*,[83] containing several documents from national legislation (e.g., decrees, regulation, laws) of seven countries, Romania included. The authors evaluate a baseline based on BiLSTM and CRF, which takes as input a text representation obtained through FastText.[84]

*PLUE* [Chi et al., 2023] is a multi-task benchmark that collects several privacy policy datasets (including the aforementioned datasets) to evaluate NLP methods over various privacy policy tasks, namely classification, question-answering, intent classification, slot filling, name entity recognition.[85] In particular, PLUE contains the following datasets: OPP-115, APP-350 [Zimmeck et al., 2019], PrivacyQA, PolicyQA, PolicyIE, and PI-Extract (the dataset used in [Bui et al., 2021]). To enable model pre-training on the domain, [Chi et al., 2023] also provide a large corpus using *MAPS* [Zimmeck et al., 2019], a corpus of 441K mobile application privacy policies, and the *Princeton-Leuven Longitudinal Corpus* (PLLC) [Amos et al., 2021], containing bout 130K privacy policies of websites. From the combination of the two corpora, a pre-training corpus with 332M words is created. The authors evaluated several TLMs as baseline, previously pre-trained on MAPS and PLLC and then fine-tuned on the PLUE datasets.

[Drawzeski et al., 2021] introduce a multilingual corpus for the analysis of *fairness of online terms of service*. The dataset contains 100 contracts, derived from 25 ToS documents annotated in four languages (English, German, Italian and Polish) and extracted from an existing corpus [Lippi et al., 2019]. In each contract, potentially unfair clauses are labeled with one of the nine possible unfairness categories, namely arbitration, unilateral change, content removal, jurisdiction (i.e., which courts will have jurisdiction over disputes under the contract), choice of law (i.e., which law will regulate the contract), limitation of liability, unilateral termination, contract by using (i.e., the use of a service binds the consumer to the terms of use of the service without being required to indicate that she/he has read and accepted them), and privacy included (i.e., the use of the service implies the acceptance of the related privacy policy). Moreover, for each category the degree of the unfairness is indicated with three numerical values: 1 for clear fairness, 2 for potential unfairness, and 3 for clear unfairness. Four types of discrepancies are observed across the language versions of the same contract, relating to the sentence structures, errors or inaccuracies in translation into the target languages, absence of some clauses in the different language versions and the choice of legal terminology.

### 3.4.3 Hybrid data

*LexGLUE* [Chalkidis et al., 2022b] is a collection of seven existing legal NLP datasets[86] for evaluating models across several legal tasks, which include multi-label classification, multi-class classification and multiple choice question-answering:

- *ECtHR Tasks A & B* for multi-label classification of allegations regarding violations of the European Convention of Human Rights (ECHR) provisions. The dataset is used to test models on article violation prediction (Task A, [Chalkidis et al., 2019a]) and alleged violation prediction (Task B, [Chalkidis et al., 2021c]). In both Task A and Task B, the total number of ECHR articles is reduced to 10, discarding those that are rarely discussed, cannot be violated or are not depending on the facts of a case.

---

[80] `https://github.com/achouhan93/eur-lex-sum`
[81] `https://eur-lex.europa.eu/`
[82] `https://lod-cloud.net/dataset/racai-legalnero`
[83] `https://marcell-project.eu/`
[84] `https://fasttext.cc/`
[85] `https://github.com/JFChi/PLUE`
[86] `https://huggingface.co/datasets/lex_glue`

- The English part of *Multi-EURLEX*[Chalkidis et al., 2021a] for multi-label classification on European Union (EU) legislation. It includes different labeling granularity levels (from 21 to 7K EuroVoc concepts). In LexGLUE, the 100 most frequent labels from the second level of granularity (567 total labels) are considered.
- *SCOTUS*[87] for multi-class classification on US Supreme Court opinions. In LexGLUE, SCOTUS opinions are associated with 14 issue areas (e.g., Economic Activity, Criminal Procedure, Civil Rights) obtained through the Supreme Court DataBase.[88]
- *LEDGAR* [Tuggener et al., 2020] for multi-class classification on contract provisions gathered from the US Securities and Exchange Commission (SEC) documents. In LexGLUE, a subset of the original dataset is considered, derived from the 100 most frequent labels.
- *UNFAIR-ToS* [Lippi et al., 2019] for multi-label classification on contracts between providers and users of services (i.e., terms of service) with 8 classification labels.
- *CaseHOLD* [Zheng et al., 2021] for multiple choice question-answering about holdings of US court cases gathered from the Harvard Law Library case law corpus.

[Chalkidis et al., 2022b] evaluate BERT, RoBERTa, DeBERTa, Longformer, BigBird, Legal-BERT@aueb and Legal-BERT@stanford on LexGLUE. For ECtHR and SCOTUS, the authors employ a hierarchical variant of the models, following [Chalkidis et al., 2021c]. The domain-adapted models, i.e., Legal-BERT@aueb and Legal-BERT@stanford, performed overall better than competitors, with large improvement in US case law data.

*FairLex* [Chalkidis et al., 2022c] comprises four datasets (ECtHR, SCOTUS, Swiss-Judgment-Prediction and CAIL [Wang et al., 2021]) for the evaluation of *model fairness*.[89] To this end, it includes three groups of fairness attributes: demographics, regional and legal area. The first group regard biases relating to factors such as gender, age and race. The second and third groups aim to alleviate disparity, respectively, in regions of a given jurisdiction and in different areas of law. Moreover, it contains five languages (Chinese, English, French, German and Italian) and four jurisdictions (China, USA, Switzerland and European Council). The authors also provide four hierarchical BERT-based models, one for each dataset, as baselines for the benchmark. Such models are similar to [Chalkidis et al., 2021c] and further pre-trained on the specific dataset. The models are warm-started from MiniLMv2 checkpoints, using a distilled version of RoBERTa for the English version and a distilled version of XLM-R for the other languages. Experimental results show that the models have some disparity in performance. In particular, in the ECtHR task there is a disparity related to defendant state and applicant's gender, while for the FSCS task there is disparity related to language (Italian versus French and German), legal areas (penal law versus the others) and court regions (Switzerland courts versus federation courts). Court regions disparity is noted also in the CAIL task (Beijing courts versus Sichuan courts). However, disparities in performance can be influenced by general factors based on the distribution of data.

[Bhattacharya et al., 2020a] collect documents of the Supreme Court of India and statutes in the Indian judiciary through the Thomson Reuters Westlaw India website,[90] for a task of document similarity. In particular, they propose and evaluate an approach based on *precedent citation network* augmented with the hierarchy of legal statutes, in order to encompass also the knowledge of legal text's hierarchy.

*Pile of Law* [Henderson et al., 2022] is a legal corpus designed to address *ethical issues* in the pre-training phase.[91] It is collected from 35 EU and US data sources and covers several legal sub-domains, such as court opinions, administrative rules, contract and legislative records, for a total of about 10M documents. Such data has already implicit filters which reflect legal standards of the specific jurisdiction, but the authors note that not all of them have been detected and such norms can vary respect to the jurisdiction. By performing filtering rules on the data, the proposed dataset respects the legal norms of governments and courts regarding the presence of toxic (offensive or obscene terms) or private content and prevents a model to learn such information. The authors demonstrate that the dataset can be used to learn contextual privacy/toxicity rules, as it respects the variation in the different privacy/toxicity norms. For example, they demonstrate models pre-trained on Pile of Law can learn contextual privacy rules with regard to the use of pseudonyms in immigration court and in civil litigation. In particular, a BERT-based model is trained on the data to predict whether a pseudonym should be used. Moreover, [Henderson et al., 2022] provide a baseline by pre-training BERT on the corpus from scratch. The resulting model, called *PoL-BERT-Large*, is fine-tuned and evaluated for a legal reasoning task on CaseHOLD, reaching about the same performance reported in LexGLUE, and outperforms BERT. However, it does not outperform a BERT model trained exclusively on case law data. This is probably due to the extreme data diversity in the corpus that limits the pre-training efficacy respect to competitors trained exclusively on task-oriented data such as CaseHOLD.

---

[87] https://www.supremecourt.gov/
[88] http://supremecourtdatabase.org/
[89] https://huggingface.co/datasets/coastalcph/fairlex
[90] https://www.westlawasia.com/
[91] https://huggingface.co/datasets/pile-of-law/pile-of-law

**Table 3.** Summary of benchmarks and datasets discussed in Section 3.4.

| Reference | Category | Data type | Data source | Data size | Tasks |
|---|---|---|---|---|---|
| [Strickson and Iglesia, 2020] | cases | UK court judgments | web sources | 5K documents | LJP |
| [Chalkidis et al., 2019a] | cases (allegations) | ECHR cases | ECHR's public database (HUDOC) | 11.5K documents | LJP (binary violation), ALVP, case importance prediction |
| [Chalkidis et al., 2021c] | cases (allegations) | ECHR cases | ECHR's public database (HUDOC) | 11K documents | ALVP |
| [Niklaus et al., 2021] | cases | Federal Supreme Court of Switzerland cases | entscheidsuche platform | 85K documents | LJP |
| [Wrzalik and Krechel, 2021] | cases | case laws from Open Legal Data | Open Legal Data platform | 123K query passages, 131K documents | LPR |
| [Urchs et al., 2021] | cases | cases from Bavarian courts | gesetze-bayern platform | 32K docs, 200 docs with 25K sentences | automatic detection of sections |
| [Zhong et al., 2019b] | cases (appeals) | US Board of Veterans' appeals cases | US Department of Veterans Affairs platform | 92 documents | ES |
| [Walker et al., 2019] | cases (appeals) | US Board of Veterans' appeals cases | US Department of Veterans Affairs platform | 50 documents, 6K sentences | SC |
| [Shen et al., 2022] | cases | writings providing information on US federal civil rights cases | Civil Rights Litigation Clearinghouse platform | 9K documents | AS |
| [Zheng et al., 2021] | cases (holdings) | US court cases | existing corpus (Harvard Law Library corpus) | 53K questions | MCQA |
| [Feijó and Moreira, 2018] | cases | Brazil federal court decisions | Supremo Tribunal Federal platform | 10K documents | ES |
| [Lage-Freitas et al., 2022] | cases | Brazilian State higher court cases | web sources | 4K documents | LJP, unanimity decision |
| [Xiao et al., 2019] | cases | Supreme People's Court of China cases | China Judgments Online platform | 8K document triplets | CM |
| [Yu et al., 2022b] | cases | Supreme People's Court of China cases | existing dataset (CAIL) Faxin platform | 6K documents pairs 5K documents pairs | explainable CM |
| [Malik et al., 2021] | cases | Supreme Court of India cases | IndianKanoon platform | 35K documents | CJPE |
| [Kalamkar et al., 2022] | cases | Indian courts' legal judgments | Indian court platforms (Supreme Court of India platform, eCourt India services platform) | 354 documents, 40K sentences | automatic legal document structuring |
| [Bhattacharya et al., 2021] | cases | Supreme Court of India cases UK Supreme Court | Thomson Reuters Westlaw India platform UK Supreme Court platform | 50 documents 50 documents | RRL |
| [Paul et al., 2022b] | cases | Supreme Court and High Courts of India cases | web platforms (Supreme Court of India platform, IndianKanoon platform) | 5.4M documents | pre-training |
| [Bhattacharya et al., 2019a] | cases | Supreme Court of India cases | Thomson Reuters Westlaw India platform | 17K documents | AS |
| [Bhattacharya et al., 2020a] | cases | Supreme Court of India cases, statutes in the Indian judiciary | Thomson Reuters Westlaw India platform | 1.8K documents | document similarity |
| [Shukla et al., 2022] | cases | Supreme Court of India case judgments, UK Supreme Court case judgments | Legal Information Inst. of India platform, UK Supreme Court platform | ∼7K docs (India), 793 docs (UK) | AS, ES |
| [Niklaus et al., 2022] | cases | Federal Supreme Court of Switzerland and Indian cases | existing datasets (SJP and ILDC) | ∼180K Swiss docs, 96K Indian docs | LJP, cross-domain/regional/ jurisdiction transfer learning |
| [Savelka et al., 2021] | cases | documents from several courts | web platforms, random courts (CA); Constitutional/Supreme/Supreme admin. courts (CZ); Cour de cassation (FR); federal courts (DE); criminal courts (IT); Supreme Court, Constitutional tribunal (PL); Federal district court, Dept. of Labor (US) | 807 documents with 89K sentences | functional segmentation |

**Table 4.** *

Table 3. *(Cont'd)* Summary of benchmarks and datasets discussed in Section 3.4.

| Reference | Category | Data type | Data source | Data size | Tasks |
|---|---|---|---|---|---|
| [Holzenberger et al., 2020] | statutes (rules) | US tax law | Office of the Law Revision Counsel platform - US Internal Revenue Code | 276 train, 100 test docs | entailment and tax amount prediction |
| [Louis and Spanakis, 2022] | statutes | Belgian codes (federal/regional authorities) | Droits Quotidiens organization | 1.1K questions, 22K articles | SAR |
| [Papaloukas et al., 2021] | statutes, regulations | Greek legislation (laws, Royal/Presidential decrees, regulations and decisions, from the Official Government Gazette) | Permanent Greek Legislation Code (Raptarchis) through e-Themis platform | 47K documents | multi-level TpC |
| [Hendrycks et al., 2021] | contracts | 25 types of legal contracts (e.g., license, consulting, service, manufacturing) | SEC's EDGAR database | 510 documents | legal contract review, relevant text prediction |
| [Leivaditi et al., 2020] | contracts | lease contracts | SEC's EDGAR database | 179 documents | sentence detection, entity extraction |
| [Tuggener et al., 2020] | contracts | Exhibit-10 material contract agreements | SEC's EDGAR database | 60K documents | text classification |
| [Koreeda and Manning, 2021] | contracts | non-disclosure agreements | web sources and SEC's EDGAR database | 607 documents | NLI |
| [Aumiller et al., 2021] | contracts | terms of service | web sources | 40K documents | topic similarity |
| [Wang and Li, 2023] | contracts | merger agreements | SEC's EDGAR database | 39K question-answer annotations | MCRC |
| [Manor and Li, 2019] | contracts | contracts relating to software licenses, privacy policies, terms of service/use | web platforms (TL;DRLegal, TOS;DR) | 84 agreement-summary pairs, 421 agreement-(up to 3)summary pairs | abstractive similarity |
| [Ahmad et al., 2021] | contracts | privacy policies | mobile applications and web sources | 5.2K sentences | intent classification, SF |
| [Ravichander et al., 2019] | contracts | privacy policies | mobile applications | 1.7K questions, 3.5K sentences | question answerability, evidence extraction |
| [Ahmad et al., 2020] | contracts | privacy policies | web sources | 25K triples | answer span prediction, QA |
| [Bui et al., 2021] | contracts | existing dataset (OPP-115) based on web sources | | 4.1K sentences | personal data object extraction, NER |
| [Tziafas et al., 2021] | regulations | COVID-19 exceptional measures | web platforms | 4K sentences | SC |
| [Chalkidis et al., 2021a] | statutes, regulations | European Union laws | EUR-Lex platform | 65K documents | TpC |
| [Aumiller et al., 2022] | statutes, regulations | legal acts of 20 domains (e.g., taxation, energy, transportation, industrial policy, health protection) | EUR-Lex platform | 32K doc-summary pairs, ~375 docs as val/test sets for all languages, remaining docs as lang.-specific train set | AS, ES |
| [Pais et al., 2021] | statutes, regulations | decrees, regulation, laws of several countries | existing dataset (MARCELL-RO) based on the Romanian legislative platform | 370 documents 370 documents | NER NER |
| [Chi et al., 2023] | contracts | privacy policies | existing datasets (OPP-115, APP-350, PrivacyQA, PolicyQA, PolicyIE, PI-Extract) related to mobile applications and web sources | 18.7K segments, 1.7K questions, 25K triples, 9.2K sentences | intent classification, SF, QA, NER |
| [Lippi et al., 2019] | contracts | terms of service | web platforms | 9.4K sentences | clause detection and type classification |
| [Drawzeski et al., 2021] | contracts | terms of service | existing dataset (UNFAIR-ToS) related to web platforms | 100 documents | sentence-level annotations projection |
| [Chalkidis et al., 2022b] | cases (allegations, holdings), contracts | ECHR cases, US Supreme Court opinions (Criminal Procedure, Civil Rights, Economic Activity), US court cases, terms of service | existing datasets (ECHR Tasks A & B, Multi-EURLEX, SCOTUS, LEDGAR, UNFAIR-ToS, CaseHOLD) based on web platforms, | ~164 docs, 9.4K sentences, 53K questions | AVP, ALVP TpC, multi-class/label classification, MCQA |
| [Henderson et al., 2022] | cases, regulations, contracts, | EU and US documents (Court Listener Opinions, US Board of Veterans' Appeals, EUR-Lex, EDGAR contracts, ECHR, US Code) | web sources, existing datasets | 10M documents | pre-training |
| [Chalkidis et al., 2022c] | cases | ECHR cases, US Supreme Court opinions, Federal Supreme Court of Switzerland cases, Supreme People's Court of China cases | existing datasets (ECtHR, SCOTUS, SJP and CAIL), based on web platforms | 209K documents | fairness on ALVP, TpC, case approval prediction, crime severity prediction |

# 4 Methods

In this section we discuss the main methods based on the TLMs described in Section 2 and their application to the problems and tasks presented in Section 3. More specifically, we organize our presentation of the methods into six parts. Starting from early methods to more complex architectural settings, the first part (Section 4.1) includes methods that are mainly based on fine-tuning BERT and the other TLMs. Given the importance of the COLIEE competition over the past few years, we dedicate a separate part (Section 4.2) to the methods that were designed and competed for the COLIEE tasks. Section 4.3 discusses domain-adaptive approaches, which require pre-training of TLMs on legal corpora. Section 4.4 focuses on approaches coping with the text limitation length of BERT and other TLMs. Section 4.5 discusses GPT-based methods. Sections 4.6 and 4.7 are devoted to non-English and multilingual approaches, respectively. Finally, Section 4.8 discusses explainability and interpretability in TLM-based methods.

As a guide to our discussion, Tables 5–8 report on main characteristics of the approaches that we shall describe through this section, focusing on the adopted TLMs, the downstream tasks for which the approaches were designed, the types of data and languages, and whether a method is conceived to deal with long documents.

It should be noted that our objective is not providing a fully detailed description of each work, and most details about techniques that are not pertinent to TLMs might be discarded; also, we will not delve into the experimental results. Clearly, we refer the interested reader to the original works for further information.

## 4.1 Task-adaptive methods

Early applications of BERT and related models to the legal domain concern approaches that make TLMs *adaptive* to a legal specific task, i.e., they directly fine-tune an original, general-domain pre-trained model to the task at hand. Accordingly to the date of release of BERT, such applications have started being developed since 2019. For instance, [Howe et al., 2019] provide a comparative evaluation of various text classification approaches, including BERT, topic models and word embeddings, for classifying Singapore Supreme Court judgments into legal areas. [Gain et al., 2019] fine-tune BERT for a sentence-pair classification task to address the COLIEE-2019 Task 4 context. [Rabelo et al., 2019a] fine-tune BERT on the training data available for the COLIEE-2019 Task 2, and use the model in the post-processing stage of a framework that exploits a combination of a multi-word token similarity score and a noun-phrase similarity score to identify candidate entailing paragraphs. This combined approach was ranked first for the COLIEE-2019 Task 2, with an F-score of 0.70 on the official COLIEE test dataset. In the same context of COLIEE-2019 Task 2, [Yamada and Tokunaga, 2019] fine-tune BERT for a sentence-pair classification task, where each entailed fragment and target paragraphs are regarded as sentences. Three variants are defined to handle different lengths of an input sentence-pair, and an ensemble model BERT-vote is implemented using a simple voting method based on a majority prediction for each instance according to the three BERT-based models.

In the context of the AILA-2020 task of rhetorical role labeling, [Gao et al., 2020] combine TF-IDF features with the fine-tuned BERT embeddings (i.e., the feature vectors of the [CLS] token) to fed logistic regression, SVM, and AdaBoost classification models for final training and prediction. Results have shown that the joint semantic-statistical feature modeling improves upon the individual models.

[Elwany et al., 2019] fine-tune BERT on a proprietary corpus of legal contracts for the task of extracting the fixed-term and the auto-renewing-term portions of an agreement. Two main strategies are developed to build the model depending on whether the BERT layers are frozen or left unfrozen during the fine-tuning process.

Marginal yet practically valuable improvements in both accuracy and training speed were observed for the fine-tuned BERT model.

[Sanchez et al., 2020] employ BERT in its regression form for passage retrieval, i.e., to predict a relevance score for a query-document pair in input, where each document is represented by a combined title and summary field along with the content of a news article. On a similar fashion, [Mistica et al., 2021] fine-tune BERT on a dataset of online help requests for a multi-label classification task pertinent to different areas of law.

[Ravichander et al., 2019] fine-tune BERT on the PrivacyQA benchmark for two tasks: to identify if a question is answerable and to identify the evidence in a privacy policy that answers the question. to predict only the answerability of the question, then at inference time, if the question is considered answerable, the model produces the evidences. Analogously, [Ahmad et al., 2020] fine-tune BERT on the PolicyQA benchmark, whose task is to predict the answer span of a related question, given the policy segment. This time, however, the model is first pre-trained on a dataset, SQuAD, which is regarded as related to the PolicyQA corpus. The positive impact of pre-training with SQuAD and fine-tuning on the benchmark is demonstrated by the fact that the same model without pre-training or fine-tuning (or both) performed lower. A different use of BERT in the policy context is adopted by [Bui et al., 2021], which propose *PI-Extract* for the extraction of personal data and related actions in privacy policies. PI-Extract consists of four BiLSTM-CRF-based NER models trained on a privacy policy corpus introduced by the authors (cf. Section 3.4) which contains four possible labels to be assigned to the personal data objects contained in an input sentence, where each label expresses the type of privacy action performed on the personal data. Each NER model of PI-Extract is specifically trained to predict only one of four possible labels. In PI-Extract, BERT is used as a more effective alternative to word embedding based on GloVE [Pennington et al., 2014] to encode the text input.

**Table 5.** Summary of main methods discussed in Sections 4.1 and 4.2.

| Method | Ref. | Downstream Tasks | Lang. | Data | Long docs? |
|---|---|---|---|---|---|
| Fine-tuned BERT | [Howe et al., 2019] | TC | EN | legal cases | No |
| Fine-tuned BERT | [Rabelo et al., 2019a] [Yamada and Tokunaga, 2019] | COLIEE-2019 Task 2 | EN | legal cases | No |
| Fine-tuned BERT | [Gain et al., 2019] | COLIEE-2019 Task 4 | EN | civil code | No |
| Fine-tuned BERT w/ TF-IDF | [Gao et al., 2020] | RRL | EN | legal cases | No |
| Fine-tuned BERT w/ and w/o frozen layers | [Elwany et al., 2019] | NER | EN | contracts | No |
| Fine-tuned BERT in regression form | [Sanchez et al., 2020] | PR | EN | legal news articles | No |
| Fine-tuned BERT | [Ravichander et al., 2019] | QA | EN | contracts | No |
| Fine-tuned BERT | [Ahmad et al., 2020] | QA | EN | contracts | No |
| BiLSTM-CRF + BERT (PI-Extract) | [Bui et al., 2021] | NER | EN | contracts | No |
| Fine-tuned BERT | [Mistica et al., 2021] | TC | EN | help requests | No |
| Fine-tuned RoBERTa | [Westermann et al., 2020] | COLIEE-2020 Task 4 | EN | civil code | No |
| Fine-tuned RoBERTa + BiLSTM | [Majumder and Das, 2020] | RRL | EN | legal cases | No |
| Fine-tuned RoBERTa | [Vold and Conrad, 2021] | QA | EN | PRIVACYQA | No |
| Fine-tuned BERT and RoBERTa | [Chalkidis et al., 2020a] | TC | EN | EURLEX57K; MIMICIII; AMAZON13K | No |
| Fine-tuned RoBERTa | [Savelka et al., 2020] [Savelka and Ashley, 2021] | TC | EN | legal cases | No |
| Fine-tuned T5 | [Hudzina et al., 2020] | COLIEE-2020 Task 4 | EN | civil code | No |
| Fine-tuned BERT and SpanBERT | [Kruiper et al., 2021] | MWE discovery | EN | SPAR.txt | No |
| Fine-tuned BERT + BigBird | [Hong et al., 2021] | IE | EN | parole hearings | No |
| Fine-tuned XLM-RoBERTa w/ GLoVe embeddings | [Trias et al., 2021] | NER | EN | CAP | No |
| Fine-tuned BERT, RoBERTa, DistilBERT, XLNet and m-BERT | [Shaheen et al., 2020] | LMTC | EN | JRC-Aquis and EURLEX57K | No |
| Fine-tuned BERT | [Chalkidis et al., 2019b] | LMTC | EN | EURLEX57K | No |
| Triplet-loss-network w/ fine-tuned SBERT encoder | [Sarkar et al., 2021] | TC | EN | promissory and non-promissory sentences | No |
| Fine-tuned BERT, RoBERTa, Transformer + BiLSTM + CRF; Fine-tuned MiniLM, UniLM, UniLMv2, MASS and BART | [Ahmad et al., 2021] | intent classification, SF | EN | contracts | No |
| BERT; fine-tuned BERT; BERT in regression form; ensemble of BERT and fine-tuned BERT; BERTLaw | [Nguyen et al., 2020] | COLIEE-2020 Tasks 1-4 | EN | legal cases; civil code | No |
| Ensemble of fine-tuned BERT and BM25 | [Nguyen et al., 2021a] | COLIEE-2021 Tasks 1, 2 | JP | legal cases | No |
| Fine-tuned BERT, BART, and RoBERTa + Random Forest | [Rabelo et al., 2020] | COLIEE-2020 Task 2 | EN | legal cases | No |
| Fine-tuned BERT | [Alberts et al., 2020] | COLIEE-2020 Task 2 | EN | legal cases; NLI | No |
| monoT5-zeroshot; fine-tuned monoT5 and DeBERTa; DeBERTa + monoT5 Ensemble | [Rosa et al., 2021] | COLIEE-2021 Task 2 | EN | case paragraphs, decision fragments | No |
| Fine-tuned BERT w/ truncated texts | [Shao et al., 2020b] | COLIEE-2020 Task 2 | EN | legal cases | No |
| Fine-tuned BERT and AlBERT | [Shao et al., 2020a] | COLIEE-2020 Tasks 3, 4 | JP | civil code | No |
| Fine-tuned RoBERTa w/ NLI type detection | [Rabelo et al., 2020] | COLIEE-2020 Task 4 | EN | civil code; SNLI | No |
| Fine-tuned BERT | [Schilder et al., 2021] | COLIEE-2021 Task 3 | JP | civil code | No |
| Fine-tuned SBERT + TF-IDF vectorization + data enrichment; Legal-BERT-FP@aueb + TF-IDF vectorization + data augmentation; BERTScore | [Wehnert et al., 2021] | COLIEE-2021 Task 3 | EN | civil code | No |
| Fine-tuned BERT + XGBoost | [Alberts et al., 2020] | COLIEE-2020 Task 4 | EN | civil code | No |
| Ensembles of fine-tuned ELECTRA, T5 and Multee | [Schilder et al., 2021] | COLIEE-2021 Task 4 | JP | civil code | No |
| DistilRoBERTa and GPT-3 | [Schilder et al., 2021] | COLIEE-2021 Task 5 | JP | civil code | No |
| Fine-tuned BERT ensemble w/ augmentation | [Yoshioka et al., 2021a] [Yoshioka et al., 2021b] | COLIEE-2021 Tasks 3-5 | JP | civil code | No |
| Fine-tuned BERT and Legal-BERT@aueb; Fine-tuned BERT w/ thesaurus concept number | [Kim et al., 2021] | COLIEE-2021 Tasks 2, 4, 5 | EN | legal cases; civil code | Yes |

A few studies have also focused on RoBERTa. [Westermann et al., 2020] fine-tune RoBERTa for sentence-pair classification on the COLIEE-2020 Task 4. [Majumder and Das, 2020] provides RoBERTa embeddings to a BiLSTM for the AILA-2020 task of rhetorical role labeling, reaching the highest rank at the competition. [Vold and Conrad, 2021] fine-tune RoBERTa on the PRIVACYQA dataset, a corpus of questions about privacy policies associated with mobile applications, for binary or graded QA pair classification. [Chalkidis et al., 2020a] evaluate BERT and RoBERTa, along with RNN-based Label-Wise Attention Networks (LWANs) and Probabilistic Label Trees (PLTs) on three English datasets for large-scale multi-label text classification: EURLEX57K [Chalkidis et al., 2019b], MIMICIII, and AMAZON13K. Experimental results have shown that BERT and RoBERTa outperform both PLT-based methods and LWANs, and that a combination between BERT and LWAN can further improve performance.

On the task of determining whether a sentence describes facts or not, [Savelka et al., 2020] develop a meta type system to train models on data from different domains and jurisdictions of annotated legal cases and assess performance of the models across the various domains. Datasets have been re-labeled in such a way as to obtain from the original rhetorical roles only "Fact" and "Non-Fact" labels. To this purpose, RoBERTa base is fine-tuned on the different datasets and combined datasets.

[Savelka and Ashley, 2021] focus on the identification of explanatory sentences, i.e., to detect sentences that are useful for explaining selected legal concepts. In order to classify sentences according to four levels of usefulness, three RoBERTa-*base* models are fine-tuned on different tasks: classifying retrieved sentences in terms of their value for explaining the legal concepts, sentence pair classification, and another type of sentence pair classification where one sentence is a provision of written law and the second is a retrieved sentence. Results point out important interactions among a legal concept, the provision of law in which it is embedded, and retrieved sentences.

Besides the RoBERTa alternative to BERT, [Hudzina et al., 2020] resort to the pre-trained T5-*base* model for the COLIEE-2020 Task 4 and for denoising tasks on Japanese Civil Code article texts and titles and on Wikibook articles about the Japanese Civil Code. Results have shown that T5 appears to overfit the training data despite the multiple domain-specific tasks. [Kruiper et al., 2021] utilize pre-trained BERT-*base* and SpanBERT-*base* cased[92] to learn a sequence tagger for a particular task of multi-word expression (MWE) for Automated Compliance Checking (ACC), which is to identify low-level constituent parts of a sentence, i.e., spans. The objective is to learn a semantic lexicon for ACC, which is essential for semantic parsing as it enables the grounding of information units identified in a text (e.g., objects, interactions, and constraints). SpanBERT shows to be less effective than BERT, which is likely due to a mismatch between the span types and sizes used for the training of the original SpanBERT and the ones used to train the sequence tagger.

[Hong et al., 2021] investigate on information extraction of case factors from a corpus of parole hearing transcripts, which is characterized by longer texts with fewer labels than in other NLP datasets. A two-step open-domain question answering approach is followed by using a Reducer to extract relevant text segments from a hearing and a Producer model to generate answers from the text segments selected by the Reducer. A combination of a rule-based Reducer and a neural Producer show to yield improved performance. In particular, as default Producer, a combination of RoBERTa and BigBird base is chosen due to its balance of long input length, low computation requirements, and performance on different choices of prediction heads, namely classification head, MLM head and QA head.

[Trias et al., 2021] propose an ensemble language model consisting of a combination of a Transformer architecture with a finite state machine to extract names from American English documents.[93] This exploits one pre-trained general-purpose English language NER model based on Flair[94] trained on CoNLL03 data, and another one trained on Harvard Caselaw Access Project, using GLoVe embeddings and XLM-RoBERTa.

[Shaheen et al., 2020] focus on the problem of large multi-label text classification (LMTC) in the legal domain, using JRC-Aquis and EURLEX57K annotated with the EuroVoc labels. In LMTC, the label space is comprised of thousands of labels, typically following a power-law label distribution and hierarchically organized. Pre-trained BERT, RoBERTa, DistilBERT, XLNet and m-BERT are applied on this task, using various training strategies such as gradual unfreezing[95] and slanted triangular learning rates,[96] in addition to fine-tuning. Results indicate that DistilBERT is better in retrieving the most probable label compared with RoBERTa and BERT. Also in the LMTC context, [Chalkidis et al., 2019b] apply a number of methods, including BERT, to investigate which portions of the documents in EURLEX57K are more informative. Results show that the title and recitals of each document contains enough information (and also allows to overcome the BERT's maximum text length limit); however, the approach fails in zero-shot learning.

[Sarkar et al., 2021] define a triplet network for legal sentence classification. A triplet network consists of three instances of the same neural network with shared parameters, it takes as input three objects, i.e., the positive example and the anchor, which belong to the same class, and the negative examples, which belongs to a different class, and it outputs the distance between the anchor and the positive example and the distance between the anchor and the negative example. The representation of the anchor, positive and negative sentences are then used to compute the triplet loss function to minimize. The network encodes each input sentence using Sentence-BERT as encoder, and the contextual sentence embedding is then fed to a two-layer perceptron. For the downstream classification task, a SVM with Radial Basis Function (RBF) kernel is used to compute the probability that an input sentence is promissory; the sentence is eventually classified as either promissory or non-promissory depending on a user-specified threshold on the SVM output probability.

[Ahmad et al., 2021] deal with fine-grained information extraction of privacy policies, specifically to identify sentences expressing privacy practices (intent classification task) and to detect specific details as text spans into the sentences (slot filling task). Two alternative approaches are proposed to tackle the two tasks: modeling intent classification and slot tagging either jointly as a sequence tagging task, or separately by generating respectively labels and spans. For the first approach, BERT, RoBERTa, a Transformer model along with a BiLSTM encoder are trained to get contextual representations of the input; in particular, the embedding of the special token [CLS] is used by a softmax classifier to predict the target intent, while the embeddings of the input tokens are used by another softmax classifier to predict the slot labels combined with a conditional random field (CRF). Results show that BERT-based and RoBERTa-based sequence tagging models outperform the other baselines by a wide margin, and the use of CRF is benefical for the slot tagging task but slightly degrades the performance for intent classification. For the second approach, MiniLM, UniLM, UniLMv2, MASS and BART are fine-tuned on the benchmark, and results show that seq2seq models outperform the sequence tagging models in the slot filling task, with BART and UniLM-based models reaching best performances, while they have similar results on the intent classification task.

---

[92] `https://github.com/rubenkruiper/SPaR.txt`

[93] `https://harvard-almit.github.io/legal-nlp`

[94] https://github.com/flairNLP/flair

[95] The training process is divided into multiple cycles, each consisting of several training epochs. Training starts after freezing all layers except for the last few layers in the first cycle one, then more layers are unfrozen gradually (from last to first layers) during later cycles.

[96] The learning rate is increased at the beginning of a training epoch up to the maximal learning rate, then slowly reduced to refine the parameters.

## 4.2 Enhanced methods for COLIEE Tasks

### 4.2.1 Case Law Retrieval (Task 1)

In the COLIEE-2021 competition, [Nguyen et al., 2021a] consider BERT as semantic model and Rank-BM25[97] as lexical model. In particular, the lexical model restricts the searching space of candidate cases to the top-100 cases for a given query, according to a lexical similarity score. After that, query and candidates are split into paragraphs and a lexical score matrix for each query-candidate pair is derived. In a similar way, a semantic score matrix for each query-candidate pair is derived using BERT, which is fine-tuned on a silver dataset based on the Task 1 raw data. Finally, the most relevant cases for the given query is obtained by combining lexical and BERT-based scores. The overall performance on the test set turns out to be quite poor, probably due to prior passage of searching space restriction performed by the lexical model.

In the previous edition of COLIEE competition, the same team uses a similar approach, where BERT is fine-tuned for a text-pair classification task with the support of specific heuristics to extract text-pairs, and BM25 is used as lexical model [Nguyen et al., 2020]. In that case, for each query the searching space consists of only 200 possible candidates, thus leading to significantly better results.

[Rossi and Kanoulas, 2019], [Li et al., 2021b], [Althammer et al., 2021], [Ma et al., 2021] and [Shao et al., 2020b] propose methods dealing with the document length limitation; therefore a discussion of such works is postponed to Section 4.4.

### 4.2.2 Case Law Entailment (Task 2)

One of the first approaches to case law entailment is provided by [Rabelo et al., 2019a], which has been described in Section 4.1. In the 2020 edition, [Rabelo et al., 2020] propose a Random Forest classifier to score each pair query-paragraph using features from the entailment score of BERT fine-tuned on the Task 2 data, BART and RoBERTa fine-tuned on a generic entailment dataset, BERT fine-tuned on paraphrase detection (zero-shot setting), cosine similarity between sentence embeddings of the input pair and cosine similarity on bag-of-words of the noun phrases contained in the input pair. Data augmentation based on back translation is also carried out, however without significant evidence of improvement.

[Alberts et al., 2020] exploit BERT in two ways to address the task in the COLIEE-2020 competition: the first one is to compute cosine similarity scores between BERT embeddings after fine-tuning on the COLIEE training data, and the second way is to derive features for a natural language inference (NLI) task by applying BERT on the NLI dataset. The two types of features, combined with BM25 based ones, are fed as input to a XGBoost-based classifier [Chen et al., 2015], which achieved the third place in the competition.

[Nguyen et al., 2020] experiment with the same solution proposed for Task 1 (described in Section 4.2.1) together with BM25 scores. This is also improved by fine-tuning BERT on the training data provided for this task. A further variant is proposed in which BM25 is replaced with BERT fine-tuned on SigmaLaw,[98] a corpus of legal cases including cases from the US Supreme Court. Among the three versions, the second one achieved the best F1 score reaching the leading position in the competition.

In the COLIEE-2021 competition, [Nguyen et al., 2021a] starts from their own approach implemented for Task 1 to define a binary classification task for BERT fine-tuned using both the silver dataset, obtained from the Task 1 data, and a further gold dataset obtained from the Task 2 data. One of their submissions is a combination of lexical score and a model trained on the *Next Foreign Sentence Prediction (NFSP)* task [Nguyen et al., 2021b], that is a binary classification task in which the text-pairs are composed of sentences with different languages and the model has to understand the correct meaning of each sentence to determine if their semantics belong to two consecutive sentences (cf. Section 4.7). However, best results are obtained using their earlier approach.

[Rosa et al., 2021] apply pre-trained Transformer models directly to the Task 2 of COLIEE 2021, without any preliminary fine-tuning on the legal task.[99] More specifically, they test monoT5 in two settings: zero-shot and fine-tuned on the Task 2 data of COLIEE 2020. In addition to monoT5, they also fine-tune DeBERTa on the legal task using the Task 2 data of COLIEE 2021. To balance the amount of negative and positive examples, they expand the positive examples generating a set of artificial fragments from the base case paragraphs through a sliding window strategy. As a further solution, an ensemble of monoT5 and DeBERTa, both fine-tuned on COLIEE 2020 data, is proposed. The models estimate a score for each fragment-candidate paragraph pair to select the best set of candidate paragraphs for each case fragment. A candidate paragraph is chosen if its score complies with specific rules. For the ensemble model, the selected paragraphs are the concatenation of the final set of DeBERTa and monoT5. The fine-tuned version of DeBERTa, monoT5 and the ensemble model achieve the first 3 positions in the competition, with the ensemble model on top position. However, once the ground-truth annotations of the test set have been released, the authors find out that monoT5 with zero-shot setting performs better than the single fine-tuned monoT5 and DeBERTa. This result leads the authors to the conclusion that, in case of limited labeled data, a model with no adaptions to the target domain can be more robust than fine-tuned models.

---

[97] https://pypi.org/project/rank-bm25/
[98] https://osf.io/qvg8s/
[99] https://github.com/neuralmind-ai/coliee

In the attempt of dealing with the text length limitation of BERT, [Shao et al., 2020b] consider either to truncate the input queries and paragraphs symmetrically, or to limit the tokens of the query to 128 and truncate the paragraph if the total tokens of query and candidate exceeds the limit. This is motivated by the observation that most of the decision fragments in the training data is no longer than 128 tokens. In both cases, the final hidden vector corresponding to the [CLS] token is fed to a fully connected layer for the classification. The results on test set demonstrate that the asymmetric truncation significantly improves the performance of the model. A third solution for the task proposed by the authors is to combine the output vectors of BERT (both with symmetric and asymmetric truncation), BM25 score and some structural features to be fed to a RankSVM model.[100]

Note that there have been other approaches, such as those proposed in [Li et al., 2021b, Kim et al., 2021], effectively dealing with the length of the documents; as such, they will be discussed in Section 4.4.

### 4.2.3 Statute Law Retrieval (Task 3)

In COLIEE 2020, [Nguyen et al., 2020] fine-tune `bert-base-uncased` on a text-pair classification task considering all query-article pairs. Due to the unbalancing between positive and negative pairs, they filter all the possible candidates to top-$k$ articles according to the cosine similarity calculated comparing the TF-IDF vectorization of article and query. They combine a `bert-base-uncased` model fine-tuned on text-pair classification and a variant of BERT, dubbed *BERT-CC*, previously fine-tuned on all data of the Civil Code with a masked language modeling task and further fine-tuned on text-pair classification. This is motivated by the assumption that BERT-CC can compensate for the lack of domain-specific knowledge of BERT, so that together they can learn different linguistic features. This hypothesis is confirmed by the results on the test set in the competition, where the ensemble model performed higher than the two separate models.

[Shao et al., 2020a] address the task as a binary classification problem using the Japanese version of the data provided by organizers. Each question is paired with every possible article and, due to the imbalance between positive and negative samples, the positive samples are oversampled up to 100 times. A Japanese version of BERT with whole word masking mechanism, called `BERT-base_mecab-ipadic-char-4K_whole-word-mask` (for short, *BERTjpcwwm*), and a Japanese version of AlBERT[101] are trained. In the first case, BERTjpcwwm is equipped with two settings, considering a maximum text length of 384 and 512, respectively. The relevance of an article for a query is decided through a threshold. The ensemble of these two BERTjpcwwm models and AlBERT model ranked first at COLIEE-2020.

For the 2021 edition, [Schilder et al., 2021] fine-tune a pre-trained Japanese BERT model on pairwise sequence classification task, where a text-pair is composed of queries and articles. The task is addressed by varying the BERT score along with the maximum cosine similarity on validation set and using cosine similarity to sort entailing articles. However, the proposed system produced unsatisfactory results, probably due to the lack of a sufficient number of training examples to adequately fine-tune BERT on COLIEE language domain.

[Wehnert et al., 2021] propose three approaches based on BERT model for the English version of the task. In the first approach, SBERT embeddings are combined with TF-IDF vectorization. To this end, the training data is increased with document enrichment strategies, in order to obtain article vectors that are as much unique as possible. In particular, each article can be enriched with hierarchical relations between articles as metadata, crawling the Japanese open source commentary on the article[102] and using the training data of Task 4. For each query-article pair, the cosine similarity scores are calculated ove TF-IDF and SBERT embeddings (`paraphrase-distilroberta-base-v1`). In the second approach, LEGAL-BERT@aueb is fine-tuned on sentence classification task. Data manipulation is performed in order to have an augmented dataset. The softmax score derived from the model is then combined to the cosine similarity calculated on TF-IDF vectorization (cf. Section 4.3). The third approach involves the use of BERTScore, which works as follows: the pairwise cosine similarities of all token-wise contextual embeddings from two sentences are first computed, then the token pairs between the two sentences which have the highest cosine similarity are selected, and those similarities are summed up and discounted by the words in the sentence to obtain precision, recall and the according F1-score. The first approach proved to be the best for the competition, which was ranked first at the 2021 competition.

---

[100] https://www.cs.cornell.edu/people/tj/svm_light/svm_rank.html
[101] https://github.com/alinear-corp/albert-japanese
[102] https://ja.wikibooks.org/

Note that [Aydemir et al., 2020] address the COLIEE-2020 task based on a multi-language approach, which is described in Section 4.7. Also, in Section 4.4, we discuss the COLIEE-2021 methods by [Nguyen et al., 2021a] and [Yoshioka et al., 2021b] dealing with the length limitation of BERT.

### 4.2.4 Statute Law Entailment (Task 4)

For the statute law entailment task, early Transformer-based solutions have been proposed in [Gain et al., 2019], [Westermann et al., 2020], and [Hudzina et al., 2020] (cf. Section 4.1).

In COLIEE 2020, [Alberts et al., 2020] use the last hidden layers `bert-base-cased` as the input for the XGBoost classifier, inspired by [Gain et al., 2019]. However, such a combination reaches poor performance, probably due to the fact that BERT was not trained on legal tasks. [Shao et al., 2020a] apply the same ensemble system used for Task 3, without however replicating the enhancement of the training samples, as for Task 4 the goal is to answer the queries based on the provided relevant articles. This led to lower performance of the model, which ranked sixth in the competition.

[Rabelo et al., 2020] address the statute law entailment problem using RoBERTa for natural language inference. They identify multiple NLI types in the statute law dataset, from which the model needs to recognize condition, conclusion and exceptional cases. To add more training data, they include the SNLI dataset [Bowman et al., 2015]. A pre-processing step is performed to help the detection of the different NLI types. This step consists in splitting the data in condition, conclusion and exceptional case, replacing references to prior cases or paragraphs with the referenced text, re-generating a sentence by negating the conclusion for the exceptional case and extracting one relevant sentence that matches most terms with the query. The model reaches the second position in the competition. The authors also adopt a BERT model for natural language inference to address Task 4 of COLIEE 2021 [Kim et al., 2021]. They select the most relevant candidate sentences for the query following their previous approach [Rabelo et al., 2020] and training a BERT classifier over triplets of query, article' sentence and binary label (yes/no). To help the pragmatic reasoning of BERT, the authors incorporate additional features consisting of the semantic category numbers of the Kadokawa thesaurus to the content words. The inclusion of such a semantic information boosts the performance of the model, reaching the fourth position in the competition for Task 4, while the same model without the additional information reached the third-to-last position.

[Nguyen et al., 2020] exploit the questions and the results of their system for the Task 3 (cf. Section 4.2.3) as input pairs for the classification prolem. On the development set, top-2 articles extracted by TF-IDF from the Civil Code are added to the gold data for each question. The answer for the question is positive if at least one pair is classified as positive by `bert-base-uncased`. As a second setting, only the content of the questions is used as input, following the approach they used in COLIEE 2019 [Rabelo et al., 2019b]. Essentially, the entailment problem is translated to a *lawfulness* classification task. BERT and BERTLaw (cf. Section 4.3) are trained with law sentences of the Civil Code and previous years' bar questions available from COLIEE. As shown in the final scores on test set, the second setting with BERTLaw reaches a score considerably higher than the other solutions proposed by the same team as well as by other teams that contributed to the competition.

[Schilder et al., 2021] define ensemble models based on ELECTRA and T5 to address the Japanese version of the statute entailment task. In the first case, a Japanese ELECTRA-*small* variant is chosen,[103] which turns out to be the best pre-trained architecture for Japanese according to the Jensen-Shannon divergence [Lin, 1991], and measures the embedding distribution distances between positive and negative examples. The model is trained layer by layer on the few training examples available. Although the model obtains strong performance in the validation step, it achieves a low accuracy on test set, probably due to a potential over-fitting on the available data. As regards T5, the best performing approach consists in fine-tuning on COLIEE dataset and evaluating T5 embeddings (the other two approaches consist in adding span corruption in the fine-tuning step and pre-training T5 embeddings from scratch, respectively). Besides ELECTRA and T5, the authors also investigate a multi-sentence natural language inference model, *Multee* [Trivedi et al., 2019]. The final submissions comprise ELECTRA, Multee and an ensemble of the first T5-based approach, ELECTRA and Multee. However, none of the three submissions obtained satisfactory results.

[Yoshioka et al., 2021a] address the Japanese statute law entailment task using BERT-based ensemble methods with data augmentation. In particular, they augment training data with a focus on the logical mismatches between articles and questions. The positive examples are composed of the same sentence used both as the query and as the article's sentence, while negative examples are pairs of original sentences and their versions with judicial decision flipped. If one article contains multiple decisions, then it is divided into smaller sentences containing one judicial decision, and if the split sentence expresses an exceptional case, then the judicial decision is flipped. Then, ten BERT-japanese models[104] are fine-tuned on this dataset to evaluate if the article entails a question or not. The models differ in the non-deterministic characteristics of fine-tuning process and the randomly selection of training set. Three ensemble models are then obtained by combining subsets of the aforementioned ten models. In the competition, they reach the best three accuracy scores, thus taking the top three positions.

---

[103] `https://huggingface.co/Cinnamon/electra-small-japanese-discriminator`
[104] `https://github.com/cl-tohoku/bert-japanese`

Further studies on Task 4 are [Nguyen et al., 2021a] and [Kim et al., 2021], which address the task by accounting for the document length limitation (cf. Section 4.4), [Aydemir et al., 2020], which propose a multi-language approach (cf. Section 4.7), and [Goebel et al., 2021], which address the task based on a domain-adaptive solution (cf. Section 4.3).

### 4.2.5 Statute Law Question Answering (Task 5)

[Schilder et al., 2021] propose two solutions for the Japanese version of Task 5: using DistilRoBERTa without any domain-specific training, and using GPT-3 [Brown et al., 2020b] with a few-shot learning. DistilRoBERTa is a distilled version of RoBERTa for paraphrase detection. The authors use the implementation[105] of SBERT containing a pre-trained DistilRoBERTa model trained on paraphrase text. The relevant articles to a given query are determined according to a similarity score. GPT-3 is a massive auto-regressive language model based on Transformer that obtained impressive results in generation and question-answering tasks with a few-shot learning scenario. The model is available via OpenAI interface[106] in different size versions. The authors use the largest model, dubbed *davinci*, and the fastest model, dubbed *ada*. From the experimental results, it is observed that GPT-3 models could not reach a sufficient level of comprehension on the legal domain, while the system based on DistilRoBERTa shows promising performance on the task.

[Kim et al., 2021] combine the output of a traditional TF-IDF technique with the NLI system described for the English version of Task 4 (cf. Section 4.2.4). As for Task 3, the additional semantic information enhances the model performance, allowing the team to achieve the second position at the competition.

[Yoshioka et al., 2021b] also consider the Japanese version of Task 5, although they could not meet the deadline for the submissions. Three configurations of ensemble models are defined by combining the submitted runs for the Task 3 [Yoshioka et al., 2021b] (cf. Section 4.2.3) and for Task 4 [Yoshioka et al., 2021a] (cf. Section 4.4), under the same conditions of the other participants. One of the three configurations, based on the best run of Task 4, obtained similar performance compared to the best model of the competition [Nguyen et al., 2021a]. Also, [Nguyen et al., 2021a] propose a multilingual approach for the task, therefore its discussion is referred to Section 4.7).

## 4.3 Domain-adaptive pre-training methods

In contrast to task-adaptive fine-tuning methods so far discussed, *domain-adaptive pre-training* allows for deeply tailoring an out-of-the-box model to a specialized language domain [Wang et al., 2020a], [Gururangan et al., 2020], [Song et al., 2022]. In the legal domain, there are two main strategies that stand as alternative to the direct application of a TLM for the downstream task, namely (i) to continue pre-training the model on a legal corpus, or (ii) to pre-train the model from scratch on a legal corpus. In this section, we discuss methods that developed TLMs based on one or both the pre-training strategies and compared their performance effects on various tasks.

On different tasks of document review, [Shaghaghian et al., 2020] propose to further pre-train and pre-train from scratch BERT models according to different strategies of tokenization and configuration of the initial weights of the model. More specifically, SentencePiece tokenization is performed on a corpus of US SEC legal agreements to generate the same number of cased tokens as in `bert-base-cased`; only 36% of tokens were found to be shared between tokens from the input legal corpus and the tokens in the original BERT. A hybrid tokenization is also devised whereby the 500 most frequent words in the input legal corpus are added to the token set if they do not occur as unbroken tokens in the set of general-domain tokens. Moreover, each of the BERT variants is also used as the teacher to customize as many DistilBERT models.

*BERTLaw* [Nguyen and Nguyen, 2021] is a model pre-trained on a collection of American legal cases with 8.2 million sentences (182 million words). BERTLaw has the same architecture and training objectives as BERT-base, however with an overlap between its vocabulary and the one in the original BERT-*base* model that is less than 50% of the size of the union of the two vocabularies. On the COLIEE-2019 Task 4, BERTLaw has shown to outperform BERT-*base* by almost 4% on validation data and over 16% on test data.

[Chalkidis et al., 2020b] are the first to propose both further pre-training (FP) and pre-training from scratch (SC) BERT on legal corpora. Their models, hereinafter referred to as *Legal-BERT-FP@aueb* and *Legal-BERT-SC@aueb*,[107] were trained on a collection of legal documents, including EU and UK legislations, ECJ cases, ECHR cases, US court cases, and US contracts. Legal-BERT-FP@aueb models are the result of prolonged in-domain pre-training of BERT-*base* up to 500K steps, in all or a selection of the training legal corpora, whereas Legal-BERT-SC@aueb has still the

---

[105] https://github.com/UKPLab/sentence-transformers
[106] https://openai.com/api/
[107] https://huggingface.co/nlpaueb

**Table 6.** Summary of main methods discussed in Sections 4.3 and 4.4.

| Method | Ref. | Downstream Tasks | Lang. | Data | Long docs? |
|---|---|---|---|---|---|
| Pre-trained BERT w/ legal or hybrid tokens; Pre-trained DistilBERT w/ legal or hybrid tokens | [Shaghaghian et al., 2020] | PR, NER, TM, SA | EN | US SEC legal agreements | No |
| From-scratch pre-trained BERT (BERTLaw) | [Nguyen and Nguyen, 2020] | COLIEE-2020 Task 4 | EN | civil code | No |
| Further pre-trained BERT (Legal-BERT-FP@aueb); From-scratch pre-trained BERT (Legal-BERT-SC@aueb) | [Chalkidis et al., 2020b] | TC, NER | EN | EURLEX57K; ECHR cases; CONTRACTS-NER | No |
| Further pre-trained BERT (Legal-BERT@jhu) | [Holzenberger et al., 2020] | CTR | EN | CAP cases | No |
| Further pre-trained BERT (Legal-BERT-FP@stanford); From-scratch pre-trained BERT (Legal-BERT-SC@stanford) | [Zheng et al., 2021] | QA, OR, ToS DS | EN | CaseHOLD; Overruling; Terms-of-Service | No |
| Fine-tuned Legal-BERT@aueb | [Mahari, 2021] | LPP | EN | CAP judicial opinions | No |
| Further pre-trained AlBERT (ALeaseBERT) | [Leivaditi et al., 2020] | red flag detection, entity extraction | EN | contracts | No |
| Further pre-trained BERT, SpanBERT and ELECTRA | [Chi et al., 2023] | intent classification, SF, QA, NER | EN | contracts | No |
| Fine-tuned Legal-BERT@aueb (InLegalBERT); Further pre-trained Legal-BERT@stanford (InCaseLawBERT) | [Paul et al., 2022b] | SAR, RRL, CJP | EN | legal cases | Yes |
| Fine-tuned BERT, Legal-BERT@aueb, Legal-BERT + BiLSTM and CRF | [Ostendorff et al., 2021] | DR | EN | Open Case Book; Wikisource US Courts | Yes |
| Fine-tuned BERT and LegalBERT@aueb + BiLSTM and CRF | [Bhattacharya et al., 2021] | RRL | EN | legal cases | No |
| Fine-tuned BERT, SBERT and LegalBERT@aueb | [Chalkidis et al., 2021b] | RIR | EN | EU2UK; UK2EU | No |
| Fine-tuned BERT (C-BERT) | [Chalkidis et al., 2021b] | EUROVOC concept classification | EN | EU2UK; UK2EU | No |
| Fine-tuned BERT + KERMIT | [Wehnert et al., 2022] | COLIEE-2021 Task 4 | EN | civil code | No |
| Fine-tuned BERT, RoBERTa, LegalBERT@aueb, DeBERTaV3, BigBird | [Wang et al., 2023] | MCRC | EN | contracts | Yes |
| Fine-tuned Legal-PEGASUS, Legal-LED, BART, BERT-BART; BERTSUM | [Shukla et al., 2022] | ES, AS | EN | legal cases | Yes |
| | | | | | |
| Fine-tuned BERT w/ summarized texts | [Rossi and Kanoulas, 2019] | COLIEE-2019 Task 1 | EN | legal cases | Yes |
| Two-stage fine-tuned BERT w/ attentive aggregation (HIER-BERT) | [Chalkidis et al., 2019a] | AV, CIR | EN | ECHR cases | Yes |
| Two-stage fine-tuned BERT w/ attentive aggregation (BERT-PLI) | [Shao et al., 2020c] | COLIEE-2019 Tasks 1, 2 | EN | legal cases | Yes |
| LMIR + fine-tuned BERT-PLI | [Shao et al., 2020b] [Ma et al., 2021] | COLIEE-2020 Task 1 | EN | legal cases | Yes |
| Fine-tuned BERT, ALBERT, RoBERTa, DeBERTa | [Hendrycks et al., 2021] | QA | EN | CUAD legal contracts | Yes |
| Fine-tuned BERT and SRoBERTa | [Aumiller et al., 2021] | STP | EN | ToS documents | Yes |
| Transformer summarizer + fine-tuned BERT | [Kim et al., 2021] | COLIEE-2021 Task 2 | EN | legal cases | Yes |
| Fine-tuned BERT and RoBERTa | [Nguyen et al., 2021a] | COLIEE-2021 Task 3 | JP | civil code | Yes |
| Hier-BERT variation based on LegalBERT@aueb | [Chalkidis et al., 2021c] | rationale extraction | EN | ECtHR | Yes |
| Pre-trained DPR (LawDPR); Fine-tuned Longformer-Encoder-Decoder | [Althammer et al., 2021] | COLIEE-2021 Tasks 1, 2 | EN | Canada case laws | Yes |
| Fine-tuned BERT (BERTSum) | [Deroy et al., 2021] | ES | EN | legal cases | Yes |
| Fine-tuned BERT, RoBERTa, DistilBERT | [Klaus et al., 2022] | ES | EN | EUR-LexSum | Yes |
| Further pre-trained BERT and RoBERTa | [Lam et al., 2020] | employment notice prediction | EN | notice cases | Yes |
| Fine-tuned BERT, Legal-BERT@aueb and Legal-BERT@stanford; Fine-tuned BigBird and LongFormer | [Limsopatham, 2021] | AV, OR | EN | ECHR Violation dataset; Overruling Task dataset | Yes |
| From-scratch pre-trained BERT + Siamese Legal BERT + Fine-tuned Legal BERT | [Khazaeli et al., 2021] | PR, QA | EN | caselaw headnotes, RFCs | No |
| Further pre-trained BERT (BERT-Legal) + LSTM | [Li et al., 2021b] | COLIEE-2021 Task 1 | EN | legal cases | Yes |
| Fine-tuned BERT-Legal w/ augmentation; Fine-tuned BERT-Legal w/ fast gradient; | [Li et al., 2021b] | COLIEE-2021 Task 2 | EN | legal cases | Yes |
| Fine-tuned Legal-BERT@aueb + TF-IDF vectors | [Furniturewala et al., 2021] | AILA-2021 Tasks 2a-b | EN | legal cases | Yes |
| Fine-tuned Legal-BERT@aueb | [Jain et al., 2021] | AILA-2021 Tasks 2a-b | EN | legal cases | Yes |
| Ensemble of BM25 w/ KLI summarizer and fine-tuned BERT w/ LED summarizer | [Askari and Verberne, 2021] | COLIEE-2020 Task 1 | EN | legal cases | Yes |
| Fine-tuned BERT (Span NLI BERT) | [Koreeda and Manning, 2021] | NLI, span detection | EN | contracts | Yes |
| Fine-tuned BART, PEGASUS, LED and PRIMERA | [Shen et al., 2022] | AS | EN | legal cases | Yes |
| BERT + Bi-LSTM + CRF (SciBERT-HSLN); Fine-tuned BERTSUM + rhetorical roles (BERTSUM RR) Legal-PEGASUS + rhetorical roles (Legal-PEGASUS RR) XLNet + rhetorical roles (BERTSUM RR) | [Kalamkar et al., 2022] | RRL, ES, AS, LJP | EN | legal cases | Yes |
| BERT, RoBERTa, DeBERTa, Longformer, BigBird, Legal-BERT@aueb and Legal-BERT@stanford | [Chalkidis et al., 2022b] | AVP, ALVP, TpC, QA | EN | legal cases, contracts | Yes |
| Fine-tuned Longformer (Longformer-8192 & variants), Fine-tuned LegalBERT@aueb + Longformer (LegalLongformer & variants) Fine-tuned LegalBERT@aueb+TFIDF (TFIDF-SRT-LegalBERT) | [Mamakas et al., 2022] | AVP, ALVP, TpC, TC, QA | EN | legal cases, contracts | Yes |
| Fine-tuned Hierarchical Attention Transformer (HAT) | [Chalkidis et al., 2022a] | ALVP, TC, NLI | EN | legal cases, contracts | Yes |
| Fine-tuned BERT, DistilBERT, RoBERTa and XLNet | [Malik et al., 2021] | CJPE | EN | legal cases | Yes |

same architecture but with a new vocabulary equivalent in size to BERT-base. Also, a smaller Legal-BERT-SC@aueb model is derived with 6 layers, 512 hidden units, and 8 attention heads (35M parameters, 32% the size of BERT-base). All models were trained over the legal corpora in batches of 256 samples, including up to 512 sentence-piece tokens, with Adam optimizer and learning rate of 1e-4, as in the original BERT; nonetheless, for the fine-tuning stage, early-stopping was used instead of a fixed number of epochs, and more batch sizes, learning rates and drop-out regularization rate than the recommended ones in [Devlin et al., 2019] were experimented. Legal-BERT-SC@aueb has shown to achieve a significantly lower training loss than the smaller model, while the latter turns out to be comparable to the original BERT-base. As for the Legal-BERT-FP@aueb models, the training loss can actually be improved when further pre-training on some selected corpora (but not on the whole collection of corpora). Considering the downstream tasks (EurLex57K, ECHR-Cases, Contracts-NER), Legal-BERT-FP@aueb and Legal-BERT-SC@aueb models outperform BERT-*base* and are as good as or better than the fine-tuned BERT-*base*, with more substantial gain in the ECHR-Cases multi-label task. Legal-BERT-FP@aueb and Legal-BERT-SC@aueb models also tend to be comparable to each other on all but the ECHR-Cases tasks, where Legal-BERT-SC@aueb is clearly better. The small Legal-BERT-SC@aueb is highly competitive with at least one of the larger variants in most tasks, suggesting that architecturally complex model may not be necessary when dealing with domain-specific sub-languages.

A legal BERT model is also developed by [Holzenberger et al., 2020], in a context of evaluation of statutory reasoning (i.e., how to reason about an example case, based on rules extracted from statutes), particularly for US tax regulations. This model, hereinafter referred to as *Legal-BERT@jhu*,[108] is obtained by further pre-training `bert-base-cased` on a portion of the case.law corpus, including both state and federal cases, for a total of 900M tokens. On the legal questions and answers of the corpus, Legal-BERT@jhu shows better perplexity than `bert-base-cased` (i.e., 2.7 against 14.4), which indicates improved adaptation to the legal queries. Legal-BERT@jhu was fine-tuned on the task of recognizing legal terms in the test set split of case.law, i.e., tokens or multi-token collocations that are defined in Black's Law Dictionary, a well-known legal dictionary.

[Zheng et al., 2021] further investigate the use of BERT-based methods to pre-train TLMs on legal documents, starting from the premise that a substantial gain from law-specific pre-training might be prevented by fine-tuning tasks that are too easy and/or based on data that are inconsistent with the pre-training corpus. The proposed models, here dubbed *Legal-BERT@stanford*[109] are pre-trained on the Harvard Law case corpus (i.e., case.law), where a random sample of 10% of decisions was extracted from the corpus and used as holdout set. More specifically, the variant Legal-BERT-FP@stanford is derived from the BERT-base model pre-trained for an additional 1M steps using the case.law corpus and the same vocabulary as BERT (uncased), while the variant Legal-BERT-SC@stanford is pre-trained from scratch for 2M steps using the case.law corpus, with a custom legal domain-specific vocabulary created using SentencePiece tokenization. Both variants use the same training objectives as BERT, i.e., MLM and next sentence prediction; for the latter, regular expressions are used to ensure that legal citations are included as part of a segmented sentence. Through a number of legal tasks varying in difficulty and domain-specificity, BERT has shown to already achieve high performance for easy tasks, while legal-specific pre-training takes small, resp. significant, advantage for mid-difficulty tasks that are not highly domain-specific, resp. high-difficulty and domain-specific tasks.

[Leivaditi et al., 2020] further pre-train AlBERT on lease contracts. The resulting model, named *ALeaseBERT*, is fine-tuned and evaluated on the tasks of red flag detection and entity extraction. In particular, for the red flag detection, the additional pre-training proves to enhance the performance not only compared to lexical competitors, but also against AlBERT pre-trained from scratch as well as a fine-tuned AlBERT on the task.

[Chi et al., 2023] further pre-train BERT, SpanBERT and ELECTRA on a pre-training corpus for privacy policy understanding provided in PLUE (cf. Section 3.4); the resulting pre-trained models are then fine-tuned on the benchmarks included in PLUE. Results reveal the advantage of these further pre-trained models compared to their only task-adapted counterparts (i.e., fine-tuned on the PLUE benchmarks). Moreover, the pre-trained models show comparable results w.r.t. a task-adapted Legal-BERT@aueb.

[Paul et al., 2022b] further pre-train Legal-BERT@aueb and Legal-BERT@stanford on a corpus of about 5.4M Indian court cases. The resulting models, called InLegalBERT (derived from Legal-BERT@aueb) and InCaseLawBERT (derived from Legal-BERT@stanford), are evaluated on three downstream tasks: legal statute retrieval, rhetorical role labeling, and court judgment prediction. To this regard, the two proposed models have been fine-tuned and assessed on various benchmarks against BERT, the original Legal-BERT@aueb and Legal-BERT@stanford and the best performers according to the benchmark's paper. In particular, for the legal statute retrieval task, [Paul et al., 2022b] consider the ECtHR Task B of LexGLUE and a dataset of criminal case documents provided in [Paul et al., 2022a]. A hierarchical model — which consists of a lower-level encoder to provide the [CLS] representation of each sentence in the input text, followed by an higher-level encoder that aggregates the sentence embeddings to get the document representation — is used to get the relevant statutes, in a multi-label classification manner. Five versions of this model are defined, choosing the lower-level encoder among BERT, Legal-BERT@aueb, Legal-BERT@stanford, InLegalBERT and InCaseLawBERT, whereas the high-level encoder is based on a LSTM with the attention mechanism. For the rhetorical role labeling task, the models are evaluated on a dataset of 50 documents from the Supreme Court of India [Bhattacharya et al., 2019b] and a dataset of as many documents from the Supreme Court of UK [Bhattacharya et al., 2021]; a hierarchical model is again used, with the same variations of the lower-level encoder as in the previous task. For court judgment prediction, the authors consider the ILDC$_{multi}$ dataset and the hierarchical BERT model provided by [Malik et al., 2021], again with variations as above. Results show that both InLegalBERT and InCaseLawBERT improve the performances against LegalBERT@aueb and LegalBERT@stanford, respectively. Moreover, both InLegalBERT and InCaseLawBERT outperform the best performer of [Paul et al., 2022a] and [Malik et al., 2021], and InLegalBERT is better than the best performer of [Bhattacharya et al., 2021], ECtHR Task B [Chalkidis et al., 2022b]. Overall, results suggest that further pre-training on Indian legal data can improve the performance of a model in several legal tasks across domains and countries.

**Further applications of legal pre-trained models.** The above discussed legal pre-trained models have recently attracted attention, especially Legal-BERT@aueb models which have started to be widely used.

---

[108] `https://nlp.jhu.edu/law`

[109] `https://github.com/reglab/casehold`

In [Mahari, 2021], Legal-BERT-SC@aueb is fine-tuned for the legal precedent prediction (LPP) task, i.e., to predict relevant passages from precedential court decisions given the local context surrounding the passage in the opinion citing it (cf. Section 3.3). Using the US federal judicial opinions from the Case Law Access Project (CAP), the neighborhood of a passage is here comprised of 300 characters before the passage, 300 characters after the passage, 300 characters from the Introduction and 300 characters from the Conclusions. In 96% of unseen test examples, the correct passage is found to be within the top-10 predicted passages.

[Ostendorff et al., 2021] compare Legal-BERT@aueb and Legal-BERT@jhu — along with BERT (base and large), RoBERTa, SBERT and SRoBERTa, Longformer — against word-vector- and citation-based approaches on legal document recommendation as a text similarity task.[110] A legal document is represented as a numerical vector, and the top-5 documents with the highest cosine similarity are selected through nearest neighbor search. Results have shown no absolute winner across all datasets and criteria, although a legal-domain adaptation of the FastText word embedding model [Bojanowski et al., 2017,Joulin et al., 2017] was found the best performing method on average.

[Bhattacharya et al., 2021] address the problem of rhetorical role labeling and propose different deep learning models to identify seven rhetorical classes over documents from five different law domains and from two different jurisdictions. BERT and LegalBERT@aueb are fine-tuned on this task and used as bottom layers of a BiLSTM-CRF-based architecture. More specifically, the sentence embeddings produced by BERT and LegalBERT are used with a Hierarchical BiLSTM classifier and a Hierarchical BiLSTM CRF classifier with or without attention mechanism. Results show different outcomes in the model comparison, and in particular BERT and LegalBERT embeddings reveal to be beneficial for a Supreme Court of the UK dataset, but not for a Supreme Court of India dataset, compared to *Sent2Vec* [Pagliardini et al., 2018] embeddings. (Sent2Vec is a simple unsupervised model that considers sentence embeddings as the average of source word embeddings, considering both unigrams and n-grams).

[Chalkidis et al., 2021b] address the task of regulatory information retrieval, where the query is a document (e.g, an EU directive) and the goal is to retrieve a set of relevant documents from the available law corpora. Upon two new datasets, EU2UK and UK2EU, a two-stage approach is defined, where document re-ranking is carried out after pre-fetching, i.e., retrieval of the top-$k$ prominent documents for a query. As retrieval models, BERT, SBERT and LegalBERT@aueb are applied but showed to perform worse or comparably to BM25. As an improvement, following [Chalkidis et al., 2019b], BERT is fine-tuned to predict EUROVOC concepts that describe the core subjects of each text: the resulting model, called C-BERT, turns out to be the best pre-fetcher by a large margin in EU2UK, while being comparable to BM25 in UK2EU; also, an ensemble combining C-BERT with BM25 further improves the results. For document re-ranking, several neural re-ranking methods are selected to yield a relevance score for each of the top-$k$ documents returned by the best pre-fetcher: DRMM [Guo et al., 2016], PACRR [Hui et al., 2017], BERT-based re-rankers. DRMM is a deep matching model focused on relevance matching in ad-hoc retrieval. PACRR is a IR model that focuses on preserving positional information by incorporating domain insights. [Chalkidis et al., 2021b] use DRMM and PACRR on top of BERT embeddings, and the resulting methods are called C-BERT-DRMM and C-BERT-PACRR. A major finding is that the re-rankers fail to improve the performance w.r.t. the ensemble pre-fetcher, as their term matching mechanisms tend to degenerate and over-utilize the pre-fetcher score.

For the statute law retrieval task of COLIEE 2021, [Wehnert et al., 2021] fine-tune an ensemble model of Legal-BERT@aueb (`legal-bert-base-uncased`) and TF-IDF on sentence classification, to predict the relevance of an article to a query. Training and validation data are handled in three settings: in the original form (i.e., pairs of queries and relevant articles), decomposing instances with multiple relevant articles, and performing a data augmentation strategy to reduce data imbalance. In particular, the decomposed pairs are obtained separating the multiple relevant articles in different instances, producing a query-article pair for each relevant article. Data augmentation is performed by retrieving the top-50 non-relevant articles according to the highest cosine similarity between TF-IDF vectors of relevant articles and all the non-relevant articles, analogously to the approach in [Nguyen et al., 2020] to reduce the unbalance between positive and negative samples. The resulting dataset is fed to Legal-BERT@aueb and TF-IDF, to obtain a softmax score of relevance and cosine similarity score, respectively. The overall score is computed as the average of the two scores after normalization. The top-n relevant articles are filtered using a threshold based on the precision-recall trade-off in the validation set.

The same team also applied for the Task 4 of COLIEE 2021 [Goebel et al., 2021], proposing two approaches: SBERT combined with a *Graph Neural Network* (GNN) model and SBERT combined with LegalBERT@aueb. In the first approach, GNN is expected to model the entailment relationships between nodes, where nodes represent articles/queries. The result is a bi-partite graph, in which queries can have connections to multiple relevant articles, and articles are enriched with metadata from the section titles. From the content of articles and queries, contextual sentence embeddings are generated using SBERT and used as node features encoding external knowledge. Such node features are used to get query node embeddings following [Morris et al., 2019], which encodes information taking into account also relevant articles as direct neighbor nodes. The query node embeddings are then used for query node classification to address the entailment task. In the second approach, the team adapt the LegalBERT@aueb encoder they proposed for Task 3 [Wehnert et al., 2021], re-initializing the classification head with new trainable parameters and splitting training data instances having multiple articles for a query in order to obtain additional instances. From the competition results, LegalBERT@aueb submissions outperform the GNN-based approach; according to the authors, it is likely that the GNN was unable to generalize as well as LegalBERT@aueb due to the limited amounts of data and the lack of document enrichment to encode external knowledge. [Wehnert et al., 2022] further propose a new method for the COLIEE-2021 Task 4, which combines BERT and KERMIT [Zanzotto et al., 2020], an encoder that provides embeddings of syntactic parse trees. The idea is to inject further linguistic knowledge from the syntactic parse trees of query and articles. The output is the concatenation of the two embedding vectors. The trees for query and articles are obtained using the parser from the Stanford CoreNLP, as in [Zanzotto et al., 2020]. The concatenation of query and article is given to `bert-base-uncased` to get the sequence representation through the [CLS] resulting embedding. Both the embeddings of KERMIT and BERT are then passed to a fully connected decoder layer to get the entailment label. This third approach to Task 4 reaches higher performance than the GNN-based run and comparable results with the best LegalBERT@aueb-based run.

---

[110] https://github.com/malteos/legal-document-similarity

[Wang et al., 2023] address the multiple-choice reading comprehension task of MAUD by fine-tuning several TLMs, in particular BERT, RoBERTa, LegalBERT@aueb, DeBERTaV3 and BigBird for a single-task setting, and RoBERTa, LegalBERT@aueb and DeBERTaV3 for a multi-task setting, which requires to fine-tune a model for all questions of MAUD. Newer and larger models provided better results, with BigBird and LegalBERT@aueb being the best performers in the single-task and multi-task setting, respectively.

[Shukla et al., 2022] evaluate different extractive and abstractive summarization methods on their own collected datasets (cf. Section 3.4): domain-agnostic methods, such as SummaRunner [Nallapati et al., 2017] and BERTSUM for extractive summarization, BART for abstractive summarization; domain-adapted methods, such as CaseSummarizer [Polsley et al., 2016] and Gist [Liu and Chen, 2019] for extractive summarization, Legal-PEGASUS[111] for short abstractive documents, Legal-LED[112] for long abstractive documents; task-adapted methods, in which Transformers are fine-tuned on data generated with sentence similarity techniques. In addition, a newly hybrid extractive-abstractive method, called *BERT_BART*, is introduced and evaluated in both domain-adapted and task-adapted versions. As mentioned in Section 3.4, methods are to be evaluated w.r.t. document-level summaries, how well the summary represents the logical rhetorical segments in the legal case document, i.e., segment-wise evaluation, and how the summaries are evaluated by domain-experts. As regards the document-level evaluation, on the IN-Ext dataset, BERTSUM turns out to be the best extractive method, while the task-adapted Legal-PEGASUS and BERT-BART obtain the highest ROUGE and BERTScore values, respectively; on the IN-Abs dataset, the task-adapted Legal-LED, BART and Legal-PEGASUS obtain the best BERTScore, R-1 ROUGE and R-L ROUGE scores, respectively; on the UK-Abs dataset, the task-adapted BART is the best performer in terms of R-L ROUGE and BERTScore. As regards the segment-wise evaluation for the IN-Ext and UK-Abs datasets, results show a mixed picture, with Transformers obtaining the best score in most metrics; however, a discrepancy is noticed between the metrics used for summarization and the expert judgments: in fact, according to the involved law experts, extractive summaries are more satisfactory than the abstractive counterparts, since abstractive summaries are considered to be less organized and often incomplete.

## 4.4 Dealing with long legal documents

As discussed in Section 2, BERT and its early derivations can process texts having 512 tokens as maximum length. Input truncation to the maximum length has often been performed, which also holds for the methods discussed in the previous sections. However, this choice is clearly not optimal, especially when important information are spread through the whole document in different locations, which is also the case for legal documents. In this regard, text segmentation and text summarization can certainly be helpful.

In this section, we discuss methods that adopt different strategies for dealing with long documents. We organize this presentation by distinguishing between task-adaptive fine-tuning methods and domain-adaptive pre-training methods.

### 4.4.1 Task-adaptive methods

[Rossi and Kanoulas, 2019] address the COLIEE-2019 Task 1 as a ranking problem formulated as a pairwise relevance classification task, given the binary labels "Noticed" or "Not noticed". This is treated as a downstream task for the pre-trained `bert-base-uncased`, where the hidden state of the [CLS] token is used as the pairwise embedding for a pair of summarized query case and candidate case, and the TextRank method [Barrios et al., 2016] is used for the summarization steps. The model has shown good generalization ability, although the widespread distribution of scores for the positive class and the small proportion of pairs with high score for the negative class might affect negatively the performance of the system.

---

[111] `https://huggingface.co/nsi319/legal-pegasus`
[112] `https://huggingface.co/nsi319/legal-led-base-16384`

[Chalkidis et al., 2019a] face the length issue in legal cases proposing *Hier-BERT*. BERT is used to produce fact embeddings, then a self-attention mechanism applies to the fact embeddings and generates a single case embedding that is fed to a task-specific output layer. Three tasks are evaluated: binary violation classification, multi-label violation classification, and case importance prediction. The model is tested on ECHR (cf. Section 3.4). The list of facts in the fact description of each case are extracted using regular expressions. The violated articles are mapped to the violating case. An importance score is assigned for each case by the European Convention of Human Rights. The authors demonstrate that Hier-BERT significantly improves BERT in the aforementioned tasks.

*BERT-PLI* [Shao et al., 2020c] utilizes BERT to model paragraph-level interactions for legal case retrieval tasks, using a cascade framework.[113] Initially, BM25 is used to filter the set of candidates and BERT is fine-tuned for sentence-pair binary classification on the COLIEE-2019 Task 2 data. The fine-tuned parameters are kept for the subsequent stage whose goal is to train the BERT-PLI model for relevance prediction. BERT-PLI receives a query and a selected document each broken into paragraphs. Each pair of query paragraph and document paragraph is an input to the fine-tuned BERT. The final hidden state vector of the [CLS] token is considered as the aggregate representation of the input pair. For each query and document, an interaction map of paragraph is finally produced. Then, for each query paragraph, a max-pooling strategy is used to capture the strongest signal with a candidate document. The ordered sequence of vectors resulting from the max-pooling on all paragraphs of the query is encoded by a RNN. An attention mechanism is applied to infer the importance of each position, and a document-level representation of the query-document pair is obtained via attentive aggregation. The resulting representation is provided to a fully connected layer with *softmax* to make a prediction of binary relevance. BERT-PLI has shown to achieve better recall and F1-score than the top-2 teams at COLIEE-2019.

To address the binary classification COLIEE-2020 Task 1, [Shao et al., 2020b] apply a cascade framework in which a bi-gram LMIR [Song and Croft, 1999] is used to select top-$N$ candidates to be further classified by BERT-PLI. The proposed system focuses mainly on the semantic understanding of documents, so that an additional solution is proposed to combine the semantic aspect of BERT-PLI with the exact matching among documents applied by a word-entity duet framework [Xiong et al., 2017]. In particular, several features are extracted from BERT-PLI, BERT fine-tuned on sentence-pair classification using the first paragraph of query and candidate case, word-entity duet and similarity scores based on words and entities. A RankSVM model is then used to re-rank the top-$N$ candidates, based on the aforementioned features. The resulting system performs better than the single BERT-PLI based solution, suggesting that the union of semantic understanding and exact matching can be beneficial. Analogously, in COLIEE 2021, [Ma et al., 2021] apply BERT-PLI in Task 1 with some revisions. First, a LMIR model [Ponte and Croft, 1998] is used to filter candidates, then BERT is fine-tuned for a next sentence prediction task on the COLIEE-2019 Task 2. While [Shao et al., 2020c] compute interactions with all query paragraphs and all top-$N$ candidate paragraphs, [Ma et al., 2021] consider only query paragraphs with sentences with a citation token, i.e., sentences with placeholders (for example '*REFERENCE_SUPPRESSED*') that indicate the presence of mentions to other noticed cases. They also limit the number of paragraphs for a document. The proposed model outperformed the BERT-PLI based solutions and reached the first position.

[Hendrycks et al., 2021] fine-tune and evaluate BERT, RoBERTa, AlBERT and DeBERTa on the Contract Understanding Atticus Dataset (CUAD, see Section 3.4) for legal contract review, which is composed of contracts annotated with 41 category labels by legal experts. To deal with long documents, a sliding window mechanism is used over each document. DeBERTa turns out to perform the best over the other models, however the overall performances show that this is a challenging task, where the size of a model and the amount of training data play a very important role.

[Aumiller et al., 2021] propose an approach to document segmentation of legal text to predict topical coherence among paragraphs and to detect topical change (cf. Section 3.2).[114] The approach consists in fine-tuning RoBERTa and Sentence-RoBERTa on a topic similarity task, which is formulated as a binary classification problem, i.e., to detect if paragraphs belong to similar topics or not. To this purpose, a dataset of Terms-of-Service documents, dubbed ToS (cf. Section 3.4), is created which includes hierarchical topic annotations. The classification task is performed under the assumption of topical independence of the context, i.e., the topic probability of a paragraph is not affected by the belonging of context paragraphs to the same topic. The fine-tuned models are then evaluated for sequential inference on entire documents, where segments are delimited by topical change. The approach has shown to outperform segmentation baselines based on TF-IDF and bag-of-words models.

For the COLIEE-2021 Task 2, [Kim et al., 2021] fine-tune BERT on a binary classification problem using pairs of query and candidate paragraphs from the relative COLIEE dataset. To address the BERT token limit, a Transformer-based model generates a summary of the input. A subsequent phase of post-processing limits the maximum number of outputs for a given case. The final system is submitted to the competition in three runs, which differed in the setting of post-processing parameters.

[Nguyen et al., 2021a] split the articles in multiple chunks using sliding windows to address the text length limitation of BERT and RoBERTa for the COLIEE-2021 Task 3. This is motivated from the observation that only some parts of an article entail the corresponding query. Training data are built as pairs of query and entailing article (positive sample) or any other article (negative sample), limiting the maximum number of negative samples. To avoid noisy training examples, a self-labeling technique is exploited, consisting of multiple fine-tuning phases in which training data labels are modified according to specific rules [Triguero et al., 2015]. Different models, such as `bert-base-japanese`, `bert-base-japanese-whole-word-masking`, and `xlm-roberta-base`,[115] are trained on a sentence-pair classification task using the COLIEE-2020 dataset. The ensemble of those models performed better than the individual ones, reaching the second position in the competition. The same approach is also adapted for Task 4, using `bert-base-japanese-whole-word-masking`.

---

[113] https://github.com/ThuYShao/BERT-PLI-IJCAI2020
[114] https://github.com/dennlinger/TopicalChange

For the Japanese version of COLIEE-2021 Task 3, [Yoshioka et al., 2021b] improve their previous proposal, introduced in COLIEE 2020, based on the ensemble of a BERT-based retrieval system [Sakata et al., 2019] and the keyword-based Indri retrieval system.[116] A new article database is built combining articles and referred articles. To deal with long article sentences, the articles composed of a list of conditions or multiple judicial decisions are divided into small sequences through the numbered paragraph structure. Oversampling and negative sampling strategies are also applied to the training examples. The BERT-based and Indri scores for each article are combined to get an overall score. Three versions of the system are defined using different ensemble settings: the first version consists of BERT with oversampling, BERT with negative sampling and Indri, the second version uses BERT with oversampling only, and the third version corresponds to BERT with oversampling and Indri.

[Chalkidis et al., 2021c] propose a method to learn automatically extracting paragraphs (i.e., the rationales) of the input so to justify decisions in alleged violation prediction. To this end, the ECtHR dataset (cf. Section 3.4) is used as it includes annotations for paragraph-level rationales. Extraction of the rationales is driven by rationale constraints. The proposed method is a variation of Hier-BERT [Chalkidis et al., 2019a], dubbed HierBERT-HA, following the framework of [Lei et al., 2016] for the construction of rationales and consisting in three sub-networks: the first reads the text, the second extracts rationales through hard masking mechanism, and the third classifies the hard-masked text. More specifically, the paragraphs are first represented by the [CLS] embeddings obtained by LegalBERT@aueb [Chalkidis et al., 2020b]. Then, a two-layer Transformer [Vaswani et al., 2017] obtains a contextualized representation of the paragraph embeddings which are fed to two separated fully-connected layers, the one producing paragraph encodings for the classifier sub-network, and the other one producing paragraph encodings for the rationale extraction sub-network. The rationale sub-network obtains a binarized attention score for each paragraph. Each paragraph is then hard-masked using the corresponding attention score. The resulting paragraphs are then fed to the classifier sub-network to predict the alleged violated articles. Different versions of the model using rationale constraints as regularizers (such as sparsity, comprehensiveness, singularity) are compared. A HierBERT-HA version without hard masking, dubbed HierRBER-ALL, and rationale constraints is tested too. Results show that models with hard attention and rationale constraints perform well and comparably to HierBER-ALL.

[Althammer et al., 2021] propose a two-stage pipeline for case law retrieval followed by re-ranking.[117] For the first stage, each query case and the cases in the corpus are split into their paragraphs, so that relevant prior cases can be retrieved for each paragraph of the query case based on the relevance of their paragraphs. Both lexical and semantic retrieval models are exploited, where the first is based on BM25 and the second corresponds to DPR based on two BERT-base-uncased Siamese Encoders (i.e., the one for encoding the query passage and the other one for the candidate passage). The relevance score between a query and a candidate passage is computed as the dot-product between their encoded vectors. The resulting model, called *LawDPR*, is trained on the entailing query-paragraph pairs of COLIEE-2021 Task 2. Results have shown that the paragraph-level retrieval is beneficial in terms of recall, and that LawDPR outperforms BM25, although a combination of both models can lead to further improved performance. In the stage of re-ranking of the retrieved cases, the approach is to summarize the texts of cases and queries, and then to predict whether the summarized query case is relevant to a summarized case. To this purpose, the LED model is fine-tuned on the corpus for the corresponding binary classification task.

[Deroy et al., 2021] analyze expert-generated summaries and model-generated summaries of legal case documents, to understand which parts of the documents are selected from models and domain-experts, and which sentences marked as important by domain-experts are easier or harder to identify for models. To this purpose, 15 extractive summarization models are evaluated over 50 case documents from the Indian Supreme Court. One of these models is BERTSum [Liu and Lapata, 2019b], which is a BERT model fine-tuned for extractive summarization. Results show that BERTSum tends to select the initial portions of the document, even after fine-tuning. This is probably due to the pre-training of BERTSum, which is performed on news article corpus in which the first sentences are usually indicative of the content of the entire document.

[Klaus et al., 2022] focus on extractive summarization of the European regulatory documents, in order to allow non-jurists (e.g., companies that want to ensure compliance with current regulations) a more facilitated comprehension of the documents and decide which regulations need more follow-up. They create the *EUR-LexSum* corpus,[118] consisting of 4K curated European regulatory documents with their summaries and structured into 32 policy fields (e.g., Taxation, Public Health and so on). Such documents are gathered from the Web,[119] which contains also the corresponding summaries in an abstractive form. To this regard, the authors perform a greedy strategy to obtain the extractive version of the summaries, before fine-tuning and feeding the data to a classifier. The obtained dataset is then split into 75% training, 15% validation, and 10% test documents. They fine-tune base uncased BERT, base uncased DistilBERT, and base RoBERTa (gathered from *TransformerSum*,[120] a library based on *BertSum* model [Liu and Lapata, 2019b]), and address the summarization task as a binary classification of the sentence representations of the documents. Results prove that TLMs achieve superior performance compared to the *TextRank* [Mihalcea and Tarau, 2004] baseline, with BERT and DistilBERT performing better than RoBERTa. Furthermore, a combination of TextRank and TLMs is also evaluated, using the former as a pre-filter of candidate sentences, which achieves the highest F1 and precision scores in terms of *Rouge-1* metrics [Lin, 2004].

---

[115] `https://huggingface.co/xlm-roberta-base`

[116] `https://www.lemurproject.org/indri/`

[117] `https://github.com/sophiaalthammer/dossier_coliee`

[Koreeda and Manning, 2021] introduce *Span NLI BERT*, a BERT-based model that performs document-level NLI for non-disclosure agreements contracts. As explained in Section 3.2, the task is to find the implication relation (NLI) between a set of hypotheses and a contract ("entails", "contradicts", or "is neutral"), as well as to identify the spans in the contract which provide evidence for the associated relation. To deal with long documents, a dynamic context segmentation process is involved to split the text into overlapping contexts, assigning a pre-defined number of tokens to each context and marking the span boundaries within the context. The span boundaries are identified by special tokens [SPAN], whereas the input consists of a [CLS] token followed by the hypothesis and the context separated by [SEP]. The task is thus reduced to a multi-label binary classification over the [SPAN] tokens for the evidence identification, and a classification task over the [CLS] token for NLI. Span-NLI-BERT was fine-tuned and tested on the ContractNLI dataset (cf. Section 3.4) against several baselines, based on SVM, TF-IDF and BERT. Span-NLI-BERT was also tested with different backbone models, including BERT without fine-tuning, BERT pre-trained from scratch with a case law corpus [Zheng et al., 2021], BERT fine-tuned on case law and contract corpora [Chalkidis et al., 2020b], DeBERTaV2 without fine-tuning and DeBERTa V2 fine-tuned on span identification [Hendrycks et al., 2021]; the latter turned out to be the best performing model. Overall, it was found that the correct identification of evidence greatly improves the NLI task, the performance of the model degrades in presence of rare labels, the negation by exception in the text damages the model's accuracy for NLI, while the presence of references to definitions does not hurt performance.

Using the ILDC benchmark, [Malik et al., 2021] compare BERT, DistilBERT, RoBERTa and XLNet models, including their hierarchical versions, against non-TLMs. An input text is divided into overlapping chunks of 512 tokens, and the [CLS] representation of each chunk produced by a TLM is passed to a sequential or feed-forward model to get the final case-decision prediction. The best performing model turns out to be the hierarchical model based on XLNet and a BiGRU on top. [Malik et al., 2021] also consider a number of explainability approaches as a post-prediction step (cf. Section 4.8).

Inspired by SciBERT-HSLN [Brack et al., 2021], [Kalamkar et al., 2022] propose a method based on BERT word embeddings along with Bi-LSTM and CRF, for the automatic prediction of rhetorical roles. Results show that SciBERT-HSLN outperforms a CRF model, which uses BERT sentence embeddings, and the model proposed in [Cohan et al., 2019], which only uses BERT. [Kalamkar et al., 2022] also experimented how rhetorical role prediction can be helpful in two applications: extractive and abstractive summarization of court judgments and court judgment prediction. For extractive summarization, BERTSUM is fine-tuned on the LAWBriefs corpus, chunking the input in case of exceeding 512 tokens and pairing each input sentence with its rhetorical role. The resulting model, called *BERTSUM RR*, is compared against BERTSUM fine-tuned on data without information on the rhetorical roles. Similarly, for the abstractive summarization, Legal-PEGASUS is used by splitting the input in chunks (each of 1024 tokens) in one setting, and segmenting the document in terms of rhetorical roles in another setting, which is called *Legal-PEGASUS RR*). Results demonstrate that the use of rhetorical roles improves performance on both extractive and abstractive summarization. As regards the court judgment prediction task, XLNet is used on the model proposed in [Malik et al., 2021], whereby a version of the model is trained on the last 512 tokens of the judgment text of the ILDC corpus, and another version is trained on the same data but filtered on the last 512 tokens sentences associated to the "analysis" role, which turns out to improve the predictions.

In [Chalkidis et al., 2022b] hierarchical versions of several TLMs are evaluated on the LexGLUE data. These methods are particularly motivated by the inclusion of three datasets in LexGLUE, i.e., ECtHR (A and B), SCOTUS and Multi-EURLEX, whose texts substantially exceed the limit of 512 tokens. Even models that handle longer text sequences, such as Longformer, are not able to avoid the truncation of texts in these datasets. Following [Chalkidis et al., 2021c], each paragraph is encoded using the corresponding TLM-based encoder, then a shallow version of the TLM is fed with paragraph encodings and obtains a new representation of a paragraph that is aware

---

of the surrounding paragraphs. The document representation is then obtained from the context-aware paragraph representations through max-pooling. Results show the benefit of this hierarchical approach.

Using the same three LexGLUE datasets, [Mamakas et al., 2022] propose to modify Longformer in order to deal up to 8192 tokens and to combine Legal-BERT@aueb with TF-IDF representations. In the former case, the proposed *Longformer-8192* is designed to increase the maximum input length up to 8192 tokens and to decrease the local attention window size from 512 to 128 to reduce computational burden. A variant called *Longformer-8192-PAR* is also introduced which is Longformer-8192 with the addition of the global token [SEP] between paragraphs to get paragraph-level representations. Also, the following models are introduced: *LegalLongformer*, which is a Longformer warm-started from LegalBERT@aueb and able to handle up to 4096 tokens, *LegalLongformer-8192* and *LegalLongformer-8192-PAR*, which are Longformer-8192 and Longformer-8192-PAR, respectively, based on LegalLongformer. For the second approach, *TFIDF-SRT-LegalBERT* is proposed as a version of LegalBERT@aueb in which duplicate sub-words are removed from the input text to decrease its length. The remaining sub-words are ordered by decreasing TF-IDF, so that the model is induced to attend more the earlier sub-words (with higher TF-IDF). This model is also extended by including a TF-IDF embedding layer. Both the approaches are fine-tuned and evaluated on LexGLUE tasks. The models of the first approach achieve better results on almost all tasks, and surpass the best model for [Chalkidis et al., 2022b], i.e., the hierarchical version of LegalBERT. This demonstrates the effectiveness of warm-start from a legally pre-trained model and the adding of further positional embeddings and global tokens.

A hierarchical approach for handling long documents is also proposed in [Chalkidis et al., 2022a] by using *Hierarchical Attention Transformer* (HAT) models. The HAT models are based on a multi-level hierarchical attention pattern, which comprises a segment-wise encoder (SWE) that applies to segments independently, followed by a cross-segment encoder (CSE), to get contextual representation across segments. Each segment is preceded by the [CLS] token, which represents the segment encoding. These two encoders can be arranged to form different architectural topologies. The text segmentation process is based on a dynamic segmentation strategy that aims to preserve sentence integrity while minimizing padding. The HAT models are pre-trained with the MLM objective task and the parameters of the HAT models are warm-started following BERT and RoBERTa. Thus, they are fine-tuned and evaluated on benchmarks of different domains, including ContractNLI and ECtHR. Results demonstrate that HATs perform better across all tasks than sparse-attention models like Longformer and BigBird, and also they are less demanding in terms of computation and memory resources.

[Shukla et al., 2022] also evaluate three methods to deal with long legal documents: TLMs designed for long text (e.g., Longformer), chunking approaches along with TLMs designed for short summaries (e.g., BART and Legal-PEGASUS), and extractive summarization techniques to reduce the size of the input document combined with abstractive methods. For the latter approach, [Shukla et al., 2022] propose BERT_BART, which first uses a BERT-based extractive summarization model to extract relevant sentences, then carries out a chunking-based BART model to get the final abstractive summary. Nonetheless, the chunking-based methods (i.e., Legal-PEGASUS and BART task-adapted through fine-tuning) and Legal-LED turn out to be the best performers for the document-level evaluation and for the segment-wise evaluation, respectively.

[Shen et al., 2022] fine-tune and evaluate BART, PEGASUS, LED and PRIMERA for abstractive summarization on the Multi-LexSum benchmark. In particular, for the multi-document summarization task, LED and PRIMERA, which deal with long inputs, perform better than PEGASUS and BART on all the three tasks of Multi-LexSum, although no model has shown to be able to generate long summaries of length similar to those provided by humans. When considering short summaries, gold summaries provided in Multi-LexSum are used as input to summarize, instead of the source document. In this setting, on single-document summarization tasks, PRIMERA reaches the performance of PEGASUS and BART, while the performance degrades when BART-generated summaries are used as input instead of the gold ones. Long and tiny versions of generated summaries have benefited from multi-document summarization compared to the summaries of the single-document counterpart.

### 4.4.2 Domain-adaptive pre-training methods

There is also a body of works jointly focusing on domain-adaptive pre-training and handling long documents. The common approach is to apply in-house or existing (e.g., Legal-BERT@aueb) further/from-scratch pre-trained models on segmented or summarized versions of long legal documents.

In the context of employment notice prediction, [Lam et al., 2020] adapt BERT-*base* and RoBERTa-*base* on a corpus of notice case texts in addition to the Harvard case law dataset. Using only the MLM criterion objective, the two models are further pre-trained for text classification with ten epochs and fine-tuned on 409 remaining cases. The goal is to predict reasonable notice based on a free-text summary of the case law, where the summaries are unstructured text data written in plain, non-legal English collected from WestLaw's Quantum service. Surprisingly, however, results have shown a decrease in performance by the domain-adapted models against the out-of-the-box RoBERTa.

[Limsopatham, 2021] compare BERT, Legal-BERT@aueb, Legal-BERT@stanford, and RoBERTa on a text classification task. After removing tokens in the rear, resp. front, of the input texts to ensure the 512 token length, each model is applied on chunks of 200 tokens. An average pooling layer, or alternatively max pooling layer, is used before the classification layer. Results obtained on the ECHR Violation dataset and the Overruling Task dataset, which are multi-label and binary classification tasks, respectively, show that Legal-BERT@aueb and Legal-BERT@stanford achieve better results than the other methods. However, fine-tuned BigBird and fine-tuned Longformer (even though not pre-trained on in-domain documents) outperform the other approaches that adapt BERT to deal with long documents by truncating them.

[Khazaeli et al., 2021] propose a QA system for both factoid and non-factoid questions. It comprises a search repository, which contains the passages and their learned embeddings, a search engine, which retrieves passages by text and embedding similarity, and an answer finder, which re-ranks the retrieved passages. In the stage of passage retrieval, it makes use of a Siamese BERT architecture (a.k.a. SBERT) [Reimers and Gurevych, 2019]. A Siamese Legal BERT system is hence trained on a collection of headnotes to retrieve similar passages, with a regression objective function with cosine loss. A legal BERT model pre-trained from scratch, with a custom legal vocabulary on US case-law documents, receives in input a sentence embedding, and uses mean pooling of the tokens embedding. The answer finder is trained by fine-tuning the in-house Legal BERT model, whose classifier uses the [CLS] representation with two fully connected layers with a final softmax layer. For each input question-passage pair $(q, p)$, the answer finder concatenates their texts in the form '[CLS]$\langle q \rangle$[SEP]$\langle p \rangle$[SEP]', and computes the probability that the passage answers the question.

[Li et al., 2021b] consider the COLIEE-2021 Task 1 as a ranking problem and address the BERT's input length limitation handling the text at paragraph and document level. Firstly, the input is divided into paragraphs. Each paragraph is encoded by a `bert-base-uncased` model that is further pre-trained on the Task 1 corpus, dubbed *BERT-Legal*. The pre-training task is similar to the MLM task but relies on the n-gram masking method to obtain the masked input. Each paragraph is represented by the hidden vector obtained for the [CLS] token. Then, a paragraph level coherence matrix of the query-candidate pairs is derived. The document level representation is obtained encoding the coherence matrix through a max-pooling strategy and a LSTM model. The output of the LSTM is adopted to get the contextual based semantic ranking through relevance prediction. The BERT-Legal ranking is part of a retrieval pipeline, in which the first stage is to recall top-50 relevant cases using BM25-based similarity score.To reduce the training consumption, the authors set a maximum number of paragraphs for a document. The proposed solution has not achieved satisfactory results, probably due to a bad candidates selection operated by the BM25-based recall method. For the Task 2, BERT-Legal is fine-tuned for binary classification on the case law entailment dataset. The authors propose two further versions of the model: the one leverages Task 1 dataset to augment the positive and negative samples in the training set, the other one adversarially train BERT-Legal with Fast Gradient Method [Miyato et al., 2017] to be robust to embedding perturbations. However, both the strategies have shown to decrease the performance of BERT-Legal on the test set, probably due to the different data distribution between Task 1 and Task 2.

[Furniturewala et al., 2021] address both summarization tasks at AILA-2021. For Task 2a, two approaches are proposed: (i) to fine-tune Legal-BERT@aueb, and (ii) to concatenate statistical feature vectors obtained by TF-IDF with the semantic feature vectors obtained by the Legal-BERT@aueb in correspondence to the [CLS] token. The final joint features are classified by a SVM model and logistic regression. For Task 2b, the sentences that were labeled as relevant in Task 2a were concatenated into one summary. The authors reached the first position for Task 2a and the best scores as regards some of the ROUGE metrics for Task 2b.

Concerning the same tasks at AILA-2021, [Jain et al., 2021] use Legal-BERT@aueb to get contextual embeddings of the sentences and fed them to an MLP model. Five training datasets are obtained, so that, for Task 2a, the probability for a sentence to be relevant is the average of all the five probabilities computed by the models. a sentence is labeled as relevant if the average probability computed over the five models exceeds a certain threshold. A second run is proposed for Task 2a, in which both rhetorical and relevance labels are considered using a multi-task learning MLP model. This is motivated by the intuition that rhetorical labels can be helpful for predicting relevance labels. A third run is also proposed, combining the results of the first two runs by averaging all the individual probabilities of each individual trained model. For Task 2b, the average probabilities computed by each run of Task 2a are used to rank the sentences. The sentences in top positions are picked to get the summary, until the summary length limitation imposed by the organizers is reached. All the runs for Task 2a reached leading positions, surpassed only by [Furniturewala et al., 2021]. For Task 2b, the first run achieved the highest F1 score for all the ROUGE metrics, the best recall for some of them, and the second-highest precision for all the metrics.

[Askari and Verberne, 2021] address the COLIEE-2020 and 2021 Task 1 by creating shorter query documents based on three approaches: term extraction with Kullback-Leibler divergence for informativeness (KLI), noun phrase or entity extraction, and abstractive summarization using LED fine-tuned on the COLIEE-2018 data containing human-written summaries of the case documents. Such summaries are then combined using lexical ranking methods based on BM25 and statistical language modeling as well as neural ranking methods based on DRMM, Legal-BERT@aueb, and a version of BERT and Longformer designed for document ranking following [MacAvaney et al., 2019]. Ensemble models are also proposed using the scores of neural and lexical rankers as features for a classifier (the authors experimented with SVM, Naive Bayes and MLP). Such methods are primarily evaluated on COLIEE-2020 data. The results show that the best performance is obtained by the ensemble of BM25 (with KLI summarizer) and BERT (with LED summarizer), surpassing the first team at COLIEE-2020 [Westermann et al., 2020]. On COLIEE 2021, BM25 with KLI summarizer surpassed the first team at the competition [Ma et al., 2021].

**Table 7.** Summary of main methods discussed in Section 4.5.

| Method | Ref. | Downstream Tasks | Lang. | Data | Long docs? |
|---|---|---|---|---|---|
| Zero-shot GPT-3 | [Il and Katz, 2022] | MCQA | EN | bar exams | Yes |
| Zero-shot/Few-shot GPT-3 | [Blair-Stanek et al., 2023] | NLI, QA | EN | CAP cases | Yes |
| Zero-shot/Few-shot/Fine-tuned GPT-3 | [Yu et al., 2022a] | COLIEE 2021 Task 4 | EN | civil code | Yes |
| Fine-tuned GPT-3 (LawGPT 1.0) | [Nguyen, 2023] | QA | EN | legal cases | Yes |
| Zero-shot ChatGPT | [Choi et al., 2023] | QA | EN | constitution | Yes |

### 4.5 GPT-based methods

Since the launch of ChatGPT in November 2022, there has been a worldwide renewed interest towards generative language models, which has inevitably impacted on the legal domain as well. In the following, we overview a few works focusing on GPT-based models that have very recently appeared in the literature.

[Il and Katz, 2022] evaluate the performance of a zero-shot GPT-3 (text-davinci-003) model on the multiple choice component of the US Bar Exam. The questionnaire is typically divided into eight categories, seven of which regard specific law areas and one is used to experiment with the test design. Each category contains 25 questions and 4 candidate answers for each question. Different types of prompts were assessed to get the best answers from the model, and the best strategy turned out to be asking the model to sort three multiple-choice answers for the question. Results show that the model is not able to pass the entire exam with its first choices, however it greatly outperforms baselines based on random choices. Moreover, it reaches the average passing range (58%-62%) in two categories. When considering the top two answers, the model passes the average passing range in all categories and exceeds the average results obtained by human examiners in 5 out of 7 categories, with an overall average score on all categories of 71%, in contrast to the average score of 68% obtained by humans. Results get even better when considering the top three answers, with an overall average score of 88%.

[Blair-Stanek et al., 2023] evaluate several prompting approaches on GPT-3 (text-davinci-003) to perform statutory reasoning on the SARA benchmark. In particular, they consider dynamic few-shot prompting, chain-of-thought prompting and zero-shot prompting. In the few-shot scenario, the four most similar training cases are provided for each test case; the prompt pattern for the training case consists of the text of case labeled with the premise, the hypothesis and the answer (entailment or contradiction), whereas the prompt pattern of the test case contains only the premise and the hypothesis. In the zero-shot prompting, no training cases are used. In the chain-of-thought prompting, a prompt consists of ten training cases, each having the same labels as for the few-shot prompting, followed by a chain-of-thought that explains the reasoning leading to the conclusion (i.e., entailment or contradiction). The same prompt pattern is used for the test cases. Moreover, the prompts can be augmented by adding the statute(s) required to solve the entailment reasoning, since GPT-3 may also have been trained on the US tax code. Following [Kojima et al., 2022], if the model does not explicitly indicate the result of the reasoning (entailment or contradiction), the model is prompted again with the original prompt augmented with the model's answer and adding a clear answer imposition string at the end, i.e., *"Therefore, the answer (Entailment or Contradiction) is"*. Optionally, in zero-shot and few-shot prompting, the model is forced to explain the chain-of-thought reasoning by adding the phrase *"Let's think step by step"* [Kojima et al., 2022]. Results show that the model is highly sensitive to the prompt patterns, with some of the prompts outperforming the previous state-of-the-art BERT-based model [Holzenberger and Durme, 2021]. Also, the chain-of-thought imposition improves the performance but not systematically, whereas the worst prompt combination is, as expected, the zero-shot setting without the text of the statue(s). However, examining the chain-of-thought reasoning, it turns out that the model has a certain knowledge of the US tax code, yet imperfect, meaning that the model tends to refer to the wrong part of the statutes, even when the statute is provided in the prompt.

[Yu et al., 2022a] test the performance of GPT-3 (text-davinci-002) on the Task 4 of COLIEE 2021, adopting the following approaches: zero-shot/few-shot setting (with and without the chain-of-thought reasoning) and fine-tuning. In the zero-shot setting, the prompt pattern consists of instructions, the premise-hypothesis pairs from COLIEE dataset and the question phrase *"True or False?"*. In the few-shot setting, hypothesis-answer demonstration pairs of previously evaluated bar exam questions are given to the model in a 1-shot, 3-shot and 8-shot manner. Results show that the worst performing prompt in the zero-shot setting reaches the best model of COLIEE 2021, which is outperformed when a more instructive prompt is provided to the model. In the few-shot setting, all the $k$-shot combinations exceed the COLIEE 2021 winner, with 3-shot and 8-shot reaching the same performance and surpassing the 1-shot. Following [Kojima et al., 2022], the authors consider another configuration adding the chain-of-thought reasoning imposition (*"Let's think step by step"*) in the prompt and query again the model with its first response and the answer imposition (*"Therefore, the hypothesis is (True or False)"*). Moreover, they also fine-tune the model on the COLIEE 2021 training set in two completion settings: providing only the binary entailment answer or providing also the explanations. In the second setting, the explanations are either pseudo-explanations (i.e., for each premise the most relevant sentence for hypothesis) or explanations created by the model (i.e., for each hypothesis-premise-answer triplets, the model is prompted to generate explanations, with a specific prompt pattern). Results show that fine-tuning with pseudo-explanation surpasses fine-tuning with its GPT-3 explanation, but the use of explanations underperforms the fine-tuning without explanations. This suggests that allowing the model to reason independently might lead to better results. However, the best results are achieved using legal reasoning prompts [Burton, 2017] under the zero-shot setting. This approach aims to lead the model to "think like a lawyer", encouraging it to consider, for example, the facts and circumstances that lead to the legal case, to find the rules governing the issue, to apply the rules to the fact and to determine the entailment task.

[Nguyen, 2023] introduces *LawGPT 1.0*, a GPT-3 based model that provides legal assistance in a conversational manner. LawGPT 1.0 is obtained by fine-tuning GPT-3 on a large legal corpus and is evaluated on several tasks, including the generation of legal documents, legal question-aswering and provide legal advice. Results reveal its competitiveness w.r.t. existing virtual legal assistants.

[Choi et al., 2023] focus on evaluating ChatGPT on four law school exams at the University of Minnesota courses of Constitutional Law, namely Federalism and Separation of Powers, Torts, Employee Benefits and Taxation. The exams consist of 95 multiple choice questions and 12 essay questions. The model is questioned with a uniform set of prompts (i.e., without adapting the prompts to a specific course or question). For multiple choice questions, three prompting approaches are evaluated: simple prompting, which requires to give only the answer, the chain-of-thought prompting (described in [Blair-Stanek et al., 2023]) and the rank-order prompting, which consists in ranking the top three choices (like in [II and Katz, 2022]). ChatGPT was able to pass all the exams, although reaching a low average score; in general, it performed better on the essay questions than on the multiple choice questions. On the essays, in some cases it was as good as or even better than the average performance of human students, but in other cases it strongly failed, particularly when essay questions regard specific cases or theories explained in the course. On the multiple-choice questions, it performed much worse on questions involving numbers and better in questions involving relatively uniform legal rules across the jurisdictions. Based on these results, the authors suggest some guidelines for constructing prompts to help the model generate better outputs, such as specifying at the end of the prompt the tone of the writing of the legal essay, or a word range.

### 4.6 Methods for non-English legal languages

Prompted by the success in English text data, in the past few years there has been an increased interest in replicating the advances in NLP based on TLMs for many other languages. As a result, a plethora of non-English BERT and related models have been trained and developed, such as CamemBERT [Martin et al., 2020], BARThez [Eddine et al., 2021] and FlauBERT [Le et al., 2020] for French, GermanBERT,[121] GBERT and GELECTRA [Chan et al., 2020] for German, AlBERTo [Polignano et al., 2019], GilBERTo[122] and UmBERTo[123] for Italian, KoBERT[124] and KoBART[125] for Korean, RobBERT [Delobelle et al., 2020] and BERTje [de Vries et al., 2019] for Dutch, RoBERT [Masala et al., 2020] and Romanian BERT [Dumitrescu et al., 2020] for Romanian, Greek-BERT [Koutsikakis et al., 2020] for Greek, RoBER-Tuito [Pérez et al., 2022] and BETO[126] for Spanish, AraBERT [Antoun et al., 2020] for Arabic, MacBERT [Cui et al., 2020], PERT [Cui et al., 2022] and MarkBERT [Li et al., 2022] for Chinese, AlephBERT [Seker et al., 2021] for Hebrew, BERTimbau [Souza et al., 2020], RoBERTa-PT-BR[127] and GPorTuguese-2 [Guillou, 2020] for Brazilian, and so on.

In specialized domains, such as the legal one, the challenge underlying the development of TLMs is even more evident, due to the lack of large domain-specific training data for specific languages. In this section, we briefly describe main methods for non-English TLM-based legal learning, organized by language.

**Brazilian.**   [Feijó and Moreira, 2019] experiment with extractive and abstractive models to summarize documents of legal rulings. To this purpose, the *RulingBR* dataset is used, which contains 10K rulings from the Brazilian Supreme

---

[121] `https://www.deepset.ai/german-bert`
[122] `https://github.com/idb-ita/GilBERTo`
[123] `https://github.com/musixmatchresearch/umberto`
[124] `https://github.com/SKTBrain/KoBERT`
[125] `https://github.com/SKT-AI/KoBART`
[126] `https://github.com/dccuchile/beto`
[127] `https://huggingface.co/josu/roberta-pt-br`

**Table 8.** Summary of main methods discussed in Sections 4.6 and 4.7.

| Method | Ref. | Downstream Tasks | Lang. | Data | Long docs? |
|---|---|---|---|---|---|
| Fine-tuned Transformer and TransformerAAN | [Feijó and Moreira, 2019] | AS | BR | legal judgments | Yes |
| Fine-tuned BERTimbau | [Lage-Freitas et al., 2022] | case law classification | BR | legal judgments | No |
| Fine-tuned BERTimbau | [Aguiar et al., 2021] | lawsuit classification | BR | petitions, indictments | No |
| Fine-tuned m-BERT and BERTimbau | [Serras and Finger, 2022] | case law classification | BR | summaries, headers | Yes |
| Fine-tuned Transformer + GCN (R-former) | [Dong and Niu, 2021] | LJP | CN | legal cases | No |
| BERT + BiLSTM + CRF | [Sun et al., 2021] | NER | CN | legal cases | No |
| CEEN (BERT as alternative to BiLSTM module) | [Lyu et al., 2022] | LJP | CN | legal cases | No |
| Further pre-trained LegalBERT@OpenClap+attention (EPM) | [Feng et al., 2022] | LJP | CN | legal cases | No |
| Inverse optimal transport-based rationale extraction approach + fine-tuned Chinese T5-PEGASUS (IOT-Match) | [Yu et al., 2022b] | case matching prediction | CN | legal cases | Yes |
| Fine-tuned BERT + BiLSTM +CRF; UniLM (CLASS) | [Li et al., 2021a] | RRL; AS | CN | legal cases | Yes |
| Fine-tuned Transformer and BERT + Graph Transformer | [Huang et al., 2021] | AS | CN | legal opinion news | Yes |
| Pre-trained Longformer (Lawformer) | [Xiao et al., 2021] | LJP; CR; RC; QA | CN | CAIL-Long; LeCaRD; CJRC; JEC-QA | Yes |
| Fine-tuned FlauBERT | [Salaün et al., 2020] | lawsuit classification | FR | legal cases | No |
| From-scratch pre-trained BERT (JuriBERT); Further pre-trained CamemBERT (JuriBERT-FP) | [Douka et al., 2021] | pleading classification | FR | legal cases | No |
| Further pre-trained BARThez (CriminelBART) | [Garneau et al., 2021] | charge prediction (Cloze tests) | FR | legal cases | No |
| Zero-shot and fine-tuned CamemBERT | [Louis and Spanakis, 2022] | SAR | FR | federal/regional codes | No |
| Further pre-trained CamemBERT+Transformer+graph model (G-DSR) | [Louis et al., 2023] | SAR | FR | federal/regional codes | Yes |
| Fine-tuned GBERT and GELECTRA | [Wrzalik and Krechel, 2021] | case law classification | DE | legal cases | No |
| GermanBERT and DistilBERT | [Tang and Clematide, 2021] | document retrieval | DE | articles and legal cases | No |
| Fine-tuned m-BERT, XLM-RoBERTa, Greek-BERT, and Greek-Legal-BERT | [Papaloukas et al., 2021] | article classification | GR | civil code | No |
| Fine-tuned m-BERT | [Tarasconi et al., 2020] | NER; TC | IT | legal cases | No |
| Fine-tuned BERT (LamBERTa) | [Tagarelli and Simeri, 2022] | SAR | IT | civil code | No |
| Further pre-trained BERT (ITALIAN-LEGAL-BERT) | [Licari and Comandè, 2022] | NER, SC, SS | IT | legal cases | No |
| Fine-tuned KoBERT and KoBART | [Yoon et al., 2022] | AS | KR | legal cases | No |
| From-scratch pre-trained BERT | [Masala et al., 2021] | case law classification | RO | legal cases | No |
| From-scratch pre-trained RoBERTa | [Gutiérrez-Fandiño et al., 2021a] | NER; TC | ES | various legal texts | No |
| | | | | | |
| BERT and m-BERT | [Aydemir et al., 2020] | COLIEE-2020 Tasks 3-4 | EN, JP | civil code | No |
| Fine-tuned hierarchical BERT + BiLSTM; BERT (Long BERT) | [Niklaus et al., 2021] | LJP | IT, DE, FR | legal cases | Yes |
| Fine-tuned monolingual German-BERT, Camembert, UmBERTo Fine-tuned multi-lingual XLM- RoBERTa | [Niklaus et al., 2022] | LJP | IT, DE, FR | legal cases | Yes |
| Fine-tuned m-BERT and m-DistilBERT | [Shaheen et al., 2021] | TC | EN, DE, FR | JRC-Acquis; EURLEX57K | No |
| Fine-tuned Legal-BERT@aueb, m-BERT, WikiBERT, BERT | [Avram et al., 2021] | TC | various EU lang. | JRC-Acquis; OPOCE | No |
| Fine-tuned m-BERT- and DistilBERT-based ParaLaw Nets | [Nguyen et al., 2021b] | COLIEE-2021 Task 5 | EN, JP | civil code | No |
| zero-shot mGPT, GPT-J-6B, GPT-NeoX-20B | [Trautmann et al., 2022] | LJP | EN, IT, DE, FR | legal cases | Yes |

Court, and is divided in 60% training, 20% validation and 20% test samples. Each ruling is organized in sections, the first of which is the summary. The abstractive approaches involved are based on neural networks, and include a Transformer model and a *TransformerAAN* model [Zhang et al., 2018]. The latter uses a cumulative average attention as an alternative to the self-attention in the decoder side to accelerate the decoding procedure. Results show that abstractive approaches outperform extractive methods, with Transformer getting the highest scores.

[Lage-Freitas et al., 2022] evaluate classic machine learning techniques and deep learning models, including BERTimbau, on court decision prediction and unanimity decision prediction. In general, due to the small size of the dataset they propose (cf. Section 3.4), deep learning techniques are outperformed by other methods. One exception is for BERTimbau, which obtains comparable or slightly higher performance when predicting on three-label case outcomes (yes, no, partial) with imbalanced setting of the dataset. Regarding unanimity prediction, deep learning models perform better than the classic ones in F1 score when the data set is balanced.

[Polo et al., 2021] release a library[128] containing available pre-trained language models, including BERT, for the Brazilian legal language. The library also includes a package with useful functions and demo examples to facilitate the use of the models. The models are trained using datasets from several sources, in particular BERT is trained using three datasets, regarding clippings and motions from different Brazilian courts and longer documents from the Court of Justice of São Paulo, starting from a checkpoint of [Souza et al., 2020]. The resulting model is dubbed *Bertikal*.

[Aguiar et al., 2021] evaluate different NLP methods along with several combinations of embeddings of Portuguese language models on the lawsuit classification task. Such methods are trained on a collection of petitions and indictments from the Brazilian Court of Justice of the State of Ceará, in which BERTimbau obtains best performance when lawsuit is embedded as the concatenation of all petitions containing one or more references to a legislation.

[Serras and Finger, 2022] focus on the categorization of case laws considering a Brazilian legal dataset.[129] In particular, they fine-tune m-BERT and the *base* and *large* version of BERTimbau. In the fine-tuning process, pairs of summary and header are given, and the aim is to generate the terms of the header from the summary. From the experimental results, it is observed that BERT-based methods outperform a statistical-based baseline. Larger BERT models perform slightly better than the base ones, highlighting that base models are robust enough to the deal with the task.

---

[128] https://github.com/felipemaiapolo/legalnlp
[129] Two reformulated versions of the Kollemata dataset: https://www.kollemata.com.br/. The dataset is not publicly available.

**Chinese.** [Dong and Niu, 2021] address the legal judgment prediction task as a graph node classification problem, handling constraints among articles, charges and terms of penalty. Using the training set, a global consistency graph is derived, composed of all the possible relations of class labels (i.e., articles, charges and terms-of-penalty) treated as graph nodes. To prevent logically conflicting judgment results, relational learning is introduced into a Transformer model to achieve global and local consistency. The resulting model is dubbed *R-former*. R-former's architecture includes a node encoder module and a node classification module. The first module is composed of two Transformers [Dai et al., 2019]: the former obtains the article representations from raw text, the latter uses masking mechanisms to extract consistency information among nodes belonging to different tasks (article prediction, charge prediction and terms-of-penalty prediction tasks) and distinction information of the same task. The second module consists of a graph convolution network to obtain the relevance score of each node according to the neighbors in the consistency graph. Experiments are conducted using CAIL-*small* and CAIL-*big* datasets from Chinese AI and Law challenge (CAIL2018) [Xiao et al., 2018], with R-former performing the best among all the evaluated competitors.

[Sun et al., 2021] combine BERT, BiLSTM and Conditional Random Fields (CRF) to identify legal case entities. To build an name entity recognition model, the authors leverage BERT's ability of text feature extraction, using it as input layer in order to get word embeddings. In addition, they employ a BiLSTM to get long-term memory information. Finally, state-transition matrix in CRF is used to output the globally optimal sequence, with more accurate labeling according to specific conditions. The resulting model is evaluated using the corpus of People's Daily newspaper along with Legal Case Texts from the CAIL Law Research Cups and other legal texts manually labeled. The authors find that using BERT's word embeddings instead of Word2Vec [Mikolov et al., 2013] improves greatly the recognition accuracy.

[Lyu et al., 2022] present the so-called Criminal Element Extraction Network (CEEN) for reinforced criminal element extraction useful for predicting legal judgments. CEEN is designed to handle misleading law articles, having very similar TF-IDF representations, and indistinguishable descriptions of facts, having different targets and criminals. CEEN is composed of four parts: a fact description encoder, a reinforcement-learning-based element extractor, a criminal element discriminator and a multi-task judgment predictor. The fact description encoder is used to obtain contextual sentence representations of facts. The encoder is typically performed by four hierarchical BiLSTMs, although it can also be a BERT-based variant of CEEN, dubbed $CEEN_{BERT}$, which exploits the addition of the [CLS] token at the beginning of each sentence in the input, like [Liu and Lapata, 2019a]. After obtaining sentence representations, a reinforcement-learning-based element extractor is used to distinguish confusing fact descriptions, then a criminal element discriminator gets the discriminative criminal element representations. Finally, a multi-task judgment predictor outputs the judgment results. Experiments were conducted using CAIL-*small* and CAIL-*big* datasets from Chinese AI and Law challenge (CAIL2018) [Xiao et al., 2018] and a set of baselines, including Chinese BERT, RoBERTa and Lawformer. Experimental results show that the combination of CEEN and BERT significantly overcomes BERT and all the baselines, indicating that language knowledge from pre-training and criminal elements' identification are complementary to each other.

[Feng et al., 2022] propose to leverage event extraction from the fact description of criminal cases to constrain models for the LJP task on the CAIL-2018 benchmark. Indeed, most of the incorrect predictions are due to an inadequate identification of key events in the fact description that determine the final judgment, and the lack of consistency constraints among CAIL sub-tasks (law article, charge and term of penalty prediction). In this regard, the *EPM* model is introduced, which is based on a legal BERT [Zhong et al., 2019a] referred to as LegalBERT@OpenClap, and an attention mechanism, which is guided by event-based and cross-task consistency constraints. EPM is pre-trained on the training set of CAIL and fine-tuned on *LJP-E*, a portion of CAIL-small dataset, provided by the authors and manually-annotated with event triggers and roles. Results have shown that EPM outperforms several baselines for the benchmark and the event extraction process performed jointly with the LJP task is beneficial also for the performance of the competitors.

[Yu et al., 2022b] propose an explainable method, called IOT-Match, for the task of case matching prediction. Given the sentences of a pair of legal cases, it extracts rationales based on semantics and legal characteristics, and generates explanations so that the matching prediction (carried out on eCAIL and ELAM, cf. Section 3.4) is based on the extracted rationales and explanations. More details on IOT-Match are discussed later in Section 4.8.

[Li et al., 2021a] propose *CLASS* (Chinese LegAl judgmentS Summarization), a method to generate abstractive summaries of Chinese legal judgments. Once relevant sentences are extracted from the input, the legal judgments are split into rhetorical roles, then a summary of each rhetorical role is generated. The extraction module uses BERT to get the embedding of each sentence, then a Bi-LSTM model encodes the sequence of sentences. A BERT-BiLSTM-CRF model is trained to split the legal judgments into the rhetorical roles. The UniLM model is chosen to get the abstractive summary of each rhetorical role. CLASS is evaluated using the legal judgment summarization dataset of CAIL2020, divided in 80% training, 10% validation and 10% test samples. Results show that CLASS can achieve higher performance compared to sequence-to-sequence competitors.

[Huang et al., 2021] propose a graph-augmented abstractive summarization model for the automatic summarization of legal public opinion news. A separate graph encoder generates a structural representation of the source document. The components of a document are organized into two graphs, named element relational graph (ERG) and topic interaction graph (TIG). In ERG, the elements extracted from the document (i.e., entities, keywords, and event triples) are nodes that are connected with virtual nodes representing Person, Location, Keyword and so on, in order to get the document-level interconnections. In TIG, similarities of different nodes are represented, where the nodes are the elements of the document that can be viewed as topics and the graph can aid to detect the main topic of the document. The model is composed of a sequence encoder, to yield the sequential representation of the document, a graph encoder, to yield the structural representation of the document, and a sequence decoder, to generate the summary constrained by a dual attention mechanism on the sequential and structural representation of the encoders. The graph encoder is a Graph Transformer Network [Yun et al., 2019], which incorporates global structural information while learning from the neighbor nodes. The sequence encoder and the decoder are based on a vanilla Transformer. A BERT-based embedding mechanism is applied to improve the performance by enhancing the sequence encoder's embedding and by initializing the nodes of the element graph. The model is evaluated using a legal public opinion summarization corpus, named *LPO-news*, consisting of article-summary pairs from the Sina Weibo website. Extensive experiments demonstrate that the proposed model outperforms competing baselines in terms of both ROUGE and BERTScore metrics, also when compared on general news-oriented datasets, and that it can generate more coherent, faithful and informative summaries.

The first pre-trained model for legal long documents is *Lawformer* [Xiao et al., 2021].[130] Following Longformer, Lawformer is pre-trained with MLM objective, continuing from the checkpoint RoBERTa-wwm-ext on a collection of criminal and civil cases published by the Chinese government from China judgment Online. Lawformer is then fine-tuned on a number of tasks, namely legal judgment prediction, legal case retrieval, legal reading comprehension, and legal question answering, for which datasets CAIL-Long, LeCaRD, CJRC, and JEC-QA are used, respectively.

**French.** [Salaün et al., 2020] fine-tune the FlauBERT base cased language model [Le et al., 2020], which is a pre-trained BERT on French corpora, to a collection of lawsuits submitted to a tribunal specialized in disputes between tenants and landlords in Quebec, i.e., the Réegie du Logement du Québec (RDL). The RDL lawsuits are organized into three sections: fact description, legal reasoning and the verdict. The documents also contain metadata describing court location, judge, presence and type of each party: legal person (juridical entity) or physical person, single person or multiple people, male or female and so on. The task is to classify the outcomes of the judgments with three possible labels: penalty (judge convicts the defendant), agreement (judge imposes an agreement) and rejection (judge rejects the the plaintiff's claims). FlauBERT is trained without the use of metadata and it achieves better performance when trained on all texts (discarding verdicts).

[Douka et al., 2021] propose to pre-train small sizes of BERT (base, small, mini and tiny) from scratch using an MLM task and French legal datasets. The latter consist of the decisions of the Court and the Claimant's pleadings from the Court of Cassation, together with raw legal texts crawled from *Légifrance* website[131]. The resulting model, dubbed JuriBERT,[132] is then tested on two downstream classification tasks: to assign the Court's Claimant's pleadings to Chambers and Sections of the Court of Cassation (8 labels), and to classify the Claimant's pleadings to a set of subjects (151 labels). A variant of JuriBERT, pre-trained on Claimant's pleadings, is also provided in *tiny* and *mini* versions, as well as another JuriBERT (JuriBERT-FP) obtained by further pre-training CamemBERT [Martin et al., 2020]. From the evaluation results, it is observed that JuriBERT-*small* outperforms larger models when training on specific sub-languages like the legal one. However, due to limited resources, larger models like JuriBERT-FP and JuriBERT-*base* have been pre-trained with the use of smaller batch sizes than the other models. By fixing a model size, it is shown that JuriBERT pre-trained from scratch on the same task-specific data used in the fine-tuning can lead to better performance compared with only domain-specific models. JuriBERT-FP outperforms JuriBERT-*base*, demonstrating that further pre-training a general-purpose model can be a preferable choice.

[Garneau et al., 2021] propose a further pre-training of BARThez [Eddine et al., 2021] on a French legal corpus for criminal law, collected from the Criminal and Penal Chamber and mined from the Société Québécoise d'Information Juridique (SOQUIJ) website[133]. The legal comprehension of the resulting model, dubbed CriminelBART, was evaluated through Cloze tests regarding the prediction of criminal charges, legal provisions, and privacy.

[Louis and Spanakis, 2022] evaluate lexical and dense models on the BSARD benchmark (cf. Section 3.4). The lexical models are based on TF-IDF and BM25 and are used to retrieve the top-$k$ articles for a given question, based on scores computed for each article. The dense models are bi-encoder models, in two architectures: *siamese*, which uses a unique word embedding model for both questions and articles, and *two-tower*, which uses two separated word embedding models. For each question, a similarity score is computed for all articles that are pre-encoded, and finally the top-$k$ articles based on calculated scores are retrieved. The dense models are evaluated in a zero-shot setting, using word2vec and CamemBERT. To handle long articles for CamemBERT, each text is divided in overlapping chunks.

---

[130] `https://github.com/thunlp/LegalPLMs`
[131] `https://www.legifrance.gouv.fr/`
[132] `http://master2-bigdata.polytechnique.fr/FrenchLinguisticResources/resources#juribert`
[133] `https://soquij.qc.ca/a/fr/`

The fine-tuned CamemBERT has shown to outperform all the other competitors but, considering only the zero-shot variants, the word2vec-based model gives significantly higher performance w.r.t. BERT-based models.

The BSARD benchmark is also employed in [Louis et al., 2023], where a graph-augmented dense retrieval model, called *G-DSR*, is proposed to exploit the statute hierarchy to enrich the article information. G-DSR includes two components: a dense statute retriever and a legislative graph encoder. The first component is a bi-encoder for articles and questions, so that a question and an article relevant to the question get similar representations. Questions are encoded using the [CLS] token representation obtained by CamemBERT further pre-trained on BSARD articles, while the encoding process for statutory articles is left to a hierarchical encoder to handle the length of the documents. Given a pair of article and question, a similarity score is calculated to evaluate the relevance of the article for the question. The bi-encoder is trained through contrastive learning to get effective embedding functions. The second component of G-DSR represents the statute's hierarchy as a directed acyclic graph with two types of nodes: section nodes, corresponding to the headings of the code subdivisions, and the article nodes, representing the textual content of the articles. The edges of the graph model the hierarchy between sections and articles. The semantic information of a node is initialized through the article encoder of the first block and used as the initial node features. A graph neural network is then employed to update the node features by aggregating the local neighborhood information based on the graph structure. The resulting model show higher performance against baselines such as BM25, DPR and a combination of BM25 and mT5. Also, the enriched information given by the legislative graph encoder to the dense statute retriever contributes significantly to improve performance.

**German.** [Wrzalik and Krechel, 2021] fine-tune GBERT and GELECTRA *base* models to re-rank retrieved documents of *GerDaLIR* (cf. Section 3.4). Passages that exceed the sequence length limitation of BERT and ELECTRA are divided along sentence boundaries, then the maximum score is assigned to the passage. ELECTRA demonstrates to obtain higher re-ranking quality compared to BERT.

[Tang and Clematide, 2021] address paragraph-level semantic similarity for legal document retrieval[134], using a corpus of legal cases and statutes in German language gathered from the Swiss Federal Court[135] and Swiss Government[136] websites, and extracting from the cases citations that point to statutes. They use GermanBERT, a BERT model trained on partially legal German text, and the German variant of DistilBERT. An extended attention mask mechanism is performed to combine the *idf* scores of non-neural methods with neural models. This mechanism suppresses low informative tokens in the input, thus impeding the self-attention calculation on those tokens. The authors developed a link-based similarity method to estimate paragraph-level semantic similarity, considering the relations between paragraph cases that share citations to the same statutes. From the experimental results, it appears that GermanBERT, along with the use of the extended attention mask mechanism, offers a clear added value compared to non-neural competitors at specific *idf* threshold.

**Greek.** In [Papaloukas et al., 2021] multiple methods are evaluated on the GLC dataset (cf. Section 3.4) for classifying legal texts. In particular, the authors experimented with m-BERT, XLM-RoBERTa, Greek-BERT [Koutsikakis et al., 2020] and Greek-Legal-BERT [Athinaios, 2020]. Greek-Legal-BERT is pre-trained on documents gathered from a Greek legislative Knowledge Base, called Nomothesia[137] [Chalkidis et al., 2017]. Excluding a few exceptions, Greek-Legal-BERT proves to be the best performer for each evaluation level (i.e., volume, chapter and subject level), followed by Greek-BERT, m-BERT and XLM-RoBERTa, thus confirming the importance of both in-domain and in-language training.

**Italian.** [Tarasconi et al., 2020] evaluate different BERT-based approaches for three business problems in the processing of case law contents for electronic publishing purposes: identification of legal references in the text, new content classification based on relevance, hierarchical labeling of text according to predetermined topics. In the first case, the aim is to identify in a judgment the references to specific laws or other judgments. Judgments come from the Italian Highest Courts of Appeal. Best performances are obtained using a fine-tuned version of m-BERT for NER purposes. In the second case, the identification of the potential relevance of a document aims to select the ones to be eventually published. Documents are mostly gathered from judgments from the Italian Highest Courts of Appeal, but also T.A.R. Administrative Regional Tribunal, Italian Constitutional Court and EU courts. The task is addressed using BERT for binary classification, although best results are obtained using a Random Forest model because of the use of hand-crafted features. In the third case, the goal is to assign each document of the Italian Highest Courts of Appeal, with a set of topics belonging to a publisher's proprietary resource. The task is addressed using a fine-tuned version of m-BERT for extreme multi-label classification. The high-quality of legal data, collected over the years by publisher's managers and decision-makers, allows to successfully experiment with supervised methods.

---

[134] `https://github.com/lilytang2017`
[135] `https://www.bger.ch/it/index.htm`
[136] `https://www.fedlex.admin.ch`
[137] `http://legislation.di.uoa.gr/`

LamBERTa [Tagarelli and Simeri, 2022] is the first BERT-based framework for law article retrieval as a prediction problem, focusing on the modeling, learning and understanding of the Italian Civil Code (ICC). To this purpose, LamBERTa fine-tunes a pre-trained Italian BERT on the ICC and is designed to answer legal questions. The task is conceived as an sequence classification task with a high number (i.e., hundreds or thousands) of classes, which correspond to the number of articles in the ICC, resp. a within-book portion of it, that is used to train a LamBERTa global model, resp. book-specific model. Also, the task is a few-shot learning one, since there are few per-class examples to train a model, which are extracted from each article of the ICC according to one of several schemes of *unsupervised training-instance labeling* defined by the authors. These schemes adopt different strategies for selecting and combining portions from each article to derive the training set, while sharing the requirements of generating a minimum number of training units per article. LamBERTa models have been assessed through single-label as well as multi-label evaluation tasks, based on six different types of queries, which include jurisprudential sentences associated with the ICC articles, and annotations about the interpretation of the meanings and law implications associated to the articles. Also, [Simeri and Tagarelli, 2023] focus on an investigation of the injection of out-of-vocabulary legal terms in LamBERTa models' tokenizer and of the impact of domain-adaptive pre-training of LamBERTa models on article retrieval performance.

[Licari and Comandè, 2022] contribute with ITALIAN-LEGAL-BERT, which is the result of a further pre-training of a pre-trained Italian BERT on a corpus extracted from the National Jurisprudential Archive (pst.giustizia.it), a repository containing millions of legal documents, such as decrees, orders, and civil judgments, from Italian courts and courts of appeal. ITALIAN-LEGAL-BERT was evaluated on named entity recognition, sentence classification, and sentence similarity tasks.

**Japanese.** As previously described, many efforts have been made by researchers in the attempt of modeling TLMs on the Japanese version of COLIEE tasks, especially on statute law retrieval, entailment and question answering.

**Korean.** [Yoon et al., 2022] use a KoBERT-based version of BERT2BERT, a model composed of BERT on both the encoder and the decoder side, as well as KoBART to generate abstractive summaries of Korean legal judgments. The dataset is a collection of legal precedents gathered from the *Korean Court Comprehensive Legal Information* site.[138] Both the models show good results on the task, with KoBART performing better than BERT2BERT.

**Romanian.** [Masala et al., 2021] specialize BERT models for Romanian juridical domain.[139] To this end, the models are pre-trained from scratch using *RoJur*, a large corpus containing civil and criminal cases documents published by Romanian civil courts, and the *whole word masking* technique. In RoJur, each judgment is composed of the description of the parties, brief of the arguments, legal reasoning and final verdict. The resulting model, dubbed *JurBERT*, is then evaluated using two datasets: *RoBanking*, extracting from RoJur common types of cases relating to banking domain, and *BRDCases*, a collection of cases involving the Romanian BRD bank. The downstream task is to predict if the final verdict in a legal case is in favor of the defendant or the plaintiff, so it is addressed as binary classification.

**Spanish.** [Gutiérrez-Fandiño et al., 2021a] train a RoBERTa model on a large legal corpora obtained from a collection of different Spanish datasets, including Legal-ES [Samy et al., 2020]. The resulting model is dubbed *RoBERTalex*, and has been compared with m-BERT and a Spanish RoBERTa [Gutiérrez-Fandiño et al., 2021b]. Since there is no domain-specific evaluation dataset available, the models are tested on general-domain tasks (e.g., NER, classification), on which RoBERTalex obtains good performance.

### 4.7 Multilingual and cross-lingual methods

Unlike English and few high resource languages, many other languages have often been characterized by low resource data, sometimes resulting in limited or absent benchmarks, which negatively affect the amount of pre-training data, and hence of performed of TLMs. To overcome this issue, one common approach is to develop *multilingual language models*, which are pre-trained using large amounts of unlabeled data from multiple languages, under the assumption that low resource languages can benefit from high resource languages due to shared vocabulary and semantic relatedness aspects. Multilingual TLMs have been indeed proposed in the past three years, such as m-BERT and XLM-RoBERTa, mainly differing from each other in terms of number of languages involved, architecture components, pre-training objective functions and corpora. A comprehensive survey on multilingual TLMs is recently provided in [Doddapaneni et al., 2021]. In the following, we shall instead keep our focus on some representative studies of multilingual and cross-lingual TLM-based legal learning.

---

[138] https://glaw.scourt.go.kr/wsjo/intesrch/sjo022.do
[139] https://huggingface.co/readerbench

[Aydemir et al., 2020] propose to use m-BERT along with the BERT-*large* cased and uncased models to address the COLIEE-2020 Task 3. Results have shown that the multilingual BERT model yields better performance than BERT-*large* cased. However, for Task 4, results based on multilingual BERT appeared not be satisfactory as for Task 3, even compared to the competition baseline.

[Niklaus et al., 2021] experiment with several BERT-based methods on binary classification of the judgment outcome, using Italian, German and French legal cases from the Federal Supreme Court of Switzerland (FSCS). Among them, there is also a hierarchical version of BERT similar to [Chalkidis et al., 2019a], where a BERT (monolingual or multilingual) encoder produces fact embeddings that are used as input for a BiLSTM to get the document representation. Another variant of BERT (monolingual or multilingual) proposed in [Niklaus et al., 2021] to deal with long document is *Long BERT*, where a maximum length of 2048 tokens is reached replicating four times the original 512 positional encodings. Results unveil that monolingual models generally outperform the multilingual counterparts, with the overall best results obtained by hierarchical BERT.

[Niklaus et al., 2022] evaluate cross-lingual, cross-domain (i.e., cross-legal areas), cross-regional and cross-jurisdiction transfer learning of several TLMs on the LJP task. An augmented version of the SJP corpus (cf. Section 3.4) is provided. Since most SJP documents are around 2048 tokens, Hierarchical BERT models [Niklaus et al., 2021, Chalkidis et al., 2019a] are used to encode up to 2048 tokens for each document. For the cross-lingual transfer learning, monolingual BERT-based models (German-BERT, Camembert, and UmBERTo), and the multi-lingual XLM-RoBERTa are compared on three scenarios: fine-tuning the models for a specific language (i.e., monolingual fine-tuning), fine-tuning the models across languages (i.e., cross-lingual fine-tuning), and fine-tuning across languages but excluding the target language (i.e., zero-shot cross-lingual fine-tuning). For the first and the second scenarios, two versions of the training set are used, one containing only the documents in the original language contained in the SJP corpus and the other including also the machine-translated versions of the documents. Concerning the monolingual fine-tuning of monolingual models and XLM-RoBERTa, results have shown that monolingual models obtain better performance w.r.t. XLM-RoBERTa, with even better results than [Niklaus et al., 2021]. Moreover, the augmentation of the data through machine-translation seems to further improve the performance. For cross-lingual fine-tuning, a standard fine-tuning of XLM-RoBERTa is compared with an adapter-based fine-tuning of the model [Houlsby et al., 2019, Pfeiffer et al., 2020], which consists on the addition of adapter layers to the model and training them along with the parameters of the normalization layers. Results unveil the beneficial effect of including adapters in XLM-RoBERTa, regardless of the addition of machine-translated documents, and adapters also improve performance for XLM-RoBERTa fine-tuned with zero-shot cross-lingual fine-tuning. As regards cross-regional transfer evaluation, the SJP documents are divided w.r.t. several regions, then XLM-RoBERTa is fine-tuned in three settings: fine-tuning w.r.t. a specific region with data augmentation, fine-tuning across all the regions without machine-translated data augmentation and fine-tuning across all the regions with machine-translated data augmentation. Results show that in most cases a model fine-tuned on the same region of the target is outperformed by zero-shot models (i.e., models fine-tuned on another region). However, cross-regional models obtain better results than regional-specific models, with adapter-based models obtaining top results in most cases. As regards cross-domain transfer learning evaluation, fine-tuning is conducted over three settings: domain-specific fine-tuning with data augmentation, cross-domain fine-tuning without data augmentation and cross-domain fine-tuning with data augmentation. Results demonstrate that cross-domain models outperform domain-specific models in most cases. Unlike the cross-regional transfer analysis, models fine-tuned on the same domain of the target outperform the zero-shot models. Finally, [Niklaus et al., 2022] perform a cross-jurisdiction transfer learning evaluation, by adding to the SJP corpus, concerning the Swiss laws, the ILDC corpus, which contains Indian laws. They consider two fine-tuning scenarios: fine-tuning XLM-RoBERTa on only the machine-translated Indian cases (zero-shot fine-tuning) and fine-tuning the models with original SJP training set, machine-translated SJP cases and machine-translated ILDC cases (further augmented fine-tuning). Results show that zero-shot models perform poorly, but the further-augmented models outperform the results of the cross-lingual models evaluated in the cross-lingual transfer learning analysis. Again, adapter-based models obtain highest performance in most cases.

[Shaheen et al., 2021] evaluate m-BERT and m-DistilBERT in large-scale multi-label text classification task using a zero-shot cross-lingual transfer learning scheme and a joint learning scheme. In the first case, models are built using an English training set although performance is tested on French and German test sets. In the second case, models are trained using all the three languages. To this purpose, the multilingual version of JRC-Acquis [Steinberger et al., 2006] and an extended version of EURLEX57K [Chalkidis et al., 2019b] are exploited. From the experimental results, m-BERT outperforms m-DistilBERT in zero-shot transfer learning scheme in both datasets. As regards the joint learning scheme, m-BERT and m-DistilBERT are compared with monolingual BERT-based methods using the English test set. The multilingual models obtain about 96.83-98.39% of the performances of monolingual models in JRC-Acquis, while in EURLEX57K better results are obtained by m-BERT. Comparing the performance in zero-shot and joint learning scheme, m-BERT (resp. m-DistilBERT) in zero-shot scheme achieves about 86% (resp.79%) of the performance in the joint learning scheme on French and German test sets. The zero-shot results indicate that multilingual models are able to achieve transfer learning from English to French language more easily than from English to German.

[Avram et al., 2021] propose a tool for multi-label classification of legal documents on 22 languages. For each language, BERT-based methods are fine-tuned using the JRC-Acquis [Steinberger et al., 2006] and the *Publications Office of the European Union* (OPOCE) corpora, manually labelled with almost 7K EuroVoc[140] descriptors. The choice of BERT models for each language is conducted prioritising pre-trained models on legal domain corpora, otherwise according to the following order: models pre-trained on a specific language, models pre-trained on monolingual Wikipedia, multilingual models. As a result, Legal-BERT [Chalkidis et al., 2020b] is used for the English language, m-BERT [Devlin et al., 2019] for the Maltese language, WikiBERT models [Pyysalo et al., 2021] for Bulgarian, Lithuanian, Latvian, Slovak and Slovene, and monolingual BERT models for the remaining 14 languages. From the evaluation results, m-BERT obtain the worst performance, probably due to the low number of Maltese documents as well as due to the use of a multilingual model. In contrast, it is interesting that WikiBERT models obtain satisfactory scores, even better than Legal-BERT.

[Nguyen et al., 2021a] introduce *ParaLaw Nets* [Nguyen et al., 2021b] to address the COLIEE-2021 Task 5. This is a family of pre-trained models that rely on sentence-level translation information to reduce language ambiguity and increase performance in legal tasks. The core idea in these models is that, in the translation process, the most correct meaning of a sentence will be captured by the model. To this end, the model is pre-trained on sentence-level cross-lingual tasks, while the COLIEE task is used in the fine-tuning phase. According to the type of pre-training task, ParaLaw Nets are divided in *NFSP* (Next Foreign Sentence Prediction) and *NMSP* (Neighbor Multilingual Sentence Prediction) models. The former formulates the pre-training task as binary classification, the latter as multi-label classification. The pre-training input consists of sentence pairs, where a sentence can be in its original language or in the target language. As previously described in Section 4.2.2, the NFSP task is basically a Next Sentence Prediction task where the input is composed of sentences pairs expressed in different languages. In the NMSP models, training data also includes sentence pairs with the same language. The labels in the NMSP task correspond to four training data generation schemes, namely random sampling, normal order, reverse order, and non-contiguous. In both tasks, m-BERT and DistilBERT's architectures are used respectively as the base and the distilled version for ParaLaw Nets. The models are trained with Japanese-English legal data, but the approach can be generalized to all language pairs. The data used to pre-train ParaLaw Nets is provided by Japanese Law Translation website[141]. The data used in the fine-tuning phase come from the Japanese Civil Code and COLIEE. The authors use also augmentation strategies in the fine-tuning phase, which mainly consist in the statement negation of the original sentences. From the competition results, it is observed that the base model using NFSP obtains the highest score, while the base model using NMSP reaches the third position, proving the effectiveness in exploiting cross-lingual information.

[Trautmann et al., 2022] explore the potential of legal prompt engineering on zero-shot GPT-based models for multilingual LJP. In particular, they evaluate mGPT, GPT-J-6B[142], i.e., a 6 billion parameter version of GPT trained on the Pile dataset [Gao et al., 2021], and GPT-NeoX-20B on the ECHR and FSCS datasets. A prompt template is defined by mapping the LJP task in a question-answer form, where the question requests to detect whether or not there are violated articles in the document. Results have shown that the proposed prompting methods outperform simple unsupervised baselines, but perform worse than supervised (task-adapted) models, thus leaving much room for improvement especially in terms of defining templates suitable to different corpora and languages (as well as dealing with long documents without truncating at 2048 tokens).

### 4.8 Dealing with explainability and interpretability issues

As for any sophisticated neural network models, the TLMs' behavior at both learning and inference phases is not straightforward to fully understand. This clearly may limit their applicability, or even increase fear in legal actors or general public that who may want to use such AI-based tools, especially for high societal impact fields like law. It is therefore demanding to resort to techniques that can provide explainable justification for the models' decisions, or that can make their prediction outcomes interpretable.

**Post-hoc explanation.**   One approach is to infer post-hoc explanations. Within this view, some works such as [Savelka and Ashley, 2021] and [Tagarelli and Simeri, 2022] provide insights into the informativeness of the attention weights produced by a BERT model for selected use cases, through the lens of the interactive visualization tool *bertviz* [Vig, 2019].[143] The general objective is to inspect the formation of complex inter-token relationships and the corresponding distinctive attention patterns that most influence the final representation of a given sentence in input and, therefore, such as to justify the model's outcome for that input. A different perspective is taken in [Simeri and Tagarelli, 2023], where the explanation is accomplished by approximating a BERT-based classifier locally by an interpretable linear model, by means of a technique known as LIME - Local Interpretable Model-Agnostic Explanations [Ribeiro et al., 2016]. For a given input sentence, LIME explains a classifier's behavior "around" that query instance, by weighing perturbed versions of the input by their proximity to the original query instance, then observing the associated predictions by the underlying classifier to determine which of those changes will have most impact on the prediction of the original query.

---

[140] https://data.europa.eu/data/datasets/eurovoc
[141] https://www.japaneselawtranslation.go.jp/
[142] https://github.com/kingoflolz/mesh-transformer-jax

[Wehnert et al., 2022] address interpretability using KERMIT to encode symbolic syntactic parse trees of queries and articles in addition to BERT representation of input sentence, thus injecting further linguistic knowledge. KERMIT is specifically designed to include syntactic interpretations in deep neural networks. The KERMITviz architecture [Ranaldi et al., 2021] enables the visualization of which part of the sentence is used during the inference step.

[Malik et al., 2021] evaluate a number of explainability methods as a post-prediction step of the case decision prediction task. To this regard, legal experts are asked to mark the sentences they consider as explanations for the judgments (from a portion of the ILDC test set) and assign each explanation sentence with a score reflecting the importance of the explanation for the judgment. A hierarchical system composed of XLNet and BiGRU on top is chosen as the base model on which to experiment an explainability method based on *occlusion* and inspired from [Zeiler and Fergus, 2014] and [Li et al., 2016]. Documents are divided into chunks of 512 tokens, each chunk embedding is masked at a time, then the masked input is given to the BiGRU component of the model. A chunk explainability score is computed as the difference between the output probabilities of the prediction calculated on the masked input and the unmasked input. The sentences that explain the final decision correspond to the chunks with positive scores. Similarly, each sentence of selected chunks is masked at a time and supplied to the XLNet component of the model. The difference between the logits calculated on the masked input and the original logits of the prediction represent the explanation score, so that the top-k sentences for each chunk are selected as explanations. The analysis of the explanations given by the occlusion method indicate that most of the relevant information for the judgment is located at the end of the document; however, explanations selected by the occlusion method are shown to be significantly different from explanations given by legal experts.

**Early explanation.**  A different approach is to produce explanations during the data modeling or learning process. [Liu et al., 2021a] consider the interrelation between charges and the court view sections in legal documents. The court view is regarded as charge explanation since it contains supporting information for a charge and is charge-discriminative (i.e., strongly charge-dependent). Within this view, the JPGM method is introduced to jointly predict charges based on fact descriptions and generate court views. The key idea is to predict a group of similar charges that may lead to confusion. To this purpose, charge-discriminative keywords are defined for each charge, then an attention mechanism is involved to select the best matching keywords for a charge predicted by a classifier, finally the generation module provides the court view based on the fact description and the best keywords. The generated court view and fact description are also fused to refine the previous classifier prediction. JPGM has been shown to perform better than baselines based on CNN, GRU, vanilla Transformer, and graph neural networks.

In the context of alleged violation prediction, [Chalkidis et al., 2021c] propose to regularize the extraction of rationales, as the paragraphs of the input that support the decisions, by constraints that reward the model if its decisions are based on concise rationales. The rationale constraints include *sparsity*, *continuity*, *comprehensiveness* and *singularity*. The first two constraints encourage to select a small number of paragraphs that sufficiently justify the allegation and to prefer contiguous paragraphs, respectively. The comprehensiveness constraint requires to use a mask to ideally contain all the paragraphs that support the correct decision, whereas the reverse mask should contain the irrelevant paragraphs; this way, the output of the model based on the mask (i.e., the estimated probabilities on the allegations) should be better than the output of the model based on the reverse mask. Moreover, the singularity constraint requires that the selected mask should be not only better than its reverse version, but also w.r.t. any other mask. The effect of such constraints are evaluated w.r.t. *faithfulness* and *rationale quality*, given a sparsity threshold, where the latter is calculated comparing the predicted rationales with gold annotations, and faithfulness measures how much the rationales reflect the reasoning of the model, by computing the difference between the predicted probabilities obtained by the model over the whole text in input and the predicted probabilities obtained considering only the (complement of) extracted rationales. The singularity constraint is shown to improve faithfulness and rationale quality in relation to both silver and gold rationales; by contrast, continuity appears not to be beneficial for the task at hand.

[Santosh et al., 2022] start from the observation that LJP models on ECtHR cases tend to be confused by distracting factors in the text that might originate from the corpus construction, the case distribution, or spurious correlations with the outcome. In this regard, an expert-informed deconfounding method based on adversarial training

---

is introduced to prevent a model from being influenced by distractors recognized by ECtHR experts. To this purpose, [Santosh et al., 2022] train and evaluate BERT models on LexGLUE (ECtHR Task A and B) and ECHR benchmarks; in particular, a BERT variant of hierarchical attention networks [Yang et al., 2016] is chosen as base model, where input texts are segmented with a greedy sentence packing strategy and encoded using Legal-BERT@aueb.

To evaluate the alignment with expert rationales, the expert relevance assessments provided for ECtHR by [Chalkidis et al., 2021c] are used. The ability of the model to identify correct rationales at the paragraph level is evaluated through an interpretability technique that determines the importance score for each paragraph by measuring the impact of a particular input token on the final prediction. A token-level focus score is calculated using the integrated gradients [Sundararajan et al., 2017], then paragraph-level scores are obtained aggregating the token scores in the paragraph. The top-k paragraphs according to the paragraph scores are compared with the golden paragraph rationales. Results have demonstrated that the deconfounding process effectively improves the model alignment with expert rationales in ECHR benchmarks; however, such improvements are marginal and there is still a large gap between paragraphs suggested by models and what legal experts have annotated as relevant.

To address a legal case matching task, [Yu et al., 2022b] emphasize the importance of taking into account the difference between the roles of sentences corresponding to rationales and other sentences in a legal case, as well as distinguishing rationales that are in favor or against the matching decision. Each sentence of a given pair of legal cases is assigned one of the following labels: not a rationale, a key circumstance, a constitutive element of a crime or a focus of disputes. The goal is to extract aligned and misaligned rationales, to assign a matching label for the legal case pair (not matching, partially matching or matching) and to provide the set of sentences explaining the reasons for the label. The problem to extract aligned and misaligned rationales is formulated as an optimal transport problem, where the probability of the cross-sentence coherency (pro and con rationale pairs) is computed to provide evidence for the matching. The problem is guided by an affinity matrix, which reflects both semantics and legal feature relations between cross-case sentences. The affinity matrix and the optimal transport, based on both sentence embeddings and a manually-labeled (noisy) alignment matrix, are learned by the inverse optimal transport (IOT) process. The IOT process consists of solving a bi-level optimization problem, in which the affinity matrix is the upper-level variable and the optimal transport is the lower-level variable. The optimal transport is then fitted to the alignment matrix and used to help the generation of the explanations. To this purpose, a Chinese T5-PEGASUS [Su, 2021] model for each matching label is fine-tuned to get the label-specific candidate explanations. Finally, the model performs the matching prediction based only on the extracted rationales and the label-specific explanations. In this way, the noisy sentences of the legal cases pair are filtered out and not involved in the matching prediction, reducing the number of input sentences to be processed. The proposed IOT-Match model is fine-tuned and tested on eCAIL and ELAM datasets against legal case matching competitors. The quality of the extracted rationales and the generated explanations is verified by conducting an empirical analysis on faithfulness (described above) and *plausibility*, which measures how the model explanations are convincing to humans. In particular, to assess the plausibility of rationales, resp. explanations, the ones produced by the IOT-Match and competitors have been compared with those generated by humans, unveiling that the rationales, resp. explanations, by IOT-Match are more consistent with human annotations. Moreover, as concerns the faithfulness of rationales, IOT-Match has been evaluated with and without rationales, eventually finding that the extracted rationales play an important role for the task of legal case matching.

As previously described in Section 4.7, [Feng et al., 2022] propose an event extraction process from the fact description of criminal cases to constrain the prediction of LJP models on CAIL2018. A hierarchy is defined for the events of legal cases, based on Chinese law articles. Events are detected from specific triggers, i.e., words expressing the occurrence of the event. Each event has a role in the fact description, and the extracted events and constraints on such events are used to help the learning process of the EPM model for the LJP task. The context representation of the fact description is used to query all law article candidates and to determine the most relevant semantics in the articles. The context representation of the fact description and the most article relevant semantics are given as input for the CAIL sub-tasks. The base model is augmented with a hierarchical event extraction layer to identify event triggers and roles while jointly performing the LJP task. The judgment prediction is thus inferred through detected event features. The model is guided by *event-based constraints*, in order to search for a single trigger and its related roles, and *cross-task consistency constraints*, in order to take into account dependencies between the sub-tasks, since each article establishes the charges and the range of penalty terms.

## 5 Discussion

In this survey we investigated the use of Transformer-based language models (TLMs) for legal AI-based tasks. We conducted a detailed study of the approaches proposed in the literature in this area, and we categorized the problems that received a particular attention from the scientific community into three macro categories, namely legal search, legal document review, and legal outcome prediction; we noticed, however, that such categories are apparently interleaved and interrelated, and their definition is mainly intended for the sake of presentation. Retrieval, entailment and question answering have traditionally been the most frequently considered legal problems, also due to the role played by the COLIEE competition which has represented an important venue to foster the development of AI-based approaches in the legal domain, including those relying on TLMs. But equally important are a number of other tasks, ranging from named entity recognition to judgment prediction, from abstractive/extractive summarization to rhetorical role labeling. Yet, the various tasks are often interrelated, since a common approach to address them is to reformulate a legal-specific task in terms of a more general, machine-learning task, mainly focusing on classification and similarity. It is worth noticing, however, that legal text similarity and classification can be challenging for many reasons. For instance, there might be multiple classes to be assigned with a single case, as a case can span multiple areas of law. Moreover, the categorization of a case can vary depending on the particular court handling it. Yet, there is no universally agreed-upon set of areas of law clearly defined into a taxonomy [Mistica et al., 2021].

**TLM impact on legal tasks.** By examining the literature, there is evidence of how TLMs have been able to push forward the state-of-the-art in a variety of tasks for the legal domain. Again, a representative case corresponds to the COLIEE competition, where we notice a growing and successful use of TLMs. Nonetheless, the competition as well as other venues have shown that non-TLM methods, including the traditional vectorial space models (e.g., TF-IDF or BM25), are still useful especially when combined with TLMs, particularly as a data pre-processing or filtering step. This generally serves a twofold purpose: to exploit both the lexical modeling of traditional techniques and the semantic knowledge of TLMs (e.g., [Nguyen et al., 2021a]) and to perform an initial selection of possible candidates through the ranking score of traditional techniques and then re-ranking such results through TLMs (e.g., [Althammer et al., 2021]).

It does not come to surprise that, being the first and most popular among the TLMs, BERT is widely involved in the existing approaches, often along with its early variants, such as RoBERTa, DistilBERT, DeBERTa. Yet, recent works tend to consider more advanced architectures, depending on the downstream task, which have shown to be very competitive, such as ELECTRA and XLM-RoBERTa as encoder-only models, XLNet and the GPT family of models as decoder-only models, T5 and BART as encoder-decoder, along with task-specific and long range models such as Longformer, BigBird, DPR and PEGASUS.

Moreover, a key aspect that determines the effectiveness of a TLM-based framework compared to other deep learning approaches is the type of training adopted to build the models. In the legal domain, while most studies have focused on task-adaptive fine-tuning, there has also been an increased interest in performing domain adaptation through further pre-training or pre-training from scratch a TLM, again starting from BERT models (e.g., [Chalkidis et al., 2020b, Holzenberger et al., 2020, Zheng et al., 2021]). [Song et al., 2022] carry out an empirical evaluation on the effectiveness of domain-specific pre-training for the legal domain, on a number of datasets and tasks including binary classification, multi-label classification, multiple choice question answering, summarization, and information retrieval. Results have shown that domain-specific pre-trained models can lead to 1%-5% higher performance than general domain pre-trained models, but on condition that the datasets involved are very close to the pre-training corpora, thus concerning the same legal sub-domain.

Regardless of the type and size of TLM and its training, the current trend for best-performing frameworks is to exploit techniques that are extrinsic to the particular language model, such as data augmentation, data enrichment, and ensemble strategies. Data augmentation is performed in several ways, for example through back translation (e.g., [Rabelo et al., 2020]), focusing on the logical mismatches between articles and questions (e.g., [Yoshioka et al., 2021a]), negating the statements in the original sentences (e.g., [Nguyen et al., 2021b]), or retrieving top-$k$ irrelevant articles for a query according to a similarity score (e.g., [Wehnert et al., 2021]). In the latest editions of COLIEE, several works use previously released data to obtain more training samples for the tasks at hand. Also, a few works have adopted ensemble strategies, which combine the results of independent systems to boost the overall performance; for example, the ensemble system of [Nguyen et al., 2021a] reached the second position in the COLIEE-2021 Task 3, the one by [Shao et al., 2020a] ranked first in the COLIEE-2020 Task 3, as well as the ensemble method proposed in [Rosa et al., 2021] achieved the first position in the COLIEE-2021 Task 2. Data enrichment is typically obtained through the use of taxonomies ([Tziafas et al., 2021, Chalkidis et al., 2021a]) and thesauri ([Kim et al., 2021]).

**Resource availability on the benchmarks.** The need to devise the above mentioned strategies for improving the performance of a model also arises from the awareness that free-access legal resources are often limited or partially available. According to [Song et al., 2022], one of the most significant challenges in legal NLP is the lack of large-scale high-quality datasets, due to the costs of the annotation processes that require knowledge of the legal domain. In effect, although there exists a significant body of legal corpora that have been used for training and evaluation of TLMs for legal tasks, as we have analyzed in Section 3.4, the current landscape of legal benchmarks is still not comparable in size with the largest NLP and information retrieval datasets; in fact, apart from few exceptions corresponding to multi-task benchmarks (e.g., LexGLUE), most legal benchmarks and datasets have size of thousands or tens of thousands of documents. Moreover, we believe that attention should be paid to the maintenance of existing benchmarks, so as to keep them updated on important changes in the legislative status of a jurisdiction. Yet, there is room for building larger or new freely available benchmarks to evaluate legal tasks in new contexts (e.g., online social media companies and online trading companies involved in the Web3), as well as contexts that require controlling specific types of bias and/or specific ethical aspects concerning the diversity or disparity of legal norms within and across different countries.

Creating larger and more representative legal corpora depends on a number of aspects that include not only those strictly pertaining the selected evaluation goals for a particular task, but also factors relating to the language resources to be involved (e.g., open/free access availability, various forms of bias), the target audience with their different levels of expertise on the domain (e.g., lawyers, courts, law firms, citizens), as well as ethics-related and privacy-related aspects to be considered, to mention a few.

**Resource availability on the model training.**   The limited availability of legal data for a particular task might prevent an adequate training of the model, thus making it unable to properly internalize the meaning of legal texts and to generalize the knowledge learned in sufficient detail for successfully dealing with unknown input. Indeed, it has been shown that it is not guaranteed that a model pre-trained on a legal corpus can significantly improve upon its corresponding general-domain pre-trained model fine-tuned on the target task, in all situations. Besides the previously mentioned [Song et al., 2022] about the affinity between the data for the downstream task and the pre-training data, according to [Wang et al., 2020a], the advantage of domain-specific pre-training is also related to the size of the data used for the downstream task: that is, the benefit gained through domain-adaptive pre-training is likely to be more significant when the downstream task appears to be more low-resource. This is also confirmed in [Gururangan et al., 2020], where another form of pre-training is highlighted and named as task-adaptive pre-training, i.e., (unsupervised) training the model on a smaller but directly task-relevant corpus. This form of pre-training has shown to be not only competitive w.r.t. domain-adaptive pre-training but also beneficial when combined with it for improving the performance on the downstream task. However, the above findings were not proved for the legal domain, which certainly opens to opportunities for further research.

In addition, by also taking into account models' implementation technicalities, other difficulties can be identified. One is the imbalance between positive and negative examples, for which solutions include the filtering of possible candidates (e.g., [Nguyen et al., 2021a]), oversampling methods (e.g., [Shao et al., 2020a]), and the generation of artificial examples (e.g., [Rosa et al., 2021]).

**Resource availability on the jurisdiction language.**   Data availability issues are even more exacerbated when moving from a high-resource language, such as English, to low-resource languages. Several works have indeed been concerned with the legal domain for poor-resource language families, such as Romance, Slavic, Germanic, Uralic, but also oriental languages such as Chinese and Japanese. For such languages, the use of TLMs has enabled a breakthrough in many cases. For example, in [Serras and Finger, 2022], the TLM-based method improves upon statistical baselines; in [Sun et al., 2021], it reaches leading positions; in [Masala et al., 2021], all the baselines are consistently outperformed; in [Papaloukas et al., 2021], TLMs prove to be superior to vector-space-model based learning techniques and RNN-based methods, in particular when they are domain-adapted. However, for low-resource languages, TLMs can perform worse than others (e.g., [Lage-Freitas et al., 2022]), or large TLMs may suffer in performance in contrast to smaller counterparts (e.g., [Douka et al., 2021]). In some cases, when benchmarks are very limited or even absent, an evaluation on the specific domain is not practicable (e.g., [Gutiérrez-Fandiño et al., 2021a]). One common approach to overcome this issue is to develop multilingual and cross-lingual language models, under the assumption that low-resource languages can benefit from high-resource languages due to shared vocabulary and semantic relatedness aspect. Although there is evidence that the use of multilingual TLMs has helped to improve the state-of-the-art, there is certainly room for improvement (e.g., in [Avram et al., 2021]).

**Linguistic issues.**   Even more challenging is developing a model that can incorporate the many linguistic nuances and subtleties of legal documents already in the early stages of its construction, such as during the tokenization or masking processes. This appears to be critical especially for the case law data: in fact, while law statutes and articles are usually written in a language that should be as much understandable as possible to non-experts as well, legal cases and judgments can be particularly tricky to understand, even for humans. A related issue is the language mismatch between statutes, describing legal concerns in form of abstraction, and law cases, describing real facts; for instance, [Savelka and Ashley, 2022] focus on the interpretation of statutory terms, by investigating on how a particular term has been explained and applied in the past, thus allowing for the lawyers the construction of arguments which support or counter particular interpretations. This is a complex scenario, since there may be little lexical overlap between statutes and cases, thus making it more difficult to understand the legal text and extract key concepts from the case law to be retrieved in the statutes. In this regard, the quality of vocabulary can play an important role. The legal language is full of specific terms with a precise meaning, which are in many cases not included in general-purpose vocabularies. Few works address this issue by injecting legal words in the vocabulary (e.g., in [Tagarelli and Simeri, 2022, Simeri and Tagarelli, 2023]), in order to prevent that the TLM tokenizer will break out out-of-vocabulary terms that are found in a target legal corpus. Clearly, an enriched vocabulary cannot be enough for a model to deeply learn syntactic, lexical, and semantic patterns of the legal language. More specifically, modeling syntactic patterns as predicates, legal reasoning methodologies can provide a significant support. Legal reasoning is in fact one of the primary challenges identified in [Zhong et al., 2020], since it should in principle adhere to well-defined rules.

In addition to all these aspects that are pertinent to the legal domain, there are also general language peculiarities that still pose challenges for TLMs, such as recognition of pronominal forms, anaphora-related issues, understanding various forms with negations, etc. An example is how to distinguish texts that look very similar to each other but actually have a logical or semantic mismatch [Yoshioka et al., 2021a].

**Document length.**   Compared to the earliest approaches, whereby the limitation on the maximum number of token conditioned the ability of TLMs in handling long documents, the current trend is to learn a more conservative, "lossless" model, which can elaborate a text at paragraph-level (e.g., considering embeddings of entire paragraphs rather than individual words, as in [Shao et al., 2020c]), or that it can directly process longer documents, such as Longformer (e.g., [Xiao et al., 2021]), or DPR (e.g., [Althammer et al., 2021]). Another option can be to filter out noisy sentences with sentence extraction techniques ([Yu et al., 2022b]) or to perform a span-level approach ([Koreeda and Manning, 2021]), in which the document is divided into overlapping contexts containing spans. On the other hand, a different perspective is to incorporate a summarization step into the language understanding process with the aim of providing a concise meaningful version of the input text. Some of the discussed works indeed involve a summarization task (e.g., [Alberts et al., 2020, Rossi and Kanoulas, 2019, Kim et al., 2021]), others increase the maximum input length (e.g., [Mamakas et al., 2022]) or apply hierarchical attention patterns (e.g., [Chalkidis et al., 2022a]), but TLMs have surely the potential to be further developed for achieving enhanced performance.

However, to significantly improve their ability related to long range modeling, TLMs should cope with larger computational resources to process, model, store, and manipulate long passages of text, which also impact on advanced ability to reason coherently across long-range dependencies. Addressing these challenges will likely involve advancements in model architecture, training methodologies, memory management, and computational efficiency.

**Structures at document-level and corpus-level.**   Another important aspect is to exploit explicit structure levels within and/or across legal documents. For instance, legal source citations within the text of articles or cases could be managed by replacing the citation with the cited article, as in [Rabelo et al., 2020], although a much more promising approach would be to model and mine legal citation networks. As discussed in [Locke and Zuccon, 2022], citations play a very important role in legal decisions: since the decisions of judges must be in accordance with the doctrine of precedent, which establishes that lower courts must observe the decisions of higher courts, they usually explain the reasons for their decisions by proving the agreement with the previous decisions of the binding authority. Thus, modeling and learning from feature-rich legal citation networks is desirable, however their potential in retrieval and entailment tasks has not been fully explored yet.

[Bhattacharya et al., 2020b] consider both the logical subdivision of statutes and their citation information to measure similarity of legal case documents. To this purpose, the authors introduce *Hier-SPCNet*, a precedent citation network augmented with the information about the hierarchy of the statutes (e.g., an act is typically structured in parts, chapters, topics, sections). A statute is therefore modeled as nodes of different types, each representative of the structural levels, and two types of edges, the one reflecting the hierarchy of the nodes and the other one indicating the citations between cases or between a case and a statute. Adding this structural information shows to lead to better document similarity estimation than competitors based on precedent citation network only. A combination of textual features and legal citation networks is also proposed in [Paul et al., 2022a]. Given an heterogeneous network with nodes representing statutes and legal fact descriptions and edges representing hierarchical structures of the statutes as well as citations between statutes and facts, the goal is to predict which statute is relevant to a newly introduced fact, i.e., if a link exists between statute and the new document. The text of a statute or fact is considered as an attribute of the node. Two separate encoders play the role of encoding the attributes and the structural information given by the network, where attribute encoding is performed through hierarchical attention network [Yang et al., 2016], and structural encoding is obtained through metapath schemes [Fu et al., 2020]. Results show that the exploitation of textual and graphical features can lead to significant improvements in performance compared to state-of-art competitors (BERT-based included) for the task. Also in the work of [Louis et al., 2023], the structure of statutes is regarded as a key point for the success of neural models in statutory article retrieval tasks. A graph-augmented dense retrieval model is defined to exploit the topology of the statutes to expand the article information. The resulting knowledge-rich cross-article representations contributes significantly to the improvement of performance of the dense statute retriever. [Wang et al., 2022] state that the document structure, and in particular the relations among the participants of a litigation, is essential to recognize and classify legal cases of different categories yet with similar topics. To this purpose, four relation graphs are defined, each modeling different types of relations between participants (i.e., plaintiffs and defendants), namely the relations between the participants and the matters of the dispute, the actions performed by the participants, the topics related to the participants, and the relations between facts and third parties. Each graph consists of nodes representing the document, the participants, and a set of nouns and verbs selected from the text and whose relations to participants express either matters, actions, keywords, or facts and third parties. The graphs are then aggregated in one graph by merging the common nodes, then it is provided in input to a graph attention network to get the document representation for the classification task. Results on real-world Chinese documents regarding twenty semantically-similar disputes show that the proposed model outperforms all the considered baselines, included TLMs like BERT, RoBERTa and Lawformer.

A related aspect is also a debate on the notions of legal similarity. In fact, while generally two documents would be similar if legal experts evaluate them as similar in their contents, [Bhattacharya et al., 2020b] point out that two legal cases should be regarded as similar if they jointly cite a common statute or precedents, and also if they cite different statutes or precedents which are structurally similar in their proposed Hier-SPCNet network.

**Interpretability and knowledge injection.**    Two further challenges regard interpretability aspects and knowledge modeling [Zhong et al., 2020]. In Section 4.8, we have discussed about the existence of a number of approaches recently developed to improve explainability and interpretability of TLM-based methods, which can broadly be categorized into post-hoc explanation methods and early explanation methods. Besides that, it is important to define specific legal requirements on the interpretability of machine learning models applied to private and public decision making, as discussed in [Bibal et al., 2021]. Also, [Branting et al., 2021] highlight the trade-off between explanation quality and representation effort, stating that a key requirement for explainable systems is the ability to obtain useful and comprehensible predictions with low costs in terms of development, testing, and maintenance at scale. In this respect, knowledge injection can play a key role in improving not only the language understanding ability of TLMs but also their explainability and interpretability — recently, the term *augmented language models* has been introduced in [Mialon et al., 2023] to also refer to those approaches that aim to enhance TLMs with reasoning skills and other external tools. [Liu et al., 2021b] introduce the use of causal graphs to ensure explainability requirements for a task of similar charge disambiguation. The proposed model, named *GCI*, aims to capture explainable discriminative nuances among confusing charges through building a causal graph to detect keywords from the charges, where nodes are obtained from the charges and by clustering similar keywords, and edges (i.e., the causal relationships) are learned using a causal discovery algorithm. The graph is sampled in more graphs that are refined by estimating the strength of the relationships. GCI decides which charge is more suitable for a case by extracting keywords from fact description and mapping the case according to the graph. This approach has been integrated into LSTM models to assess its potential and improve interpretability, where in particular, causal strength constraints are included into the attention weights of the neural network. Knowledge modeling mainly refers to properly utilizing the legal knowledge, and in this regard, an increasing number of works carry out a domain-adaptation pre-training process of a TLM in the attempt of better internalizing legal knowledge in the model.

Unfortunately, as we previously mentioned, the ability of successfully modeling the legal language and capturing peculiar patterns to the domain is strongly related to the exploitation of large, possibly multiple, legal data sources. Actually, the amount of existing legal data is considerable but many resources are only available to large companies or court sections. For example, it is unlikely that the whole history of a section of a court can be recovered, which would be very useful for many tasks like court profiling or case recommendation to courts, or even to characterize the language and creation style of arguments by lawyers, which can be supportive in the preparation of a lawyer's pleading. In [Francesconi, 2022], it is pointed out that the knowledge available in the Semantic Web is essential for the AI applied to legal domain, since it provides knowledge models for a top-down approaches (legal knowledge representation, legal reasoning, planning, explainability), and data for a bottom-up approach (argument mining, rule-based/case-based systems, legal information retrieval and discovery). [Gan et al., 2021] suggest to inject legal knowledge into deep neural networks by including first-order logic rules, motivated by the fact that logic rules yield models with inductive inclination and, thus, can alleviate the dependency of deep neural networks on high amounts of training data, besides making them more interpretable because of the presence of the rules. A model for LJP on private loan cases is proposed as a co-attention network followed by a symbolic module, where the co-attention network exploits the relations between claims and fact descriptions and provide the probability distribution for judgments, and the symbolic module adapts the distribution according to logic rules to prevent outputs that violate the law. Moreover, non-differentiability of first-order rules is ensured by associating continuous real-values to the outputs of logic rules with mapping functions. The rules are then injected with a reward mechanism, i.e., the output of the co-attention network is increased, resp. decreased, if the facts in the text satisfy, resp. violate, the conditions. Results show that the gain in injecting legal knowledge into TLMs such as BERT and RoBERTa increases as the size of the training set decreases.

**Ethical aspects.**   Another critical point in the use of AI-based technologies for the legal domain is related to *ethics*. Specific recommendations for the legal NLP community are provided in [Tsarapatsanis and Aletras, 2021] according to three ethical parameters: the importance of academic freedom, the diversity of ethical and legal norms and the threat of moralism. In [Zhong et al., 2020], it is reminded that the purpose of AI is not to replace humans in legal matters but only to provide support in decision-making processes, but it is also pointed out that applying AI systems to law can inadvertently lead to ethical problems, particularly *bias* of different types such as gender and racial discrimination. Being aware of such issues and trying to alleviate them is an emergence for developing next-generation (legal-focused) TLMs. [Henderson et al., 2022] highlight the difficulty in filtering toxic and private information in data used to train a model, since its removal could affect the meaning of the text; moreover, privacy expectations can be different based on the specific country. In this regard, filtering rules should be designed to reflect the standards developed by legal and administrative experts. [Chalkidis et al., 2022c] measure the fairness of BERT-based models on three categories, namely demographics, regional and legal topic: for the first category, the goal is to evaluate if a model performs worse because it is biased by factors like gender, age, race, language, legal status, whereas for the other two categories, the goal is to assess if a model performs differently on cases associated with courts of specific regions, or in a specific field of law (e.g., it may perform better on criminal cases than on civil law cases). In this respect, some group disparity in performance is found as related to the defendant's state (Central European states versus the other European states), applicant's gender, language, legal areas, and court's regions (e.g., Switzerland courts versus federation courts, and Beijing courts versus Sichuan courts); however, some group disparities can also be influenced by general factors based on the distribution of training data. [Wang et al., 2021] measure the judicial inconsistency related to attributes like gender, race, and region, as the average disagreement of the judgments (term penalty) given by LJP models, which are used as virtual judges. The judicial disagreement is defined as the standard deviation of virtual judges' results. Results on the CAIL data are produced according to region and gender: in the first case, seven virtual judges are trained on data belonging to seven provincial-level administrative regions, whereas in the second case, two virtual judges are trained, one for each defendant's gender. Gender and, especially, regional inconsistency are found in the legal system, with regional inconsistency varying over time; Moreover, judicial inconsistency seems to be negatively correlated with the severity of criminal charges.

Finally, a special remark should be made concerning all the hype of large generative language models after the launch of ChatGPT in November 2022. Despite limitations of ChatGPT and similar tools have soon been detected (e.g., possibility of writing plausible-sounding but incorrect, misinformed or nonsensical answers, sensitivity to tweaks to the input phrasing, verbosity, possibility of responding to harmful instructions or exhibiting biased behavior, etc.), we have witnessed a technological shift in the way we work: by using ChatGPT, people build websites and apps, write novels or technical papers, even pass college and university exams, from medical degree to law degree. This rise in "bad" actors abusing of generative contents has prompted to start developing countermeasures to debunk artificially generated text as well as to check data security requirements (e.g., on March 2023, Italian data-protection authority temporarily banned ChatGPT over concerns about breaching of existing EU privacy rules). This and much more opens to further challenges that are going to be a major point of interest for researchers involved in legal AI.

# 6 Conclusions

In this work, we investigated the research advances in AI for law that have been accomplished by means of Transformer-based language models (TLMs), pushed since the advent of BERT. This is the first systematic study on this topic developed around problems, tasks and benchmarks in the legal AI area, for which we covered about thirty different TLMs used to build more than three hundred AI approaches and methods addressing retrieval, classification, prediction, entailment, summarization, generation, information extraction, and many other tasks relevant in the legal domain. Our survey included critical aspects in the design of TLM-based methods for legal AI, such as different strategies for adaptation to the legal domain, dealing with low-resource natural languages as well as with multilingual contexts. We discussed main findings and limitations of current TLM-based methods, and open challenges and future perspectives for next generation of legal AI tools. Moreover, we considered details on the implementation of TLM-based methods, providing a large number of references to the software resources available for such methods. In this regard, we envisage an interesting opportunity for the research community to gather and manage a publicly shared platform for all TLM-based systems for AI and law.